\documentclass[jair,twoside,11pt,theapa]{article}
\usepackage{jair, theapa, rawfonts}
\usepackage{amsmath}
\usepackage{amsthm}
\usepackage{amssymb}
\usepackage{latexsym}
\usepackage{mathrsfs}
\usepackage{bbm}
\usepackage[pointlessenum]{paralist}

\theoremstyle{definition} 
\newtheorem{Defn}{Definition}[section]
\newtheorem{Refine}{Refinement}[section]
\newtheorem{Lem}[Defn]{Lemma}
\newtheorem{Thm}[Defn]{Theorem}
\newtheorem{Eg}{Example}[section]
\newtheorem{Prin}{Principle}[section]
\newtheorem{Eval}{Evaluation}[subsection]

\newcommand{\defn}{\textbf}
\newcommand{\pf}{\textbf}
\newcommand{\eop}{\textbf}

\newcommand{\nl}{\newline}

\newcommand{\strArr}{\ensuremath{\rightarrow}} 
\newcommand{\defArr}{\ensuremath{\Rightarrow}} 
\newcommand{\warnArr}{\ensuremath{\leadsto}}   

\newcommand{\nnot}{\ensuremath{\mathord{\sim}}} 
\newcommand{\AND}{\ensuremath{\mathord{\raisebox{0.4ex}{\ensuremath{\scriptstyle \bigwedge}}}}} 
\newcommand{\OR}{\ensuremath{\mathord{\raisebox{0.4ex}{\ensuremath{\scriptstyle \bigvee}}}}} 
\newcommand{\e}{\ensuremath{\! \in \!}}      
\newcommand{\nte}{\ensuremath{\! \notin \!}} 
\newcommand{\vsim}{\ensuremath{\mid\!\sim}} 
\newcommand{\vminus}{\ensuremath{\mid\!\!\!-\,}} 

\newcommand{\calP}{\ensuremath{\mathcal{P}}} 
\newcommand{\calPd}{\ensuremath{\mathcal{P}\!_d}} 
\newcommand{\ZZ}{\ensuremath{\mathbb{Z}}} 
\newcommand{\T}{\ensuremath{\textbf{\textsf{T}}}} 
\newcommand{\F}{\ensuremath{\textbf{\textsf{F}}}} 
\newcommand{\tv}[1]{\ensuremath{\textbf{\textsf{#1}}}} 
\newcommand{\bbt}{\ensuremath{\mathbbm{t}}} 
\newcommand{\rse}{\ensuremath{\mathbbm{r}}} 

\newcommand{\Alg}{\ensuremath{\mathit{Alg}}} 
\newcommand{\alg}{\ensuremath{\mathit{alg}}} 
\newcommand{\arrow}{\ensuremath{\mathit{arrow}}} 
\newcommand{\Atm}{\ensuremath{\mathit{Atm}}} 
\newcommand{\Ax}{\ensuremath{\mathit{Ax}}} 
\newcommand{\Claus}{\ensuremath{\mathit{Claus}}} 
\newcommand{\Cor}{\ensuremath{\mathit{Cor}}} 
\newcommand{\CorRes}{\ensuremath{\mathit{CorRes}}} 
\newcommand{\Ctge}{\ensuremath{\mathit{Ctge}}} 
\newcommand{\Dftd}{\ensuremath{\mathit{Dftd}}} 
\newcommand{\Err}{\ensuremath{\mathit{Err}}} 
\newcommand{\Fact}{\ensuremath{\mathit{Fact}}} 
\newcommand{\Foe}{\ensuremath{\mathit{Foe}}} 
\newcommand{\For}{\ensuremath{\mathit{For}}} 
\newcommand{\Fml}{\ensuremath{\mathit{Fml}}} 
\newcommand{\From}{\ensuremath{\mathit{From}}} 
\newcommand{\Hist}{\ensuremath{\mathit{Hist}}} 
\newcommand{\Lit}{\ensuremath{\mathit{Lit}}} 
\newcommand{\Min}{\ensuremath{\mathit{Min}}} 
\newcommand{\op}{\ensuremath{\mathit{op}}} 
\newcommand{\Plaus}{\ensuremath{\mathit{Plaus}}} 
\newcommand{\pv}{\ensuremath{\mathit{pv}}} 
\newcommand{\Res}{\ensuremath{\mathit{Res}}} 
\newcommand{\Rul}{\ensuremath{\mathit{Rul}}} 
\newcommand{\Sat}{\ensuremath{\mathit{Sat}}} 
\newcommand{\Smp}{\ensuremath{\mathit{Smp}}} 
\newcommand{\smp}{\ensuremath{\mathit{smp}}} 
\newcommand{\SmpMinCtge}{\ensuremath{\mathit{SmpMinCtge}}} 
\newcommand{\SmpRes}{\ensuremath{\mathit{SmpRes}}} 
\newcommand{\Subj}{\ensuremath{\mathit{Subj}}} 
\newcommand{\Taut}{\ensuremath{\mathit{Taut}}} 
\newcommand{\Thms}{\ensuremath{\mathit{Thm}}} 
\newcommand{\Val}{\ensuremath{\mathit{Val}}} 

\ShortHeadings{Principles and Examples of Plausible Reasoning and 
Propositional Plausible Logic}
{Billington 
}

\begin{document}

\title{Principles and Examples of Plausible Reasoning \\
and Propositional Plausible Logic}

\author{\name David Billington 
\email d.billington@griffith.edu.au \\
\addr School of Information and Communication Technology, Nathan campus,\\
      Griffith University, Brisbane, Queensland 4111, Australia.}


\maketitle

\begin{abstract} 
Plausible reasoning concerns situations whose 
inherent lack of precision is not quantified; 
that is, there are no degrees or levels of precision, 
and hence no use of numbers like probabilities. 
A hopefully comprehensive set of principles that 
clarifies what it means for a formal logic to do 
plausible reasoning is presented. 
A new propositional logic, 
called Propositional Plausible Logic (PPL), 
is defined and applied to some important examples. 
PPL is the only non-numeric non-monotonic logic we know of that 
satisfies all the principles and 
correctly reasons with all the examples. 
Some important results about PPL are proved. 

\end{abstract}

\section{Introduction}
\label{Section:Introduction}


We are interested in reasoning about situations that 
(a) have imprecisely defined parts, and 
(b) this lack of precision is not quantified. 
That is, there are no degrees or layers or levels of precision, 
and in particular there are no numbers like probabilities, 
that would quantify the lack of precision. 
These situations are often indicated by the ordinary, rather than technical, 
use of words such as `mostly', `usually', `typically', `normally', `probably', 
`likely', `plausible', `believable', and `reasonable'. 
Although these words are not synonymous, they share a common property, 
which may be expressed by using 
either frequency of occurrence or weight of evidence. 
In frequency terms the property is that something is true more often than not; 
in evidence terms the property is that 
the evidence for something outweighs the evidence against it. 
An example is `Mammals usually are non-venomous'. 

We shall call these situations \textit{plausible-reasoning situations} 
because we shall call the reasoning used in such situations 
\textit{plausible reasoning}. 

This article has two aims. 
The first is to introduce principles that give a much clearer understanding of 
what it means for a formal logic to do plausible reasoning; 
that is the kind of reasoning indicated above. 
The hope is that this set of principles is comprehensive. 
Whether it is or not, this seems to be the first such set of principles 
even though plausible reasoning has been used for 
at least 2500 years \cite{WTG:2014}. 
However on page 114 of \cite{WTG:2014} there is 
a list of 11 characteristics of plausible reasoning, 
rather than characteristics of formal logics that do plausible reasoning. 

The second aim is to define a propositional logic, 
called Propositional Plausible Logic (PPL), 
that satisfies all these principles of plausible reasoning. 
This shows that all the principles together are consistent; that is, 
there is no principle whose negation is implied by all the other principles. 
A pruned version of PPL is presented in \cite{Billington:2015}. 

In this paper we shall be considering only propositional situations; 
that is, situations that can be fully represented by a 
\textit{propositionally adequate language}. 
That is, by a language which has an adequate set of propositional connectives. 
The most common adequate sets of connectives contain negation and 
at least one of conjunction, or disjunction, or material implication. 
The connectives we shall use are negation $\neg$, conjunction $\AND$, 
and disjunction $\OR$. 
A \textit{propositionally adequate logic} is a logic based on a 
propositionally adequate language. 

Moreover we shall consider only those plausible-reasoning situations that 
can be specified by a plausible-structure 
\,$\mathcal{S} = (\Fact(\mathcal{S}), \Plaus(\mathcal{S}))$\, 
where $\Fact(\mathcal{S})$ is a set of propositional formulas 
representing the factual part of $\mathcal{S}$, 
and $\Plaus(\mathcal{S})$ is a set 
representing the plausible part of $\mathcal{S}$. 
The elements of $\Plaus(\mathcal{S})$ can have a variety of forms; 
for example: defaults are used in Reiter's Default Logic \cite{Reiter:1980}, 
defeasible rules are used in ASPIC \cite{CA:2007} and ASPIC$^+$ \cite{MP:2013}, 
and defeasible and warning rules are used in 
Defeasible Logic \cite{Billington:2008}. 
The plausible-structure syntax is very general, while being specific enough 
to permit the definition of concepts needed later. 

This article is organised into the following sections. 
The next section defines some ideas and notation from 
classical propositional logic that are needed in 
Sections \ref{Section:Principles} and \ref{Section:PPL}. 
Section \ref{Section:Principles} presents the principles of plausible reasoning. 
Section \ref{Section:Some Non-Monotonic Logics} contains a survey of 
various non-numeric non-monotonic logics. 
The definition of PPL is in Section \ref{Section:PPL}. 
In Section \ref{Section:Examples} we apply PPL to some examples. 
In Section \ref{Section:Properties of PPL} we state and discuss some 
important properties of PPL, and show that PPL satisfies 
all the principles in Section \ref{Section:Principles}. 
Section \ref{Section:Conclusion} is the conclusion. 
All the proofs are in the appendices. 

\section{A Classical Propositional Logic using Resolution}
\label{Section:Set-based formulas}
Formulas in classical propositional logics are usually defined using sequences, for example \,$(a \vee (b \vee a))$. 
Sequences make unwanted distinctions which are often best removed, 
for example neither order nor repetitions are needed.
So the above example is more clearly written as \,$\OR\{a,b\}$. 
We shall define classical propositional formulas 
that are based on sets rather than sequences. 
Such set-based formulas simplify the definition of resolution. 
The classical notions of truth value, valuation, satisfaction, 
semantic consequence $\models$, tautology, contradiction, 
equivalence of formulas, and resolution are as usual. 

Let us start by agreeing on some notation. 
As usual `iff' abbreviates `if and only if'. 
$X$ is a subset of $Y$ is denoted by \,$X \!\subseteq\! Y$; 
the notation \,$X \!\subset\! Y$ \ means \,$X \!\subseteq\! Y$\, and 
\,$X \!\neq\! Y$, and denotes that $X$ is a \defn{strict subset} of $Y$. 
The empty set is denoted by $\{\}$, and the set of all integers by $\ZZ$. 
The cardinality of a set $S$ is denoted by $|S|$. 
If $m$ and $n$ are integers then we define the 
\defn{integer interval} $[m..n]$ by 
\,$[m..n] = \{i \e \ZZ : m \leq i \leq n\}$. 

Our \defn{alphabet} is the union of the following 
pairwise disjoint sets of symbols: 
a non-empty countable set, \Atm, of (propositional) atoms; 
the set $\{\neg, \AND, \OR \}$ of connectives with $\neg, \AND, \OR$
denoting negation, conjunction, and disjunction respectively; 
and the set of punctuation marks consisting of the comma and both braces. 

We now define a formula. 

\begin{Defn} \label{Defn:formula} \ 
\begin{compactenum}[1)]
\item \label{An atom is a formula.} 
	If $a$ is an atom then $a$ is a \defn{formula}. 
\item \label{A negated formula is a formula.} 
	If $f$ is a formula then $\neg f$ is a \defn{formula}. 
\item \label{AND F, VF are formulas.} 
	If $F$ is a finite set of formulas then $\AND F$ is a \defn{formula} 
	and $\OR F$ is a \defn{formula}. 
\item \label{All formulas are built by ...} 
	Every formula can be built by a finite number of applications of 
	(1), (2), and (3). 
\end{compactenum}
The set of all formulas is denoted by $\Fml$.  
\end{Defn} 

It is convenient to write $\AND F$ and $\OR F$ even though 
the set $F$ of formulas may be infinite. 

The next three definitions define some special formulas, 
the set of all literals in a clause or dual-clause, and 
the complement of a literal. 

\begin{Defn} \label{Defn:Kinds of formula} \ 
\begin{compactenum}[1)]
\item \label{A literal is ...} 
	The set, $\Lit$, of all \defn{literals} is defined by 
	\,$\Lit = \Atm \cup \{\neg a : a \e \Atm\}$.  
\item \label{A clause is ...} 
	A \defn{clause} is either a literal or 
	the disjunction, $\OR L$, of a finite set, $L$, of literals.  
\item \label{The falsum is ...} 
	$\OR \{\}$ is the \defn{empty clause} or \defn{falsum}. 
\item \label{A dual-clause is ...} 
	A \defn{dual-clause} is either a literal or 
	the conjunction, $\AND L$, of a finite set, $L$, of literals.  
\item \label{The verum is ...} 
	$\AND \{\}$ is the \defn{empty dual-clause} or \defn{verum}. 
\item \label{contingent} 
	A formula is \defn{contingent} iff 
	it is not a tautology and it is not a contradiction. 
\end{compactenum}
\end{Defn} 

\begin{Defn} \label{Defn:Lit(.)} \ 
\begin{compactenum}[1)]
\item \label{Lit(l) = ...} 
	If $l$ is a literal then \,$\Lit(l) = \{l\}$. 
\item \label{Lit(VL) = L = Lit(AND L)} 
	If $L$ is a finite set of literals then \,$\Lit(\OR L) = L = \Lit(\AND L)$. 
\end{compactenum}
\end{Defn} 

\begin{Defn} \label{Defn:complement}
Let $a$ be any atom and $L$ be any set of literals. 
\begin{compactenum}[1)]
\item \label{complement a} 
	The \defn{complement} of $a$, $\nnot a$, is defined by 
	\,$\nnot a = \neg a$. 
\item \label{complement not a} 
	The \defn{complement} of $\neg a$, $\nnot \neg a$, is defined by 
	\,$\nnot \neg a = a$. 
\item \label{complement L} 
	The \defn{complement} of $L$, $\nnot L$, is defined by 
	\,$\nnot L = \{\nnot l : l \e L\}$. 
\end{compactenum}
\end{Defn} 

Let $C$ be any set of clauses. 
We want to remove from $C$ all the clauses that, when removed, 
will not change the truth value of $\AND C$. 
By Lemma~\ref{Lem:[dual-]clauses, implication}(\ref{clauses, implication}) 
(See Appendix A) this means removing all tautologies and 
all clauses that have a strict subclause in $C$. 
We also want to simplify $C$ by replacing any clause $\OR \{l\}$ in $C$ by $l$. 
The result will be the core of $C$. 

Dually, let $D$ be any set of dual-clauses. 
We want to remove from $D$ all the dual-clauses that, when removed, 
will not change the truth value of $\OR D$. 
By Lemma~\ref{Lem:[dual-]clauses, implication}(\ref{dual-clauses, implication}) 
(See Appendix A) this means removing all contradictions and 
all dual-clauses that have a strict sub-dual-clause in $D$. 
We also want to simplify $D$ by 
replacing any dual-clause $\AND \{l\}$ in $D$ by $l$. 
The result will be the core of $D$. 

The following definition does both of these. 

\begin{Defn} \label{Defn:Ctge,Min,Smp,Cor} 
Let $G$ be either a set of clauses or a set of dual-clauses. 
\begin{compactenum}[1)]
\item \label{Ctge(G)} 
	The set of elements of $G$ that are contingent or empty, $\Ctge(G)$, 
	is defined by \nl $\Ctge(G) = \{g \e G : g$ is contingent or empty$\}$. 
\item \label{Min(G)} 
	The set of minimal elements of $G$, $\Min(G)$, is defined by \nl
	\,$\Min(G) = \{g \e G$ : if \,$g' \e G$\, then 
	\,$\Lit(g')\!\not\subset\!\Lit(g)\}$. 
\item \label{smp(g)} 
	The \defn{simplification} of the formula $f$, $\smp(f)$, 
	is defined as follows. \nl 
	If \,$f \e \{\AND \{g\}, \OR \{g\}\}$, where $g$ is a formula, then
	\,$\smp(f) = \smp(g)$; \ else \,$\smp(f) = f$. 
\item \label{Smp(G)} 
	The simplification of $G$, $\Smp(G)$, is defined by 
	\,$\Smp(G) = \{\smp(g) : g \e G\}$. 
\item \label{Cor(G)} 
	The \defn{core} of $G$, $\Cor(G)$, is defined by 
	\,$\Cor(G) = \Smp(\Min(\Ctge(G)))$. 
\end{compactenum}
\end{Defn} 

Let $C$ be any set of clauses.  
The set of all clauses derivable by resolution from clauses in $C$ is 
denoted by $\Res(C)$. 
Also we usually abbreviate $\Cor(\Res(C))$ to $\CorRes(C)$, 
and $\Smp(\Res(C))$ to $\SmpRes(C)$. 

We shall need to convert a formula $f$ into a set $\Claus(f)$ of clauses, 
such that $\AND\Claus(f)$ is equivalent to $f$. 
Unfortunately there are many such sets of clauses, 
so we shall follow Sections 2 and 3 of Chapter I of \cite{NS:1997} 
to define which set we mean. 

We shall denote the true truth value by $\T$ and the false truth value by $\F$. 
Let $\Val$ denote the set of all valuations, and 
$\Atm(f)$ denote the set of atoms in the formula $f$. 

\begin{Defn} \label{Defn:Val(A)} 
If $A$ is a set of atoms then define $\Val(A)$, 
the set of valuations which are false outside $A$, by 
\,$\Val(A) = \{v \e \Val$ : for all $a$ in $\Atm \!-\! A$, \,$v(a) = \F\}$. 
\end{Defn} 

So if \,$|A| = n$\, then \,$|\Val(A)| = 2^n$. 

\begin{Defn} \label{Defn:L(f,v),Claus(f)} 
Let $f$ be any formula, $F$ any set of formulas, and $v$ any valuation. 
\begin{compactenum}[1)]
\item \label{L(f,v)=...} 
	Define \,$L(f,v) = \{a : a \e \Atm(f)$ and $v(a)$ = \T$\}$ $\cup$ 
	$\{\neg a : a \e \Atm(f)$ and $v(a)$ = \F$\}$. 
\item \label{Claus(f)=...} \index{Claus@$\Claus(.)$} 
	Define \,$\Claus(f) = 
	\{\OR \nnot L(f,v) : v \e \Val(\Atm(f))$ and $v(f) = \F\}$. 
\item \label{Claus(F)=...} \index{Claus@$\Claus(.)$} 
	Define \,$\Claus(F) = \bigcup\{\Claus(f) : f \e F\}$. 
\end{compactenum}
\end{Defn} 

If a set $F$ of formulas is unsatisfiable then 
classical propositional logic can prove any formula from $F$; 
that is, classical propositional logic is \textit{explosive}. 
Explosive logics are not ideal because, for unsatisfiable sets of formulas, 
the idea of `proof' becomes worthless. 
For example, let \,$F_1 = \{a,\neg a, b\}$. 
Then from $F_1$ an explosive logic can prove $\neg b$, 
which does not seem sensible. 
Logics that are not explosive are called \textit{paraconsistent} logics. 
But we want more than mere paraconsistency. 
In the example above, either $a$ or $\neg a$ seems to be a mistake, 
so it would be reasonable that from $F_1$ we can conclude only $b$, 
and what follows from $b$. 

Let $F$ be an unsatisfiable set of formulas. 
There are several ways of getting a satisfiable subset of $F$ 
(see the literature on Belief Revision and Paraconsistent Logics). 
One way is to take the 
intersection of all the maximal satisfiable subsets of $F$, 
(this is the full meet contraction of $F$ by a contradiction). 
An equivalent way is to remove all the minimal unsatisfiable subsets of $F$. 
For example, let \,$F_2 = \{a, \neg a, \OR\{a,b\}\}$. 
Then the (only) minimal unsatisfiable subset of $F_2$ is \,$\{a, \neg a\}$. 
Removing this from $F_2$ leaves \,$\{\OR\{a,b\}\}$. 
Alternatively the intersection of all the maximal satisfiable subsets of $F_2$ 
is also \,$\{\OR\{a,b\}\}$. 

However, another way to get a satisfiable subset of $F$ is 
to remove all the formulas of $F$ that are `contaminated' by potential errors, 
as follows. 
First we convert $F$ to the set of clauses $\Claus(F)$.  
Then $F$ is unsatisfiable iff $\Claus(F)$ is unsatisfiable iff 
$\Res(\Claus(F))$ contains a literal, say $l$, and its complement, $\nnot l$. 
At least one of $l$ and $\nnot l$ is an error. 
Certainly both $l$ and $\nnot l$ are potential errors.
Potential errors contaminate any clause containing them, 
making the clause unreliable. 
We then remove all the contaminated clauses from $\Claus(F)$ 
to get the result $\Sat(\Claus(F))$. 

Applying this to $F_2$ we see that 
\,$\Claus(F_2) = \{\OR\{a\}, \OR\{\neg a\}, \OR\{a,b\}\}$\, and the set, 
$\Err(\Claus(F_2))$, of potential error literals of $\Claus(F_2)$ is 
\,$\Err(\Claus(F_2)) = \{a, \neg a\}$. 
Hence very clause in $\Claus(F_2)$ is contaminated by a potential error, 
and so \,$\Sat(\Claus(F_2)) = \{\}$. 

The formal definition of the functions $\Err(.)$ and $\Sat(.)$ follows. 

\begin{Defn} \label{Defn:Err,Sat} \raggedright
Let $C$ be any set of clauses. \nl
\,$\Err(C) = \{l \e \Lit : \{l,\nnot l\} \!\subseteq\! \SmpRes(C)\}$. \nl
\,$\Sat(C) = \{c \e C : c \neq \OR\{\}$\, and 
\,$\Lit(c) \!\cap\! \Err(C) = \{\}\}$. 
\end{Defn} 

Of course if $C$ is satisfiable then \,$\Sat(C) = C$. 
If $C$ is any set of clauses then $\Sat(C)$ is satisfiable, 
and so \,$\Sat(\Sat(C)) = \Sat(C)$. 

Let $C$ be a set of clauses. 
It can be shown that every literal in every clause in every minimal 
unsatisfiable subset of $C$ is a potential error literal 
and so is in $\Err(C)$. 
But $\Sat(C)$ removes all the clauses that contain any potential error literal, 
not just the clauses that are composed entirely of potential error literals. 
Hence the clauses in $\Sat(C)$ may be regarded as at least 
as reliable as the clauses in the intersection of 
all the maximal satisfiable subsets of $C$, 
as some of these clauses may contain potential error literals, 
as happens with $F_2$. 

We can now define our paraconsistent propositional logic by 
mimicking the standard definition of proof by resolution, 
which we shall also define. 

\begin{Defn} \label{Defn:|-, ||-} \raggedright 
Let $F$ be any set of formulas and $f$ be any formula. \nl
As usual, define $F$ \defn{proves} $f$ (by resolution), 
denoted by \,$F \vdash f$, as follows. \nl
$F \vdash f$\, iff \,$\OR\{\} \in \Res(\Claus(\{\neg f\} \!\cup\! F))$. \nl
Define $F$ \defn{judiciously proves} $f$, denoted by \,$F \Vdash f$, 
as follows. \nl
$F \Vdash f$\, iff 
\,$\OR\{\} \in \Res(\Claus(\neg f) \!\cup\! \Sat(\Claus(F)))$. 
\end{Defn} 

Explosiveness is a symptom of the fact that, from a set $F$ of formulas, 
classical logic proves formulas that, arguably, do not follow from $F$. 
For example, if $a$ and $b$ are different atoms then 
from the contradiction $\AND\{a, \neg a\}$ we can prove $b$. 
Because $b$ has nothing to do with $a$ or $\neg a$, 
our intuition is that $b$ does not follow from $\AND\{a, \neg a\}$. 
Although `follows from' is an intuitive concept rather than a formal one, 
we shall attempt to formally define the concept. 
To refine our intuition let us consider tautologies. 

Let $F$ be a set of formulas, $f$ and $g$ be formulas, 
$L$ and $M$ be finite sets of literals, and $\bbt$ be a tautology. 

\begin{compactenum}[A)]
\item \label{Tautologies are independent} 
	We could argue that tautologies stand on their own, 
they do not depend on any other formula. 
So if \,$\bbt \nte F$\, then $\bbt$ does not follow from $F$. 

\item \label{Syntax independence} 
	We would like `follows from' to be syntax independent. 
That is, if $f$ follows from $F$ and 
$f$ is equivalent to $g$ then $g$ follows from $F$. 

\item \label{isin implies follows from} 
	If \,$f \e F$\, then it seems reasonable that $f$ follows from $F$. 

\item \label{superclause follows from subclause} 
	If $\OR L$ follows from $F$ and \,$L \subseteq M$ 
then it seems reasonable that $\OR M$ follows from $F$. 

\item \label{Syntax dependence example} 
	If $a$ and $b$ are different atoms then it seems reasonable that 
$\OR\{a, \neg a\}$ does not follow from $\{\OR\{b, \neg b\}\}$. 
By (\ref{isin implies follows from}), 
$\OR\{b, \neg b\}$ follows from $\{\OR\{b, \neg b\}\}$. 
But this would make `follows from' syntax dependent, 
contrary to (\ref{Syntax independence}). 
\end{compactenum}
So tautologies present difficulties for any definition of `follows from'. 
However, in Section \ref{Section:Principles} 
we do not want to force all tautologies to be provable, 
so we shall declare that 
tautologies do not follow from any set of formulas. 
Thus we arrive at the following definition. 

\begin{Defn} \label{Defn:From(.)} \raggedright 
Let $F$ be any set of formulas. 
Define $\Taut$ to be the set of all tautologies. 
Define the set of formulas that follow from $F$, $\From(F)$, by 
\,$\From(F) = \{f \e \Fml : F \Vdash f\} - \Taut$. 
\end{Defn} 

Since \,$\Sat(\Claus(F)) \subseteq\Claus(F)$\, we have \nl
\,$\From(F) \subseteq \{f \e \Fml : F \Vdash f\} 
\subseteq \{f \e \Fml : F \vdash f\}$. 

\section{Principles of Plausible Reasoning} 
\label{Section:Principles}

Lists of postulates, properties, or principles that concern special types 
of reasoning are useful for at least the following reasons. 

\begin{compactenum}[1)] \raggedright
\item They help \textit{characterise} the intended special type of reasoning. 
\item They provide a means of \textit{evaluating} existing reasoning systems to see how well they perform the intended special type of reasoning. 
\item They provide \textit{guidelines} for creating new reasoning systems for the intended special type of reasoning. 
\item They explicitly show a \textit{difference} between the intended special type of reasoning and an existing form of reasoning. 
\end{compactenum}
Notable examples of such lists are the following. 
The AGM postulates for belief change \cite{AGM:1985,Gardenfors:1988}, 
various properties of non-monotonic consequence relations 
\cite{Makinson:1988,KLM:1990}, and the postulates that a 
rule-based argumentation system should satisfy \cite{CA:2007}. 

We shall state the principles of this section by 
referring to the logic or proof algorithm directly; 
rather than by referring to consequence relations. 
A consequence relation, say $\vsim$, relates 
a set $F$ of formulas to a formula $f$; 
where \,$F \vsim f$\, means that $f$ is a consequence of $F$. 
Consequence relations are appropriate if the reasoning situations 
under consideration can be characterised by a set of formulas. 
But the plausible-reasoning situations we consider 
are specified by a plausible-structure 
\,$\mathcal{S} = (\Fact(\mathcal{S}), \Plaus(\mathcal{S}))$\, 
where the elements of $\Plaus(\mathcal{S})$ may be very different from 
the formulas in $\Fact(\mathcal{S})$. 
For these situations consequence relations are much less appropriate. 
For example consider two fundamental properties that consequences relations 
may have; namely cut and cautious monotonicity, which together are 
equivalent to cumulativity, also called lemma addition. 
If $F$ and $G$ are sets of formulas then let \,$F \vsim G$\, 
mean for all $g$ in $G$, \,$F \vsim g$.  
Then cumulativity is the following property. 
If \,$F \vsim G$\, then for all formulas $h$, 
\,$F \vsim h$\, iff \,$F \!\cup\! G \vsim h$.  
A straightforward translation of \,$F \vsim f$\, into our situation is 
\,$\mathcal{S} \vsim f$, where $\mathcal{S}$ is a plausible-structure. 
But then it is really hard to know what \,$F \!\cup\! G$\, might mean. 
Essentially we are trying to add proved formulas to $\mathcal{S}$. 
But the only set of formulas in $\mathcal{S}$ is $\Fact(\mathcal{S})$. 
So we could try letting \,$F \!\cup\! G$\, be 
\,$(\Fact(\mathcal{S}) \cup G, \Plaus(\mathcal{S}))$. 
But this is only sensible when the formulas in $G$ have been proved using 
only $\Fact(\mathcal{S})$. 
When the formulas in $G$ have been proved using $\Plaus(\mathcal{S})$ then 
it is no longer sensible to treat the formulas in $G$ as facts; 
because they are not facts, they are only plausible conclusions. 

Some of the principles of plausible reasoning are regarded as 
necessary and so use the word `must';  
the other principles are regarded as desirable and so use the word `should'. 

As well as the principles of plausible reasoning, 
we shall present several plausible-reasoning examples. 
Some of these examples are based on an \defn{$n$-lottery}, that is, 
randomly selecting a number from the finite integer interval $[1..n]$. 
We shall use $s_i$ to denote that the number $i$ was selected. 
Four examples will guide the development of some of the principles, 
and so we shall call these examples signpost examples. 
Our first signpost example is the 3-lottery example. 

\begin{Eg}[The 3-lottery example] \ 
\label{Eg:3-lottery}
Consider a 3-lottery. 
Then we have the following. 
\begin{compactenum}[1)] \raggedright
\item Exactly one element of $\{s_1, s_2, s_3\}$ is true. 
\item Each element of $\{s_1, s_2, s_3\}$ is probably false. 
\item The disjunction of any 2 elements of $\{s_1, s_2, s_3\}$ is probably true.  
\end{compactenum}
\end{Eg}

This example illustrates some important properties of plausible reasoning 
that will be considered in several of the following subsections. 

The following notation will be convenient. 
Let $\Thms(\mathcal{L},\alpha,\mathcal{S})$ denote the set of all 
formulas derivable from the plausible-structure $\mathcal{S}$ by 
using the proof algorithm $\alpha$ of the logic $\mathcal{L}$. 
If $F$ is a set of propositional formulas then $\Thms(F)$ denotes 
all the formulas derivable from $F$ by 
(the proof algorithm of) any classical propositional logic. 
This simpler notation is unambiguous because $\Thms(F)$ is independent of 
the logic (for example Hilbert systems, natural deduction, 
or resolution systems) and its proof algorithm. 

\subsection{Representation}
\label{Subsection:Representation}

Plausible-reasoning situations may contain facts 
as well as plausible information; 
for instance statement (1) of Example \ref{Eg:3-lottery} 
is a factual statement, unlike the other two statements which are plausible. 
Hence the first part of our first principle of plausible reasoning. 

Although the inherent lack of precision of plausible-reasoning situations 
is not quantified, 
a logic could represent this lack of precision with undue accuracy, 
for instance by using probabilities. 
Forbidding this is too restrictive, 
as the logic may deduce a conclusion using the probabilities but then 
present that conclusion without using probabilities. 
All we need is that the conclusions are not unduly precise. 
In particular if a formula is proved by using plausible information then 
it should not be regard as a fact. 
Hence the second part of our first principle of plausible reasoning. 

\begin{Prin}[The Representation Principle] \ 
\label{Prin:Representation} 
\begin{compactenum}[1)]
\item A logic for plausible reasoning must be able to represent, 
	and distinguish between, factual and plausible statements. 
\item The formulas proved by a logic for plausible reasoning 
	must not be more precise than the information used to derive them. 
\end{compactenum}
\end{Prin}

We note that when a situation is precisely described, 
perhaps using probabilities, 
a logic for plausible reasoning should be able to reason with the 
corresponding imprecisely defined situation; 
Example \ref{Eg:3-lottery} is such a situation. 

We infer from Principle \ref{Prin:Representation}(1) that 
we should be able to distinguish between conclusions 
that are factual and those that are merely plausible. 
One way of making this distinction is to have a 
\defn{factual proof algorithm} that only uses facts and deduces only facts, 
and also a \defn{plausible proof algorithm} that may use plausible statements 
and facts and deduces formulas that are only plausible. 
Of course if a plausible proof algorithm deduces only facts when given just 
facts then it can be regarded as both a factual and a plausible proof algorithm. 
The need for multiple proof algorithms is discussed further in 
Subsection \ref{Subsection:Ambiguity}. 

\subsection{Evidence and Non-Monotonicity}
\label{Subsection:Evidence, Non-Monotonicity}

Let us now see if we can establish some general guidelines 
concerning the provability of a given formula $f$. 
A plausible-reasoning situation will have evidence for and against $f$. 
So it seems reasonable to determine whether $f$ is provable or not 
by just comparing these two sets of evidence, 
and declaring $f$ provable iff the preponderance of evidence is for $f$. 

A consequence of the evidence criterion needs the following definitions. 
If $\mathcal{S}_1$ and $\mathcal{S}_2$ are plausible-structures 
then \,$\mathcal{S}_1 \!\subseteq\! \mathcal{S}_2$\, means 
\,$\Fact(\mathcal{S}_1) \!\subseteq\! \Fact(\mathcal{S}_2)$\, and 
\,$\Plaus(\mathcal{S}_1) \!\subseteq\! \Plaus(\mathcal{S}_2)$. 
A proof algorithm $\alpha$ of a logic $\mathcal{L}$ is said to be 
\defn{monotonic} iff for any two plausible-structures, 
$\mathcal{S}_1$ and $\mathcal{S}_2$, 
if \,$\mathcal{S}_1 \!\subseteq\! \mathcal{S}_2$\, then 
\,$\Thms(\mathcal{L},\alpha,\mathcal{S}_1) \!\subseteq\! 
\Thms(\mathcal{L},\alpha,\mathcal{S}_2)$. 
For example, the proof algorithm of a classical propositional logic 
is monotonic. 
A proof algorithm is \defn{non-monotonic} iff it is not monotonic. 
A plausible proof algorithm is non-monotonic because 
the addition of evidence against a previously provable formula 
can cause it to be unprovable, as shown in our second signpost example. 

\begin{Eg}[The Non-Monotonicity example] \ 
\label{Eg:Non-Monotonicity}
Consider the following two statements. 
The first is plausible and the second is factual. 
\begin{compactenum}[1)]
\item $a$ is probably true. 
\item $\neg a$ is (definitely) true. 
\end{compactenum}
From (1) the conclusion is `$a$ is plausible'. 
From (1) and (2), `$a$ is plausible' cannot be deduced, 
but `$\neg a$ is true' can be. 
\end{Eg}

The discussion above justifies our next principle. 

\begin{Prin} \ 
\label{Prin:Evidence, Non-monotonicity}
\raggedright
\begin{compactenum}[\ref{Prin:Evidence, Non-monotonicity}.1)] 
\item \label{Prin:Evidence} 
	\textbf{The Evidence Principle.} \nl
	A plausible proof algorithm can prove a formula $f$ iff 
	all the evidence for $f$ sufficiently outweighs 
	all the evidence against $f$. 
\item \label{Prin:Non-Monotonicity} 
	\textbf{The Non-Monotonicity Principle.} \nl
	A plausible proof algorithm must be non-monotonic. 
\end{compactenum}
\end{Prin}

Exactly what constitutes evidence for or against $f$ can only be determined 
when the particular logic for plausible reasoning is known. 
Also `sufficiently outweighs' depends on the intuition that is being modelled, 
as well as the particular logic. 

A proof algorithm that fails the Evidence Principle 
seems to be seriously flawed. 
So the Evidence Principle may be a principle that 
any sensible proof algorithm should satisfy. 

\subsection{Conjunction}
\label{Subsection:Conjunction}

We shall say a proof algorithm $\alpha$ of a logic $\mathcal{L}$ is 
\defn{conjunctive} iff for any plausible-structure, $\mathcal{S}$, 
and any two formulas $f$ and $g$, if 
\,$\{f,g\} \!\subseteq\! \Thms(\mathcal{L},\alpha,\mathcal{S})$\, then 
\,$\AND \{f,g\} \e \Thms(\mathcal{L},\alpha,\mathcal{S})$. 
For example, the proof algorithm of any classical propositional logic 
is conjunctive. 
A proof algorithm is \defn{non-conjunctive} iff it is not conjunctive. 

Conjunctions of plausible formulas behave very differently from 
conjunctions of formulas that are certain. 
In Example \ref{Eg:3-lottery}, \,$\AND\{\neg s_1, \neg s_2\}$\, 
is equivalent to \,$s_3$. 
So although $\neg s_1$ is plausible and $\neg s_2$ is plausible, 
\,$\AND\{\neg s_1, \neg s_2\}$\, is not plausible. 
Clearly plausible proof algorithms are not conjunctive. 

Although the conjunction of two plausible formulas is not necessarily plausible, 
the conjunction of two facts is a fact. 
So what about the conjunction of a fact and a plausible formula? 
Clearly it cannot be a fact, but is it always plausible? 
Intuitively, a fact $f$ is always true, 
and a plausible formula $g$ is true more often that not. 
So it seems reasonable that their conjunction be true whenever $g$ is true, 
and hence it is reasonable that the conjunction is plausible. 
After we account for explosiveness and the problem of tautologies 
we get the following definition.  
We shall say a proof algorithm $\alpha$ of a logic $\mathcal{L}$ is 
\defn{plausibly conjunctive} iff for any plausible-structure, $\mathcal{S}$, 
and any two formulas $f$ and $g$, if \,$f \e \From(\Fact(\mathcal{S}))$\, and 
\,$g \e \Thms(\mathcal{L},\alpha,\mathcal{S})$\, then 
\,$\AND \{f,g\} \e \Thms(\mathcal{L},\alpha,\mathcal{S})$. 
For example, the proof algorithm of any classical propositional logic 
is plausibly conjunctive. 

\begin{Prin}[Conjunction] \ 
\label{Prin:Conjunction} \raggedright
\begin{compactenum}[\ref{Prin:Conjunction}.1)] 
\item \label{Prin:Non-Conjunction} 
	\textbf{The Non-Conjunction Principle.} \nl
	A plausible proof algorithm must not be conjunctive. 
\item \label{Prin:Plausible Conjunction Principle} 
	\textbf{The Plausible Conjunction Principle.} \nl
	A plausible proof algorithm should be plausibly conjunctive. 
\end{compactenum}
\end{Prin}

The Non-Conjunction Principle is supported by the fact that 
the `And' rule of \cite{KLM:1990}, 
(If \,$a \vsim x$\, and \,$a \vsim y$\, then \,$a \vsim \AND\{x,y\}$.), 
is not probabilistically sound, see \cite{MH:2014}(Section 2.1) 
where they call the `And' rule the `Right$\wedge$+' rule. 
Also Definition 2.4 of \cite{HM:2007} defines an `And' rule that 
is probabilistically sound and has a similar intuition to 
our Plausible Conjunction Principle. 

\subsection{Disjunction}
\label{Subsection:Disjunction}

We shall say a proof algorithm $\alpha$ of a logic $\mathcal{L}$ 
is \defn{disjunctive} iff for any plausible-structure, $\mathcal{S}$, 
and any two formulas $f$ and $g$, if 
\,$\OR \{f,g\} \e \Thms(\mathcal{L},\alpha,\mathcal{S})$\, 
then either \,$f \e \Thms(\mathcal{L},\alpha,\mathcal{S})$\, 
or \,$g \e \Thms(\mathcal{L},\alpha,\mathcal{S})$. 
A proof algorithm is \defn{non-disjunctive} iff it is not disjunctive. 
The proof algorithm of any classical propositional logic 
is non-disjunctive. 

The 3-lottery example (Example \ref{Eg:3-lottery}) shows that, 
although $s_1$ and $s_2$ are both unlikely 
their disjunction $\OR\{s_1, s_2\}$ is likely. 
Hence our next principle is necessary. 

\begin{Prin}[The Non-Disjunction Principle] \ 
\label{Prin:Non-Disjunction} \nl 
A plausible proof algorithm must not be disjunctive. 
\end{Prin}

\subsection{Supraclassicality}
\label{Subsection:Supraclassicality}

Consider a plausible-structure $\mathcal{S}$. 
Let $\alpha$ be a plausible proof algorithm of the logic $\mathcal{L}$. 
Then it is tempting to suggest that 
\,$\Thms(\Fact(\mathcal{S})) \subseteq \Thms(\mathcal{L},\alpha,\mathcal{S})$. 
This is called \defn{supraclassicality}, and could be phrased as 
`what is true is usually true'. 

As we saw in Section~\ref{Section:Set-based formulas} after 
Definition~\ref{Defn:L(f,v),Claus(f)}, 
classical propositional logic is explosive and proves all tautologies. 
But we do not want to force logics for plausible reasoning 
to be explosive or to prove all tautologies. 

\begin{Defn} \label{Defn:plausibly supraclassical} \raggedright
A proof algorithm $\alpha$ of a logic $\mathcal{L}$ has the 
\defn{plausible supraclassicality property} and is said to be 
\defn{plausibly supraclassical} iff for any plausible-structure, $\mathcal{S}$, 
\,$\From(\Fact(\mathcal{S})) \subseteq \Thms(\mathcal{L},\alpha,\mathcal{S})$. 
\end{Defn} 

\begin{Prin}[The Plausible Supraclassicality Principle] \ 
\label{Prin:Plausible Supraclassicality} \nl
Factual and plausible proof algorithms should be plausibly supraclassical. 
\end{Prin}

Since \,$\From(\Fact(\mathcal{S})) \subseteq \Thms(\Fact(\mathcal{S}))$, 
if $\alpha$ is supraclassical (that is, 
\,$\Thms(\Fact(\mathcal{S})) \subseteq \Thms(\mathcal{L},\alpha,\mathcal{S})$) 
then it is plausibly supraclassical. 

\subsection{Right Weakening}
\label{Subsection:Right Weakening}

Right Weakening can be thought of as closure under classical inference. 
More precisely, a proof algorithm $\alpha$ has the 
\defn{right weakening property} iff for any plausible-structure, $\mathcal{S}$, 
and any formula $f$, 
if \,$f \e \Thms(\mathcal{L},\alpha,\mathcal{S})$\, and \,$f \models g$\, 
then \,$g \e \Thms(\mathcal{L},\alpha,\mathcal{S})$. 
By replacing $g$ with any tautology, 
we see that a consequence of the right weakening property is 
\,$\Taut \subseteq \Thms(\mathcal{L},\alpha,\mathcal{S})$. 
But we do not want to force logics for plausible reasoning 
to prove all tautologies. 
We say a proof algorithm $\alpha$ has the \defn{weak right weakening property} 
iff for any plausible-structure, $\mathcal{S}$, 
and any formula $f$, if \,$f \e \Thms(\mathcal{L},\alpha,\mathcal{S})$\, 
then \,$\From(\{f\}) \subseteq \Thms(\mathcal{L},\alpha,\mathcal{S})$. 

However, suppose that whenever the facts of 
the plausible-structure $\mathcal{S}$ and a formula $f$ are true 
then the formula $g$ is also true; 
in symbols \,$\Fact(\mathcal{S}) \!\cup\! \{f\} \models g$. 
Then in the situation defined by $\mathcal{S}$, 
$g$ is true at least as often as $f$. 
So if $f$ is usually true then $g$ should also be usually true. 
We say a proof algorithm $\alpha$ has the \defn{strong right weakening property} 
iff for any plausible-structure, $\mathcal{S}$, and any formula $f$, 
if \,$f \e \Thms(\mathcal{L},\alpha,\mathcal{S})$\, and 
\,$\Fact(\mathcal{S}) \!\cup\! \{f\} \models g$\, 
then \,$g \e \Thms(\mathcal{L},\alpha,\mathcal{S})$. 

Combining the ideas in the preceding two paragraphs 
produces the following definition and corresponding principle. 
A proof algorithm $\alpha$ of a logic $\mathcal{L}$ has the 
\defn{plausible right weakening property} iff 
for any plausible-structure, $\mathcal{S}$, and any formula $f$, 
if \,$f \e \Thms(\mathcal{L},\alpha,\mathcal{S})$\, then 
\,$\From(\Fact(\mathcal{S})\!\cup\!\{f\}) \subseteq 
\Thms(\mathcal{L},\alpha,\mathcal{S})$. 

\begin{Prin}[The Plausible Right Weakening Principle] \ 
\label{Prin:Plausible Right Weakening} \nl 
A plausible proof algorithm should have the plausible right weakening property. 
\end{Prin}

We note that strong right weakening implies 
all the other right weakening properties, 
and weak right weakening is implied by 
all the other right weakening properties. 

\subsection{Consistency}
\label{Subsection:Consistency}

Of the 11 characteristics of plausible reasoning 
given on page 114 of \cite{WTG:2014}, characteristic 8 is `stability'; 
which seems to mean (bottom of page 97 of \cite{WTG:2014}) 
that plausible statements are consistent. 
However, as we shall show, where consistency is concerned 
the number of plausible statements is important. 

We say a proof algorithm $\alpha$ of a logic $\mathcal{L}$ is 
\defn{$n$-consistent} iff for any plausible-structure, $\mathcal{S}$, and 
any set of formulas, $F$, if $\Fact(\mathcal{S})$ is satisfiable, and 
\,$F \!\subseteq\! \Thms(\mathcal{L},\alpha,\mathcal{S})$, and \,$|F| \leq n$\, 
then $F$ is satisfiable. 
Also a proof algorithm $\alpha$ of a logic $\mathcal{L}$ is 
\defn{strongly $n$-consistent} iff 
for any plausible-structure, $\mathcal{S}$, and any set of formulas, $F$, 
if $\Fact(\mathcal{S})$ is satisfiable, and 
\,$F \!\subseteq\! \Thms(\mathcal{L},\alpha,\mathcal{S})$, and \,$|F| \leq n$\, 
then $\Fact(\mathcal{S}) \!\cup\! F$ is satisfiable. 

So if a proof algorithm is strongly $n$-consistent then it is $n$-consistent. 
If $\Fact(\mathcal{S})$ is satisfiable then 
$\Thms(\Fact(\mathcal{S}))$ is satisfiable; 
else $\Thms(\Fact(\mathcal{S}))$ contains all formulas. 

Contradictions are not plausible, 
so plausible proof algorithms must be 1-consistent. 
Hence Principle \ref{Prin:Consistency}.\ref{Prin:1-Consistency} below. 

Suppose $\mathcal{S}$ is a plausible-structure such that 
$\Fact(\mathcal{S})$ is satisfiable. 
If \,$f \e \Thms(\mathcal{L},\alpha,\mathcal{S})$\, 
then in the situation defined by $\mathcal{S}$, 
$f$ is more likely to be true than not. 
Hence we should expect \,$\Fact(\mathcal{S}) \!\cup\! \{f\}$\, 
to be satisfiable. 
That is, strong 1-consistency should hold. 

Now consider strong 2-consistency. 
So suppose $f$ and $g$ are formulas such that 
\,$\{f,g\} \!\subseteq\! \Thms(\mathcal{L},\alpha,\mathcal{S})$. 
By strong 1-consistency, both \,$\Fact(\mathcal{S}) \!\cup\! \{f\}$\, and 
\,$\Fact(\mathcal{S}) \!\cup\! \{g\}$\, should be satisfiable. 
If \,$\Fact(\mathcal{S}) \!\cup\! \{f,g\}$\, is unsatisfiable then 
\,$\Fact(\mathcal{S}) \!\cup\! \{g\} \models \neg f$. 
If $f$ and $g$ are contingent then 
the strong right weakening property is reasonable, and so 
we should expect that \,$\neg f \e \Thms(\mathcal{L},\alpha,\mathcal{S})$. 
Thus we have 
\,$\{f,\neg f\} \!\subseteq\! \Thms(\mathcal{L},\alpha,\mathcal{S})$. 
But, a reasonable property of `likely' is that for any formula $f$, 
at most one of $f$ and $\neg f$ is likely. 
Therefore we should not have 
\,$\{f,\neg f\} \!\subseteq\! \Thms(\mathcal{L},\alpha,\mathcal{S})$. 
This unsatisfactory situation can be avoided 
if \,$\Fact(\mathcal{S}) \!\cup\! \{f,g\}$\, is satisfiable. 
So plausible proof algorithms should be strongly 2-consistent. 
Hence Principle \ref{Prin:Consistency}.\ref{Prin:Strong 2-Consistency} below. 

Consider the 3-lottery example (Example \ref{Eg:3-lottery}) and let 
\,$U = \{\neg s_1, \neg s_2, \OR\{s_1, s_2\}\}$. 
For each $x$ in $U$, $x$ is likely; and $\neg x$ is not likely. 
But $U$ is (classically) unsatisfiable. 
The set $U$ shows the necessity of Principle 
\ref{Prin:Consistency}.\ref{Prin:Non-3-Consistency} below. 

\begin{Prin}[Consistency] \ 
\label{Prin:Consistency} \raggedright
\begin{compactenum}[\ref{Prin:Consistency}.1)] 
\item \label{Prin:1-Consistency} 
	\textbf{The 1-Consistency Principle.} \nl
	A plausible proof algorithm must be 1-consistent. 
\item \label{Prin:Strong 2-Consistency} 
	\textbf{The Strong 2-Consistency Principle.} \nl
	A plausible proof algorithm should be strongly 2-consistent. 
\item \label{Prin:Non-3-Consistency} 
	\textbf{The Non-3-Consistency Principle.} \nl
	A plausible proof algorithm that can prove disjunctions 
	must not be 3-consistent. 
\end{compactenum}
\end{Prin}


\subsection{Multiple Intuitions: Ambiguity}
\label{Subsection:Ambiguity}

With the possible exception of tautologies, 
classical propositional logic captures our intuition about 
what follows from a satisfiable set of facts. 
But there are different well-informed intuitions about 
what follows from a plausible-reasoning situation. 
For example, as early as 1987 (Section 4.1 of \cite{THT:1987}) 
it was recognised that a plausible-reasoning situation 
could elicit different sensible conclusions, 
depending on whether ambiguity was blocked or propagated. 
The essence of Figure 3 in \cite{THT:1987} is our third signpost example. 

\begin{Eg}[The Ambiguity Puzzle] \ 
\label{Eg:Ambiguity}

\begin{compactenum}[1)]
\item There is evidence that $a$ is likely. 
\item There is evidence that $\neg a$ is likely. 
\item There is evidence that $b$ is likely. 
\item If $a$ then $\neg b$ is likely. 
\end{compactenum}
\end{Eg}

What can be concluded about $b$? 
The evidence for $b$ is (3). 
The evidence against $b$ comes from (1) and (4). 
If we knew that $a$ was definitely true then 
the evidence for $b$ and against $b$ would be equal. 
Ignoring (2), $a$ is only likely by (1), 
so the evidence against $b$ is weaker than the evidence for $b$. 
But (2) means that $a$ is even less likely, 
and so the evidence against $b$ has been further weakened. 
Thus $b$ is more likely than $\neg b$. 
Hence many people think that it is reasonable to be able to conclude $b$. 
Such reasoning might be called `best bet' or `most likely' 
or `balance of probabilities' reasoning.

A formula $f$ is said to be \defn{ambiguous} iff 
there is evidence for $f$ and 
there is evidence against $f$ and 
neither $f$ nor $\neg f$ can be proved.
Since (1) and (2) give equal evidence for and against $a$, $a$ is ambiguous. 

If the evidence against $b$ has been weakened sufficiently 
to allow $b$ to be concluded, then $b$ is not ambiguous. 
So the ambiguity of $a$ has been blocked from propagating to $b$. 
An algorithm that can prove $b$ (but not $\neg b$) 
is said to be \defn{ambiguity blocking}. 
This level of reasoning is appropriate if 
the benefit of being right outweighs the penalty for being wrong. 

If the evidence against $b$ has not been weakened sufficiently 
to allow $b$ to be concluded, then $b$ is ambiguous. 
So the ambiguity of $a$ has been propagated to $b$. 
An algorithm that cannot prove $b$ (or $\neg b$) 
is said to be \defn{ambiguity propagating}. 
This more cautious level of reasoning is appropriate if 
the penalty for being wrong outweighs the benefit of being right. 

It is well-known that the Anglo-American legal system uses 
a hierarchy of proof levels, 
two of which are the `balance of probabilities' or 
`preponderance of the evidence' (used in civil cases) 
which is ambiguity blocking, 
and `beyond reasonable doubt' (used in criminal cases) 
which is ambiguity propagating. 
So there is a need for a proof algorithm that blocks ambiguity and 
one that propagates ambiguity. 

To avoid confusion, one should know which algorithm is used; 
unless it is irrelevant to the point being made. 
This, and our observation at the beginning of this section that 
a logic for plausible reasoning should have a factual proof algorithm, 
leads to our next principle. 

\begin{Prin}[The Many Proof Algorithms Principle] \ 
\label{Prin:Many Proof Algorithms} \nl
A logic for plausible reasoning should have at least 
\begin{compactenum}[1)]
\item a factual proof algorithm, 
\item an ambiguity blocking plausible proof algorithm, and 
\item an ambiguity propagating plausible proof algorithm. 
\end{compactenum}
Also, the proof algorithm used to prove a formula should be explicit 
or irrelevant. 
\end{Prin}

Clearly the algorithms in (2) and (3) must be different.  
But, as indicated after Principle \ref{Prin:Representation}, 
the factual algorithm could be the same as a plausible algorithm.  

\subsection{Decisiveness}
\label{Subsection:Decisiveness}

For a formula, $f$, a proof algorithm, $\alpha$, 
will satisfy exactly one of the following conditions. 
\begin{compactenum}[i)]
\item $\alpha$ does not terminate. 
\item $\alpha$ terminates in a state indicating that $f$ is proved, 
\item $\alpha$ terminates in a state indicating that $f$ is not provable, 
\item $\alpha$ terminates in some other state. 
\end{compactenum}
A proof algorithm $\alpha$ is said to be \defn{decisive} iff 
for every formula $f$, $\alpha$ terminates in 
either a state indicating that $f$ is proved, 
or a state indicating that $f$ is not provable. 

Our next principle is clearly desirable. 

\begin{Prin}[The Decisiveness Principle] \ 
\label{Prin:Decisiveness} \nl
Factual and plausible proof algorithms should be decisive. 
\end{Prin}

\subsection{Truth Values}
\label{Subsection:Truth Values}
Let us change our focus from deduction to the more semantic notion 
of assigning truth values to statements. 
For classical propositional logic there are exactly two truth values: 
$\T$ for true and $\F$ for false. 
If $v$ is a valuation 
(that is a function from the set of formulas to the set of truth values) 
and $f$ and $g$ are formulas then \nl
1) Either \,$v(f) = \T$\, or \,$v(\neg f) = \T$\, but not both, 
(the Excluded Middle property) and \nl
2) $v(\AND\{f,g\}) = \T$ \ iff \ $v(f) = \T = v(g)$, and \nl
3) $v(\OR\{f,g\}) = \T$ \ iff \ $v(f) = \T$ or $v(g) = \T$. 

The 3-lottery example (Example \ref{Eg:3-lottery}) shows that 
the closest plausible reasoning can get to (2) and (3) is 
(4) and (5) below. \nl
4) If $v(\AND\{f,g\}) = \T$ then $v(f) = \T = v(g)$. \nl
5) If $v(f) = \T$ or $v(g) = \T$ then $v(\OR\{f,g\}) = \T$. 

Moreover consider our fourth signpost example. 

\begin{Eg}[The 4-lottery example] \ 
\label{Eg:4-lottery}
Consider a 4-lottery. 
Then we have the following. 
\begin{compactenum}[1)] \raggedright
\item Exactly one element of $\{s_1, s_2, s_3, s_4\}$ is true. 
\item Each element of $\{s_1, s_2, s_3, s_4\}$ is probably false. 
\item The disjunction of any 2 of elements of $\{s_1, s_2, s_3, s_4\}$ 
	is not probably true and not probably false. 
\item The disjunction of any 3 elements of $\{s_1, s_2, s_3, s_4\}$ 
	is probably true. 
\end{compactenum}
\end{Eg}

Intuitively some formulas concerning Example \ref{Eg:4-lottery} 
have different truth values; for example 
$\OR\{s_1, s_2, s_3, s_4\}$ is definitely true, 
$\neg \OR\{s_1, s_2, s_3, s_4\}$ is definitely false, 
$\neg s_1$ is probably true, $s_1$ is probably false, and 
$\OR\{s_1, s_2\}$ is as likely to be true as false. 
So plausible reasoning appears to need at least 3 truth values: \nl
one indicating that a formula is more likely to be true than false, \nl
one indicating that a formula is as likely to be true as false, and \nl
one indicating that a formula is more likely to be false than true. 

Hence our last principle of plausible reasoning. 

\begin{Prin}[The Included Middle Principle] \ 
\label{Prin:Included Middle} \nl
A logic for plausible reasoning should have at least 3 truth values. 
\end{Prin}

But what happens if we insist on there being exactly two truth values? 
Suppose a logic $\mathcal{L}$ for plausible reasoning 
has exactly 2 truth values, $\T$ and $\F$. 
Also suppose that for any formulas $f$, $g$, and $h$, 
the following truth conditions hold. 
\begin{compactenum}[TC1)] \raggedright
\item If $f$ is more likely to be true than false \nl
then $f$ and $\neg f$ have different truth values. 
\item If $f$ is as likely to be true as false \nl
then $f$ and $\neg f$ have the same truth value. 
\item The truth value of \,$\OR\{f, g, h\}$\, is $\T$ iff 
the truth value of at least one of $f$, $g$, or $h$ is $\T$. 
\item $\neg \OR\{f, g, h\}$\, and \,$\AND\{\neg f, \neg g, \neg h\}$\, 
have the same truth value. 
\item The truth value of \,$\AND\{f, g, h\}$\, is $\T$ iff 
the truth value of each one of $f$, $g$, and $h$ is $\T$. 
\end{compactenum}

Now apply $\mathcal{L}$ to Example \ref{Eg:4-lottery}. 
By TC1, for each $i$ in $\{1,2,3\}$, 
$s_i$ and $\neg s_i$ have different truth values. 
By TC2, $\OR\{s_1, s_2, s_3\}$\, and \,$\neg \OR\{s_1, s_2, s_3\}$\, 
have the same truth value, 
which by TC4 is the same as \,$\AND\{\neg s_1, \neg s_2, \neg s_3\}$. 

If the truth value of \,$\OR\{s_1, s_2, s_3\}$\, is $\T$ then by TC3, 
for some $i$, the truth value of $s_i$ is $\T$; 
hence the truth value of $\neg s_i$ is $\F$ and so by TC5 
the truth value of \,$\AND\{\neg s_1, \neg s_2, \neg s_3\}$\, is $\F$. 
On the other hand if the truth value of \,$\OR\{s_1, s_2, s_3\}$\, is $\F$ 
then by TC3, for each $i$, the truth value of $s_i$ is $\F$; 
hence the truth value of each $\neg s_i$ is $\T$ and so by TC5 
the truth value of \,$\AND\{\neg s_1, \neg s_2, \neg s_3\}$\, is $\T$. 

So in both cases \,$\OR\{s_1, s_2, s_3\}$\, and 
\,$\AND\{\neg s_1, \neg s_2, \neg s_3\}$\, have different truth values 
which contradicts what we had before. 

The conditions TC1, TC2, TC3, TC4, and TC5 are so closely related to 
the meaning of `true', `false', `conjunction', `disjunction', and `negation', 
that it is hard to reject any of them. 
Therefore it seems that having only two truth values is an over-simplification. 

\subsection{Correctness}
\label{Subsection:Correctness}

A logic that satisfies all the previous principles could 
nonetheless have a fatal flaw.  
It could give an unsatisfactory answer to a particular example.  
Some examples may well have no set of answers that are generally agreed upon.  
But some examples do have a set of answers that are generally agreed upon.  
We might call these answers the correct answers.  
So it is tempting to state a principle of correctness similar to 
``When correct answers exist, a logic must give all the correct answers, 
and no incorrect answers.".  

The problem with such a principle is that it is impossible to show 
that any logic satisfies it.  
The most that can be done is to produce an counter-example that shows 
a logic fails the principle, 
or demonstrate that for a chosen set of examples 
the logic gets the correct answers.  
But there might exist a counter-example that shows 
the logic fails the principle of correctness.  

Thus we shall refrain from trying to formally state a Correctness Principle.  

\section{Some Non-Monotonic Logics} 
\label{Section:Some Non-Monotonic Logics}

We shall consider the relationship between some (non-numeric) non-monotonic 
logics and the principles and examples of Section \ref{Section:Principles}. 

There are three well-known non-monotonic logics, 
namely Default Logic, Circumscription, and Autoepistemic Logic; 
see \cite{Antoniou:1997} for an introduction. 
Answer Set Programming (ASP) \cite{Baral:2003} is a well-known 
Knowledge Representation system. 

Each of the proof algorithms of these four well-known systems is 
conjunctive and so fails the Non-Conjunction Principle 
(Principle \ref{Prin:Conjunction}.\ref{Prin:Non-Conjunction}). 
Also for each of these four proof algorithms, 
the set of all provable formulas is either satisfiable or contains all formulas. 
So all four proof algorithms fail the Non-3-Consistency Principle 
(Principle \ref{Prin:Consistency}.\ref{Prin:Non-3-Consistency}). 
Hence none of these logics reasons correctly about 
the 3-lottery example (Example \ref{Eg:3-lottery}). 
Finally all four of these proof algorithms are ambiguity propagating 
but not ambiguity blocking. 
So each of these logics fails the Many Proof Algorithms Principle 
(Principle \ref{Prin:Many Proof Algorithms}). 
Hence when ambiguity blocking is required --- for instance in civil cases --- 
these logics do not get the right answers. 

Logics that deal with only literals are incapable of the reasoning required by 
Example \ref{Eg:3-lottery}. 
Logics in this category include inheritance networks \cite{HTT:1990}, 
the DeLP system of \cite{GS:2004}, 
the ASPIC system mentioned in \cite{CA:2007}, the logic in \cite{PS:1997}, 
Ordered logic \cite{GVN:1994}, and 
most Defeasible Logics \cite{Billington:2008}. 

Propositional Plausible Logic (PPL), which is defined in the next section, 
is a member of the family of Defeasible Logics. 
The only Defeasible Logics that deal with 
conjunction and disjunction, besides PPL, 
are the logic in \cite{BR:2001}, let's call it DL1, 
and the logic in \cite{Billington:2008}, let's call it DL8. 
But the plausible proof algorithms of both DL1 and DL8 are conjunctive 
and so do not satisfy the Non-Conjunction Principle 
(Principle \ref{Prin:Conjunction}.\ref{Prin:Non-Conjunction}). 
Also the Decisiveness Principle (Principle \ref{Prin:Decisiveness}) fails 
for the plausible proof algorithms that define the Defeasible Logics in: 
\cite{Billington:1993}, \cite{BR:2001}, \cite{MN:2006}, 
\cite{Billington:2008}, and \cite{Billington:2011}. 
Since all Defeasible Logics apart from PPL are closely related to 
a Defeasible Logic in these five citations, 
all Defeasible Logics apart from PPL fail the Decisiveness Principle. 
So PPL is the only Defeasible Logic that satisfies 
all the principles in Section \ref{Section:Principles}. 
Also PPL is more expressive than previous Defeasible Logics because 
the non-strict rules in PPL use formulas 
whereas previous Defeasible Logics only used literals and clauses. 
This is significant because a finite set of clauses is very different to 
the conjunction of those clauses, 
see the 3-lottery example (Example \ref{Eg:3-lottery}). 

Argumentation systems, \cite{Dung:1995}, are well-known 
non-monotonic reasoning systems that can use rules, 
for example ASPIC \cite{CA:2007} and ASPIC$^+$ \cite{MP:2013}. 
Let \,$E \e \{$admissible, complete, preferred, grounded, ideal, semi-stable, stable$\}$. 
Then the semantics of ASPIC$^+$ defined by intersecting all $E$-extensions is 
ambiguity propagating and so fails the 
Many Proof Algorithms Principle (Principle \ref{Prin:Many Proof Algorithms}(2)). 

An early argumentation system is given in \cite{SL:1992} 
and it also is ambiguity propagating and so fails the 
Many Proof Algorithms Principle (Principle \ref{Prin:Many Proof Algorithms}(2)). 
It also has other problems mentioned in \cite{GLV:1998}. 

Three postulates that a rule-based argumentation system should satisfy 
are given in \cite{CA:2007}. 
Postulate 1 is closure under strict rules; 
that is Modus Ponens for strict rules 
(Theorem \ref{Thm:Right Weakening}(\ref{Modus Ponens for strict rules})). 
It is a kind of right weakening property 
(Subsection \ref{Subsection:Right Weakening}). 
Postulate 2 requires the set of all proved literals to be consistent. 
If only literals can be proved, as in \cite{CA:2007}, 
then this is implied by the Strong 2-Consistency Principle 
(Principle \ref{Prin:Strong 2-Consistency}). 
Postulates 1 and 2 jointly imply Postulate 3. 

It is not surprising that Conditional Logics \cite{NC:2001,A-CE:2016} 
have been used to analyse non-monotonic reasoning. 
Let \,$\dashrightarrow$\, denote a weak conditional; 
so that for formulas $f$ and $g$, \,$f \dashrightarrow g$\, means 
`if $f$ then ... $g$' where `...' could be 
`normally', `typically', `probably', or any other similar word or phrase. 
A set of such weak conditionals is called a `conditional knowledge base'. 
The following two rules are particularly important for differentiating 
our plausible reasoning from other kinds of reasoning. 
\begin{compactdesc}
\item[And:] 
	If \,$f \dashrightarrow g$\, and \,$f \dashrightarrow h$\, then 
	\,$f \dashrightarrow \AND\{g,h\}$. 
\item[Or:] 
	If \,$f \dashrightarrow h$\, and \,$g \dashrightarrow h$\, then 
	\,$\OR\{f,g\} \dashrightarrow h$. 
\end{compactdesc}
The And-rule is also called the CC-rule, 
and the Or-rule is also called the CA-rule. 

Let \,$\Ax3$ = $\AND\{\OR\{s_1,s_2,s_3\}$, $\neg \AND\{s_1, s_2\}$, 
$\neg \AND\{s_1, s_3\}$, $\neg \AND\{s_2, s_3\}\}$\, be the formula 
that characterises the 3-lottery example, Example \ref{Eg:3-lottery}(1) 
As noted in Subsection \ref{Subsection:Conjunction}, 
we have \,$\Ax3 \dashrightarrow \neg s_1$\, and 
\,$\Ax3 \dashrightarrow \neg s_2$\, but not 
\,$\Ax3 \dashrightarrow \AND\{\neg s_1,\neg s_2\}$. 
So reasoning systems that satisfy the And-rule do not do plausible reasoning. 

The formula, $\Ax7$, that characterises a 7-lottery is the conjunction of 
the following 22 formulas: \,$\OR\{s_1,s_2,s_3,s_4,s_5,s_6,s_7\}$\, and 
$\neg \AND\{s_i, s_j\}$, where \,$1 \leq i < j \leq 7$. 
Let $f$ be $\AND\{\Ax7,\neg s_1, \neg s_2\}$, 
let $g$ be $\AND\{\Ax7,\neg s_3, \neg s_4\}$, and 
let $h$ be $\OR\{s_5,s_6,s_7\}$. 
Then $f$ is equivalent to exactly one of 
$s_3$ or $s_4$ or $s_5$ or $s_6$ or $s_7$, and 
$g$ is equivalent to exactly one of $s_1$ or $s_2$ or $s_5$ or $s_6$ or $s_7$. 
So \,$f \dashrightarrow h$\, and \,$g \dashrightarrow h$. 
But \,$\OR\{f,g\}$\, does not restrict the selected number at all, 
and $h$ is not the usual result of a 7-lottery. 
So we do not have \,$\OR\{f,g\} \dashrightarrow h$. 
Therefore reasoning systems that satisfy the Or-rule 
do not do plausible reasoning. 

In \cite{Delgrande:2007} it is observed that the following reasoning systems 
satisfy both the And-rule and the Or-rule and 
hence do not do our plausible reasoning: 
systems based on intuitions from probability theory such as 
\cite{Adams:1975} and \cite{Pearl:1988}, and 
from qualitative possibilistic logic \cite{DLP:1994}, 
those based on C4 \cite{Lamarre:1991}, CT4 \cite{Boutilier:1994a}, 
and S \cite{Burgess:1981}. 

Geffner and Pearl \cite{GP:1992} define a logic called `conditional entailment'. 
The second paragraph on page 235 of \cite{GP:1992} 
contains the following sentence. 
``In the propositional case, the only difference between 
conditional entailment and prioritized circumscription is 
the source of the priorities: 
while prioritized circumscription relies on the user, 
conditional entailment extracts the priorities from the knowledge base itself." 
As noted near the beginning of this section, 
circumscription fails the Non-Conjunction Principle 
(Principle \ref{Prin:Conjunction}.\ref{Prin:Non-Conjunction}), 
the Non-3-Consistency Principle (Principle \ref{Prin:Consistency}.\ref{Prin:Non-3-Consistency}), 
and the Many Proof Algorithms Principle 
(Principle \ref{Prin:Many Proof Algorithms}). 
Hence conditional entailment also fails these principles. 

The consequence function of \cite{Makinson:1988} and the 
cumulative conditional knowledge bases of \cite{KLM:1990} 
satisfy both the And-rule and the Or-rule. 
Preferential conditional knowledge bases \cite{KLM:1990} are cumulative. 
Rational conditional knowledge bases \cite{LM:1992} are preferential. 
Hence both the preferential and rational closure of a 
conditional knowledge base satisfies both the And-rule and the Or-rule; 
and so does not do the plausible reasoning we are trying to characterise. 

As noted in \cite{Delgrande:2007} the following systems are 
`essentially the same as' rational closure and 
hence do not do our plausible reasoning: System Z \cite{Pearl:1990}, 
systems based on conditional logic \cite{CL:1992}, 
on modal logic \cite{Boutilier:1994b}, 
on possibilistic logic \cite{BDP:1992}, and 
on conditional objects \cite{DP:1991}. 

The Propositional Typicality Logic (PTL) of 
\cite{BMV:2013} and \cite{BCMV:2015} has several semantics. 
Each semantics is at least preferential and 
so satisfies both the And-rule and the Or-rule. 
Hence PTL does not do the plausible reasoning we are trying to characterise. 

The conditional logic C of \cite{Delgrande:2007} does not satisfy the 
Plausible Right Weakening Principle 
(Principle \ref{Prin:Plausible Right Weakening}). 
Also C and the extensions of C considered in \cite{Delgrande:2007} 
have only one proof algorithm and so fail the Many Proof Algorithms Principle 
(Principle \ref{Prin:Many Proof Algorithms}). 

Apart from the problems mentioned in Section 5 of \cite{GP:1991}, 
System Z \cite{Pearl:1990} and System Z$^+$ \cite{GP:1991} 
are ambiguity propagating but not ambiguity blocking. 
Hence they fail the Many Proof Algorithms Principle 
(Principle \ref{Prin:Many Proof Algorithms}). 
Moreover, although they can represent the 
3-lottery example (Example \ref{Eg:3-lottery}), 
they cannot prove anything about the example because 
the set of rules is not `consistent' as defined in \cite{Pearl:1990,GP:1991}. 

The logic implemented by \textsc{theorist} \cite{Poole:1988} and 
the Preferred Subtheories logic in \cite{Brewka:1989} 
both generate consistent extensions and so fail the Non-3-Consistency Principle 
(Principle \ref{Prin:Consistency}.\ref{Prin:Non-3-Consistency}). 

Every logic reviewed above fails at least one of the following principles: 
the Non-Conjunction Principle 
(Principle \ref{Prin:Conjunction}.\ref{Prin:Non-Conjunction}), 
the Non-3-Consistency Principle 
(Principle \ref{Prin:Consistency}.\ref{Prin:Non-3-Consistency}), 
the Many Proof Algorithms Principle 
(Principle \ref{Prin:Many Proof Algorithms}), and 
the correctness principle as instanced by the 3-lottery example 
(Example \ref{Eg:3-lottery}). 
So these principles seem to be central to the difference between 
the plausible reasoning characterised in Section \ref{Section:Principles} 
and other kinds of non-numeric non-monotonic reasoning. 
As far as we know, Propositional Plausible Logic is the only non-numeric 
non-monotonic propositionally adequate logic that 
satisfies all the principles in Section \ref{Section:Principles}. 

\section{Propositional Plausible Logic (PPL)} 
\label{Section:PPL}

The purpose of this section is to define a propositional logic, 
called Propositional Plausible Logic (PPL), 
that satisfies all the principles in Section \ref{Section:Principles}. 
The plausible-structure used in PPL is defined in 
Subsection \ref{Subsection:Plausible Descriptions}. 
The proof algorithms are defined in 
Subsection \ref{Subsection:Proof Relation and Proof Algorithms}. 
The notions of `proof' and `truth' are developed in 
Subsection \ref{Subsection:Proof Theory} and 
Subsection \ref{Subsection:Truth Theory}, respectively. 

As well as the notation introduced in Section \ref{Section:Set-based formulas}, 
we shall use the following notation concerning sequences.  
The empty sequence is denoted by $()$. 
Let $S$ be a sequence. 
If $S$ is finite then $S$+$e$ denotes the sequence 
formed by just adding $e$ onto the right end of $S$. 
Define \,$e \e S$\, to mean $e$ is an element of $S$, 
and \,$e \nte S$\, to mean $e$ is not an element of $S$. 

\subsection{Plausible Descriptions}
\label{Subsection:Plausible Descriptions}

Propositional Plausible Logic (PPL) reasons about 
plausible-reasoning situations that may contain facts, 
like definitions and membership of categories.  
These facts are represented by formulas that 
are converted into clauses called axioms and 
these axioms are then converted into strict rules. 
The plausible information is represented by defeasible rules, warning rules, 
and a priority relation, $>$, on rules. 

Intuitively the various kinds of rules have the following meanings. 
The strict rule \,$A \!\strArr\! c$\, means 
if every formula in $A$ is true then $c$ is true. 
So strict rules are like material implication except that 
$A$ is a finite set of formulas rather than a single formula. 
(We have already seen that $A$ and $\AND A$ behave differently.)
For example, `nautiluses are cephalopods' could be written as 
\,$\{n\} \!\strArr\! c$, and `cephalopods are molluscs' could be written as 
\,$\{c\} \!\strArr\! m$. 

Roughly, the defeasible rule \,$A \!\defArr\! c$\, means 
if every formula in $A$ is true then $c$ is usually true. 
For example, `molluscs usually have shells' could be written as 
\,$\{m\} \!\defArr\! s$, and `cephalopods usually have no shells' could be 
written as \,$\{c\} \!\defArr\! \neg s$. 

The warning rule \,$A \!\warnArr\! c$\, roughly means 
if every formula in $A$ is true then $c$ might be true. 
So \,$A \!\warnArr\! \neg c$\, warns against concluding usually $c$, 
but does not support usually $\neg c$. 
For example, `objects that look red in red light might not be red' could be 
written as \,$\{$looks-red-in-red-light$\} \!\warnArr\! \neg r$. 
Warning rules can be used to prevent unwanted chaining. 
For example, suppose we have `if $a$ then usually $b$' ($\{a\} \!\defArr\! b$) 
and `if $b$ then usually $c$' ($\{b\} \!\defArr\! c$). 
Then it may be too risky to conclude `usually $c$' from $a$. 
Without introducing evidence for $\neg c$, 
the conclusion of `usually $c$' from $a$ can be prevented by 
the warning rule \,$\{a\} \!\warnArr\! \neg c$. 
An instance of this example can be created by letting 
$a$ be \,$x \e \{1,2,3,4\}$, $b$ be \,$x \e \{2,3,4\}$, and 
$c$ be \,$x \e \{3,4,5\}$. 
Warning rules have also been called `defeaters' and `interfering rules'. 
The formal definition of a rule and its associated terms follows. 

\begin{Defn} \label{Defn:rule} \raggedright
A \defn{rule}, $r$, is any triple \,$(A(r), \arrow(r), c(r))$\, such that 
$A(r)$, called the \defn{set of antecedents} of $r$, 
is a finite (possibly empty) set of formulas; 
\,$\arrow(r) \e \{\strArr, \defArr, \warnArr \}$; and 
$c(r)$, called the \defn{consequent} of $r$, depends on $\arrow(r)$. \nl 
If $\arrow(r)$ is the \defn{strict arrow}, $\strArr$, then $c(r)$ is either 
a formula or the conjunction of a countable set of formulas, 
and $r$ is written \,$A(r) \strArr c(r)$\, and called a \defn{strict rule}. \nl
If $\arrow(r)$ is the \defn{defeasible arrow}, $\defArr$, 
then $c(r)$ is a formula, and $r$ is written \,$A(r) \defArr c(r)$\, 
and called a \defn{defeasible rule}. \nl
If $\arrow(r)$ is the \defn{warning arrow}, $\warnArr$, 
then $c(r)$ is a formula, and $r$ is written \,$A(r) \warnArr c(r)$\, 
and called a \defn{warning rule}. 
\end{Defn} 

A priority relation, $>$, on rules is used to indicate 
the more relevant of two rules. 
For instance, the specific rule `cephalopods usually have no shells', 
($\{c\} \!\defArr\! \neg s$), 
is more relevant than the general rule `molluscs usually have shells', 
($\{m\} \!\defArr\! s$), 
when reasoning about the external appearance of cephalopods. 
Hence \,$\{c\} \!\defArr\! \neg s$\, $>$ \,$\{m\} \!\defArr\! s$. 
More generally, some common policies for defining $>$ are the following. 
Prefer specific rules over general rules; 
prefer authoritative rules, (for instance national laws override state laws); 
prefer recent rules (because they are more up-to-date); and 
prefer more reliable rules. 
If $r$ and $s$ are rules and \,$r > s$\, then we often say 
$r$ is \textit{superior} to $s$ and $s$ is \textit{inferior} to $r$. 

Although the priority relation does not have to be transitive, 
it does have to be acyclic.  

\begin{Defn} \label{Defn:priority relation} \raggedright
Let $R$ be any set of rules. 
A binary relation, $>$, on $R$ is \defn{cyclic} iff there exists a 
finite sequence, $(r_1, r_2, ..., r_n)$ where \,$n \geq 1$, of 
elements of $R$ such that \ $r_1 > r_2 >$ ... $> r_n > r_1$; 
that is, \,$r_n > r_1$\, and for all $i$ in $[1\,..\,n\!-\!1]$, 
\,$r_i > r_{i+1}$. 
A binary relation, $>$, is \defn{acyclic} iff it is not cyclic. 
\end{Defn} 

Let us now consider the conversion of the facts of a plausible-reasoning 
situation represented by a set $F$ of formulas into strict rules. 
First we form $\Claus(F)$ which is the set of clauses formed from $F$. 
Next we generate the set of axioms, $\Ax$ by defining 
\,$\Ax = \CorRes(\Sat(\Claus(F)))$. 
Finally we convert a contingent clause with $n$ literals into 
\,$2^n \!-\! 1$\, strict rules. 
The conversion is done by the function $\Rul(.)$ in the usual way 
as shown by the following example. 
\,$\Rul(\OR\{a,b,c\})$ = $\{\ \ 
\{\} \!\strArr\! \OR\{a,b,c\}$, \nl 
$\{\AND\{\neg b, \neg c\}\} \!\strArr\! a$, \quad
$\{\AND\{\neg a, \neg c\}\} \!\strArr\! b$, \quad
$\{\AND\{\neg a, \neg b\}\} \!\strArr\! c$, \nl
$\{\neg a\} \!\strArr\! \OR \{b,c\}$, \quad
$\{\neg b\} \!\strArr\! \OR \{a,c\}$, \quad
$\{\neg c\} \!\strArr\! \OR \{a,b\} \ \ \}$. 

The full definition of $\Rul(.)$ and $\Rul(.,.)$, 
as well as some useful notation, is given in the next definition.

\begin{Defn} \label{Defn:Sets of formulas and rules} \raggedright
Let $R$ be a set of rules, $F$ be a finite set of formulas, and 
$C$ be a set of contingent clauses. 
\begin{compactenum}[1)]
\item \label{Rs} 
	$R_s$ is the set of strict rules in $R$. That is, 
	$R_s = \{r \e R : r$ is a strict rule$\}$. 
\item \label{Rd} 
	$R_d$ is the set of defeasible rules in $R$. That is, 
	$R_d = \{r \e R : r$ is a defeasible rule$\}$. 
\item \label{c(R)} 
	$c(R)$ is the set of consequents of the rules in $R$. That is, 
	$c(R) = \{c(r) : r \e R\}$. 
\item \label{Rul(c)} 
	If \,$c \e C$\, then \,$\Rul(c) = \{\,\{\} \!\strArr\! c\} \cup 
	\{\,\{\AND\nnot(L\!-\!K)\} \strArr \OR K : 
	c = \OR L$\, and \,$\{\} \!\subset\! K \!\subset\! L\}$. 
\item \label{Rul(C)} 
	$\Rul(C) = \bigcup\{\Rul(c) : c \e C\}$. 
\item \label{Rul(C,F)} 
	$\Rul(C,F)$ is the set of rules in $\Rul(C)$ whose 
	set of antecedents is $F$. That is, 
	$\Rul(C,F) = \{r \e \Rul(C) : A(r) = F\}$. 
\end{compactenum}
\end{Defn} 

We note that the set of antecedents of any strict rule formed by $\Rul(.)$ 
has at most one element. 

Although $\Rul(\Ax)$ gives us the strict rules that 
characterise the set $F$ of facts we started with, 
we can reduce the number of these strict rules by `anding' 
all those that have the same antecedent. 
For example, the `anding' of \,$\{a\} \!\strArr\! c_1$, 
\,$\{a\} \!\strArr\! c_2$, and \,$\{a\} \!\strArr\! c_3$\, 
is \,$\{a\} \!\strArr\! \AND\{c_1,c_2,c_3\}$. 
We now have the set of strict rules that we want. 
This set is formally defined by PD\ref{Rs=} below. 
The formal structure used for describing plausible-reasoning situations 
is called a plausible description and is defined below. 

\begin{Defn} \label{Defn:PlausibleDescription} \raggedright
If $R$ is a set of rules then $(R,>)$ is a \defn{plausible description} iff 
PD\ref{Ax=}, PD\ref{Rs=}, PD\ref{rse}, and PD\ref{PDpriority} all hold. 
\begin{compactenum}[PD1)]
\item \label{Ax=} 
	There is a set $F$ of formulas such that 
	\,$\Ax(R) = \CorRes(\Sat(\Claus(F)))$. \nl
	$\Ax(R)$ is called the set of \defn{axioms} of $R$ and 
	is usually denoted by $\Ax$.
\item \label{Rs=} 
	$R_s = \{A \strArr \smp(\AND c(\Rul(\Ax,A))) : 
											A \e \{A(r) : r \e \Rul(\Ax)\}\}$. 
\item \label{rse} 
	If \,$\Ax \!\neq\! \{\}$\, then $\rse$ denotes the strict rule 
	\,$\{\} \strArr \AND \Ax$. 
\item \label{PDpriority} 
	$>$ is a \defn{priority relation} on $R$; that is, 
	\,$> \ \subseteq R \!\times\! (R\!-\!\{\rse\})$\, and $>$ is not cyclic. 
\end{compactenum}
\end{Defn} 

Suppose $(R,>)$ is a plausible description. 
Then $\Ax$ is empty iff $R_s$ is empty. 
If \,$\Ax \!\neq\! \{\}$\, then \,$\rse \e R_s$. 
If $R_s$ is not empty we can extract $\Ax$ from the consequent of $\rse$. 
This shows that $\Ax(R)$ is indeed dependent on $R$. 
Different strict rules in $R$ have different sets of antecedents, 
and no rule is superior ($>$) to $\rse$.

For PPL the plausible-structure is a plausible description $(R,>)$ 
and the factual part is $\Ax(R)$, 
which by PD\ref{Rs=} is equivalent to the strict rules in $R$, $R_s$. 
The plausible part consists of the non-strict rules in $R$ and 
the priority relation $>$. 

By Lemma~\ref{Lem:Err,Sat}(\ref{Sat(C) is satisfiable.}) (See Appendix A) 
and PD\ref{Ax=}, $\Ax$ is satisfiable. 
So in PPL we extend the meaning of `fact' from just being an element of $\Ax$ 
to a formula that is implied by $\Ax$. 
Explicitly, a formula $f$ is said to be a \defn{fact} iff \,$\Ax \!\models\! f$. 

\subsection{The Proof Relation and the Proof Algorithms}
\label{Subsection:Proof Relation and Proof Algorithms}

In this subsection we define what it means for a formula to be proved 
from a plausible description. 
We shall do this by defining a proof relation, $\vminus$, 
and various proof algorithms. 
This complex task will be done by giving the overall strategy, 
and then progressively refining this general plan 
until all the terms used have been defined. 

Any method of demonstrating that \,$\Ax \!\models\! f$\, 
will do as an algorithm for proving facts; 
so there is no need to specify a particular one. 
Let our top level general plan for proving a formula be the following. 
\vspace{-1 ex}
\begin{center} 
Distinguish between proving facts and proving formulas that are not facts. 
\end{center}
\vspace{-1 ex}

Lower case Greek letters will be used to denote the proof algorithms that 
will eventually be defined. 
A general proof algorithm will be denoted by $\alpha$ 
(a for alpha and algorithm). 
We shall use $\varphi$ (f for phi and fact) to denote our 
factual proof algorithm. 
Until a further refinement is needed we shall use the notation 
\,$\alpha \!\vminus\! f$\, to denote that 
a formula $f$ is proved by the proof algorithm $\alpha$. 

Since facts are always true they are (at least) probably true. 
So we shall decree that facts are provable by all proof algorithms. 
Thus we have the following. \nl
All algorithms prove all facts. 
In symbols, if \,$\Ax \models f$\, then \,$\alpha \vminus f$. \nl 
The factual algorithm proves a formula iff it is a fact. 
In symbols, \,$\varphi \vminus f$\, iff \,$\Ax \models f$. 

Now consider formulas $f$ that are not facts, that is, \,$\Ax \not\models f$. 
To (plausibly) prove $f$ we need to do two things. 
First, establish some evidence for $f$. 
Second, defeat all the evidence against $f$. 
This will satisfy the requirements of the Evidence Principle, 
Principle \ref{Prin:Evidence, Non-monotonicity}.\ref{Prin:Evidence}. 
So our first refinement of the general plan is the following. 

\begin{Refine} \label{Refine:5.1} \raggedright
Suppose \,$(R,>)$\, is a plausible description, \,$\Ax = \Ax(R)$, 
and $f$ is a formula. 
\begin{compactenum}[1)]
\item \label{refine1:fact} 
If \,$\Ax \models f$\, then \,$\alpha \vminus f$. \ 
Also \,$\varphi \vminus f$\, iff \,$\Ax \models f$. 

\item \label{refine1:fml} 
If \,$\Ax \not\models f$\, and \,$\alpha \neq \varphi$\, 
then \,$\alpha \vminus f$\, iff 
(\ref{refine1:fml}.\ref{refine1:For}) and 
(\ref{refine1:fml}.\ref{refine1:Against}) hold. 
	\begin{compactenum}[\ref{refine1:fml}.1)]
	\item \label{refine1:For} 
	Establish some evidence for $f$. 

	\item \label{refine1:Against} 
	Defeat all the evidence against $f$. 
	\end{compactenum}

\end{compactenum}
\end{Refine} 

In accordance with the intuitive meaning of the three kinds of rules 
given at the beginning of this subsection, 
the evidence for $f$ consists of strict or defeasible rules that 
have a consequent that implies $f$. 
However since the axioms are always true, we can weaken this to 
requiring that \,$\Ax\!\cup\!\{c(r)\}$\, implies $f$, 
provided that \,$\Ax\!\cup\!\{c(r)\}$\, is satisfiable. 
So if \,$R' \!\subseteq\! R$\, it will be convenient to let \nl
\hspace*{2 em}
$R'[f] = \{r \e R' : \Ax\!\cup\!\{c(r)\}$ is satisfiable and 
			$\Ax\!\cup\!\{c(r)\} \models f\}$. \nl
In the case of Refinement \ref{Refine:5.1}(\ref{refine1:fml}.\ref{refine1:For}) 
we have \,$\Ax \not\models f$\, and so $\rse$ cannot support $f$. 
Hence the following notation is convenient. \nl
\hspace*{2 em}
$R^s_d = (R_s \!\cup\! R_d) - \{\rse\}$. \nl
So the evidence for $f$ is all the rules in $R$ that support $f$, 
that is $R^s_d[f]$. 
To establish some evidence for $f$ we must prove 
the set of antecedents of a rule supporting $f$. 
So we need to find a rule $r$ in $R^s_d[f]$ and prove $A(r)$. 

But $A(r)$ is a set of formulas, not a formula. 
By proving a finite set $F$ of formulas we shall mean 
proving every formula in $F$. 
In symbols, \,$\alpha \vminus F$ \ iff \ $\forall f \e F$, \,$\alpha \vminus f$. 
So if $F$ is empty we have \,$\alpha \vminus \{\}$. 

Collecting these ideas together gives our next refinement. 

\begin{Refine} \label{Refine:5.2} \raggedright
Suppose \,$(R,>)$\, is a plausible description, \,$\Ax = \Ax(R)$, 
and $f$ is a formula. 
\begin{compactenum}[1)]
\item \label{refine2:set} 
If $F$ is a finite set of formulas then 
\,$\alpha \vminus F$ \ iff \ $\forall f \e F$, \,$\alpha \vminus f$. 

\item \label{refine2:fact} 
If \,$\Ax \models f$\, then \,$\alpha \vminus f$. \ 
Also \,$\varphi \vminus f$\, iff \,$\Ax \models f$. 

\item \label{refine2:fml} 
If \,$\Ax \not\models f$\, and \,$\alpha \neq \varphi$\, 
then \,$\alpha \vminus f$\, iff \,$\exists r \e R^s_d[f]$\, such that 
(\ref{refine2:fml}.\ref{refine2:For}) and 
(\ref{refine2:fml}.\ref{refine2:Against}) hold. 
	\begin{compactenum}[\ref{refine2:fml}.1)]
	\item \label{refine2:For} 
	$\alpha \vminus A(r)$. 

	\item \label{refine2:Against} 
	Defeat all the evidence against $f$. 
	\end{compactenum}

\end{compactenum}
\end{Refine} 

Each rule whose consequent implies $\neg f$ is evidence against $f$. 
The set of such rules is $R[\neg f]$. 
In Refinement \ref{Refine:5.2}(\ref{refine2:fml}) we have 
a rule $r$ in $R^s_d[f]$. 
So any rule in $R[\neg f]$ that is inferior to $r$ has already been defeated 
by $r$ and hence need not be explicitly considered. 
This reduces the set of evidence against $f$ that must be considered to 
the set of rules in $R[\neg f]$ that are not inferior to $r$; in symbols, 
\,$\{s \e R[\neg f] : s \not< r\}$. 

A rule $s$ in \,$\{s \e R[\neg f] : s \not< r\}$\, is defeated 
either by team defeat or by disabling $s$. 
The team of rules for $f$ is $R^s_d[f]$. 
The rule $s$ is defeated by \textit{team defeat} iff 
there is a rule $t$ in the team of rules for $f$, $R^s_d[f]$, such that 
$t$ is superior to $s$, \,$t > s$, and 
the set of antecedents of $t$, $A(t)$, is proved \,$\alpha \!\vminus\! A(t)$. 
So if \,$R' \!\subseteq\! R$\, it will be convenient to let $R'[f;s]$ 
denote the set of all rules in $R'[f]$ that are superior to $s$. 
In symbols, \nl
\hspace*{2 em}
$R'[f;s] = \{t \e R'[f] : t > s\}$. \nl
Alternatively $s$ is \textit{disabled} iff the set of antecedents of $s$, 
$A(s)$, cannot be proved, \,$\alpha \not\vminus A(s)$. 

Three notations for useful sets of rules have been introduced, 
so their formal definition is appropriate before our third refinement. 

\begin{Defn} \label{Defn:subsets of R} \raggedright
Suppose $(R,>)$ is a plausible description, \,$\Ax = \Ax(R)$, 
\,$R' \!\subseteq\! R$, $f$ is a formula, and \,$s \e R$. 
\begin{compactenum}[1)]
\item \label{R^s_d =} 
	$R^s_d = (R_s \!\cup\! R_d) - \{\rse\}$. 
\item \label{R'[f]} 
	$R'[f] = \{r \e R' : \Ax\!\cup\!\{c(r)\}$ is satisfiable and 
						$\Ax\!\cup\!\{c(r)\} \models f\}$. 
\item \label{R'[f;s]} 
	$R'[f;s] = \{t \e R'[f] : t > s\}$. 
\end{compactenum}
\end{Defn} 

\begin{Refine} \label{Refine:5.3} \raggedright
Suppose \,$(R,>)$\, is a plausible description, \,$\Ax = \Ax(R)$, 
and $f$ is a formula. 
\begin{compactenum}[1)]
\item \label{refine3:set} 
If $F$ is a finite set of formulas then 
\,$\alpha \vminus F$ \ iff \ $\forall f \e F$, \,$\alpha \vminus f$. 

\item \label{refine3:fact} 
If \,$\Ax \models f$\, then \,$\alpha \vminus f$. \ 
Also \,$\varphi \vminus f$\, iff \,$\Ax \models f$. 

\item \label{refine3:fml} 
If \,$\Ax \not\models f$\, and \,$\alpha \neq \varphi$\, 
then \,$\alpha \vminus f$\, iff \,$\exists r \e R^s_d[f]$\, such that 
(\ref{refine3:fml}.\ref{refine3:For}) and 
(\ref{refine3:fml}.\ref{refine3:Against}) hold. 
	\begin{compactenum}[\ref{refine3:fml}.1)]
	\item \label{refine3:For} 
	$\alpha \vminus A(r)$. 

	\item \label{refine3:Against} 
	$\forall s \e \{s \e R[\neg f] : s \not< r\}$ \ either 
		\begin{compactenum}[\ref{refine3:fml}.\ref{refine3:Against}.1)]
		\item \label{refine3:TeamDefeat} 
		$\exists t \e R^s_d[f;s]$\, such that \,$\alpha \vminus A(t)$; \ or 

		\item \label{refine3:DisableByNoProof} 
		$\alpha \not\vminus A(s)$. 
		\end{compactenum}
	\end{compactenum}
\end{compactenum}
\end{Refine} 

Refinement \ref{Refine:5.3} has no undefined terms, 
but unfortunately it has two failings. 
There is only one plausible proof algorithm (denoted by $\alpha$), 
and so the Many Proof Algorithms Principle, 
Principle \ref{Prin:Many Proof Algorithms}, fails.
Also, a proof may get into a loop, and 
hence the Decisiveness Principle, Principle \ref{Prin:Decisiveness}, fails. 

Before we consider looping, let us invent the other proof algorithms. 
The $\alpha$ in Refinement \ref{Refine:5.3}(\ref{refine3:fml}.\ref{refine3:Against}.\ref{refine3:DisableByNoProof})
evaluates evidence against $f$; and this need not be the same $\alpha$ as in 
(\ref{refine3:fml}.\ref{refine3:For}) and 
(\ref{refine3:fml}.\ref{refine3:Against}.\ref{refine3:TeamDefeat}) 
which evaluates evidence for $f$. 
To avoid confusion let us call the $\alpha$ in 
(\ref{refine3:fml}.\ref{refine3:Against}.\ref{refine3:DisableByNoProof}), 
$\alpha'$. 
Replacing $\alpha$ by $\alpha'$ in 
(\ref{refine3:fml}.\ref{refine3:Against}.\ref{refine3:DisableByNoProof}) 
creates the need to decide what $(\alpha')'$ is. 
Let us simplify $(\alpha')'$ to $\alpha''$. 
Some obvious choices are: \,$\alpha'' = \alpha$, or \,$\alpha'' = \alpha'$, 
or $\alpha''$ is some other proof algorithm. 
The third choice postpones and complicates the choice that 
must eventually be made. 
Experimentation shows that the second choice has some properties 
that we would rather avoid. 
So we let \,$\alpha'' = \alpha$. 

Another change that can be made is to the set 
\,$\{s \e R[\neg f] : s \not< r\}$\, of rules that 
a proof algorithm regards as evidence against $f$. 
Let $\Foe(\alpha,f,r)$ denote the set of rules that $\alpha$ regards 
as the evidence against $f$ that is not inferior to $r$. 

These ideas gives us our fourth refinement. 

\begin{Refine} \label{Refine:5.4} \raggedright
Suppose \,$(R,>)$\, is a plausible description, \,$\Ax = \Ax(R)$, 
and $f$ is a formula. 
\begin{compactenum}[1)]
\item \label{refine4:set} 
If $F$ is a finite set of formulas then 
\,$\alpha \vminus F$ \ iff \ $\forall f \e F$, \,$\alpha \vminus f$. 

\item \label{refine4:fact} 
If \,$\Ax \models f$\, then \,$\alpha \vminus f$. \ 
Also \,$\varphi \vminus f$\, iff \,$\Ax \models f$. 

\item \label{refine4:fml} 
If \,$\Ax \not\models f$\, and \,$\alpha \neq \varphi$\, 
then \,$\alpha \vminus f$\, iff \,$\exists r \e R^s_d[f]$\, such that 
(\ref{refine4:fml}.\ref{refine4:For}) and 
(\ref{refine4:fml}.\ref{refine4:Against}) hold. 
	\begin{compactenum}[\ref{refine4:fml}.1)]
	\item \label{refine4:For} 
	$\alpha \vminus A(r)$. 

	\item \label{refine4:Against} 
	$\forall s \e \Foe(\alpha,f,r)$ \ either 
		\begin{compactenum}[\ref{refine4:fml}.\ref{refine4:Against}.1)]
		\item \label{refine4:TeamDefeat} 
		$\exists t \e R^s_d[f;s]$\, such that \,$\alpha \vminus A(t)$; \ or 

		\item \label{refine4:DisableByNoProof} 
		$\alpha' \not\vminus A(s)$. 
		\end{compactenum}
	\end{compactenum}
\end{compactenum}
\end{Refine} 

Let us create our first non-factual proof algorithm, $\beta$, 
by changing Refinement \ref{Refine:5.4} as little as possible. 
So let \,$\Foe(\beta,f,r) = \{s \e R[\neg f] : s \not< r\}$. 
Let $\beta$ be defined by replacing each $\alpha$ in 
Refinement \ref{Refine:5.4} with $\beta$. 
Of course now $\beta'$ must be defined. 
First let \,$\Foe(\beta',f,r) = \{s \e R[\neg f] : s \not< r\}$. 
Then let $\beta'$ be defined by replacing each $\alpha$ in 
Refinement \ref{Refine:5.4} with $\beta'$. 
(Recall that \,$\beta'' = \beta$.) 
Later we shall show that $\beta$ is ambiguity blocking 
(b for beta and blocking). 
We are not really concerned with any primed algorithm as they only assist 
with the definition of their non-primed co-algorithm. 
But later we shall show that $\beta$ and $\beta'$ prove 
exactly the same formulas. 
So why is $\beta'$ needed? 
Without $\beta'$ it is exceedingly difficult to prove the relationship 
between $\beta$ and the other algorithms we are about to define. 

Our next algorithm, $\pi$, will be shown to be ambiguity propagating 
(p for pi and propagating). 
We want to make $\pi$ as strong as possible; that is, 
$\pi$ proves $f$ if there is no evidence against $f$. 
This can be done by making its co-algorithm $\pi'$ as weak as possible; 
that is, $\pi'$ ignores all evidence against $f$; 
hence \,$\Foe(\pi',f,r) = \{\}$. 
This is the only change we make to Refinement \ref{Refine:5.4}. 
Explicitly, let \,$\Foe(\pi,f,r) = \{s \e R[\neg f] : s \not< r\}$. 
Let $\pi$ be defined by replacing each $\alpha$ in 
Refinement \ref{Refine:5.4} with $\pi$. 
Let $\pi'$ be defined by replacing each $\alpha$ in 
Refinement \ref{Refine:5.4} with $\pi'$. 
(Recall that \,$\pi'' = \pi$.) 

Our last algorithm, $\psi$, will also be shown to be ambiguity propagating 
(p for psi and propagating). 
We want to make $\psi$ weaker than $\pi$. 
This can be done by making its co-algorithm $\psi'$ regard those rules 
that imply $\neg f$ and are superior to $r$ as evidence against $f$; 
hence $\Foe(\psi',f,r) = \{s \e R[\neg f] : s > r\}$. 
This is the only change we make to Refinement \ref{Refine:5.4}. 
Explicitly, let \,$\Foe(\psi,f,r) = \{s \e R[\neg f] : s \not< r\}$. 
Let $\psi$ be defined by replacing each $\alpha$ in 
Refinement \ref{Refine:5.4} with $\psi$. 
Let $\psi'$ be defined by replacing each $\alpha$ in 
Refinement \ref{Refine:5.4} with $\psi'$. 
(Recall that \,$\psi'' = \psi$.) 

To emphasise that we are only interested in the non-primed proof algorithms 
we note that there are examples in which both $\pi'$ and $\psi'$ can 
prove both $f$ and $\neg f$. 
This is fine as both $\pi'$ and $\psi'$ only assess the evidence against $f$, 
rather than try to defeasibly justify accepting $f$, 
as the non-primed algorithms do. 

The following two formal definitions collect together for easy reference 
the above notations concerning algorithms and $\Foe(.,.,.)$. 

\begin{Defn} \label{Defn:Alg,co-algorithms} \raggedright
Define the set, $\Alg$, of \defn{names of the proof algorithms} by 
\,$\Alg = \{\varphi,\pi,\psi,\beta,\beta',\psi',\pi'\}$. 
Define \,$\varphi' = \varphi$. 
If \,$\alpha \e \{\pi,\psi,\beta\}$\, then 
define \,$(\alpha')' = \alpha'' = \alpha$. 
If \,$\alpha \e \Alg$\, then the \defn{co-algorithm} of $\alpha$ is $\alpha'$. 
\end{Defn} 

\begin{Defn} \label{Defn:Foe(.,.,.)} \raggedright
Suppose $(R,>)$ is a plausible description, $f$ is a formula, and \,$r \e R$. 
\begin{compactenum}[1)]
\item 
	If \,$\alpha \e \{\pi,\psi,\beta,\beta'\}$\, and \,$r \neq \rse$\, 
	then \,$\Foe(\alpha,f,r) = \{s \e R[\neg f] : s \not< r\}$. 
\item 
	If \,$\alpha \e \{\varphi,\pi'\}$\, or \,$r = \rse$\, 
	then \,$\Foe(\alpha,f,r) = \{\}$. 
\item 
	$\Foe(\psi',f,r) = \{s \e R[\neg f] : s > r\} = R[\neg f; r]$. 
\end{compactenum}
\end{Defn} 

Finally, let us consider looping. 
To prove $f$ we use $\alpha$ and a rule $r$. 
While proving $f$ we may have to prove other formulas. 
During a proof of one of these other formulas, 
if we choose to use $\alpha$ and $r$ again then 
we will be in a loop and so this choice should fail. 
To prevent such a looping choice we need to 
record that $\alpha$ and $r$ have been used previously. 
We shall call such a record of used algorithms and rules a history. 
Its formal definition follows. 

\begin{Defn} \label{Defn:history} \raggedright
Suppose $(R,>)$ is a plausible description and \,$\alpha \e \Alg$.
Define \,$\alpha R = \{\alpha r : r \e R\}$. 
Then $H$ is an \defn{$\alpha$-history} iff $H$ is a finite sequence 
of elements of \,$\alpha R \cup \alpha' R$\, that has no repeated elements. 
\end{Defn} 

Unfortunately using a history complicates Refinement \ref{Refine:5.4} 
because we now no longer have just an algorithm proving a formula, 
but an algorithm and a history proving a formula. 
Therefore in (\ref{refine4:set}), (\ref{refine4:fact}), and (\ref{refine4:fml}), 
\,$\alpha \vminus x$\, becomes \,$(\alpha,H) \vminus x$. 
In (\ref{refine4:fml}.\ref{refine4:For}) $\alpha$ and $r$ have now been used 
so $H$ must be updated to \,$H$+$\alpha r$\, and hence 
\,$\alpha \vminus A(r)$\, becomes \,$(\alpha,H$+$\alpha r) \vminus A(r)$. 
Also in (\ref{refine4:fml}.\ref{refine4:Against}.\ref{refine4:TeamDefeat}) 
$\alpha$ and $t$ have been used 
so $H$ must be updated to \,$H$+$\alpha t$\, and hence 
\,$\alpha \vminus A(t)$\, becomes \,$(\alpha,H$+$\alpha t) \vminus A(t)$. 
Similarly in 
(\ref{refine4:fml}.\ref{refine4:Against}.\ref{refine4:DisableByNoProof}) 
$\alpha'$ and $s$ have been used and 
so $H$ must be updated to \,$H$+$\alpha' s$\, and hence 
\,$\alpha' \not\vminus A(s)$\, becomes 
\,$(\alpha',H$+$\alpha' s) \not\vminus A(s)$. 
Finally to prevent looping we must be sure that 
\,$\alpha r \nte H$\, in (\ref{refine4:fml}.\ref{refine4:For}), 
\,$\alpha t \nte H$\, in 
(\ref{refine4:fml}.\ref{refine4:Against}.\ref{refine4:TeamDefeat}), 
and \,$\alpha' s \nte H$\, in 
(\ref{refine4:fml}.\ref{refine4:Against}.\ref{refine4:DisableByNoProof}). 

Incorporating these changes into Refinement \ref{Refine:5.4} gives our 
formal definition of the proof algorithms and proof relation $\vminus$. 
The letter I is attached to these final inference conditions. 

\begin{Defn} \label{Defn:|-} \raggedright
Suppose \,$\calP = (R,>)$\, is a plausible description, \,$\Ax = \Ax(R)$, 
\,$\alpha \e \Alg$, $H$ is an $\alpha$-history, and $f$ is a formula. 
The \defn{proof relation for} $\calP$, $\vminus$, and the 
\defn{proof algorithms} are defined by I\ref{|-set} to I\ref{|-fml}. 
\begin{compactenum}[{I}1)]
\item \label{|-set} 
If $F$ is a finite set of formulas then 
\,$(\alpha,H) \vminus F$ \ iff \ $\forall f \e F$, \,$(\alpha,H) \vminus f$. 

\item \label{|-fact} 
If \,$\Ax \models f$\, then \,$(\alpha,H) \vminus f$. \ 
Also \,$(\varphi,H) \vminus f$\, iff \,$\Ax \models f$. 

\item \label{|-fml} 
If \,$\Ax \not\models f$\, and \,$\alpha \neq \varphi$\, 
then \,$(\alpha,H) \vminus f$\, iff 
\,$\exists r \e R^s_d[f]$\, such that 
I\ref{|-fml}.\ref{|-For} and I\ref{|-fml}.\ref{|-Against} hold. 
	\begin{compactenum}[{I}\ref{|-fml}.1)]
	\item \label{|-For} 
	$\alpha r \nte H$\, and \,$(\alpha,H$+$\alpha r) \vminus A(r)$. 

	\item \label{|-Against} 
	$\forall s \e \Foe(\alpha,f,r)$ \ either 
		\begin{compactenum}[{I}\ref{|-fml}.\ref{|-Against}.1)]
		\item \label{TeamDefeat} 
		$\exists t \e R^s_d[f;s]$\, such that \,$\alpha t \nte H$\, 
		and \,$(\alpha,H$+$\alpha t) \vminus A(t)$; \ or 

		\item \label{DisableByNoProof} 
		$\alpha' s \nte H$\, and \,$(\alpha',H$+$\alpha' s) \not\vminus A(s)$. 
		\end{compactenum}
	\end{compactenum}
\end{compactenum}
\end{Defn} 

The following notation is useful. 

\begin{Defn} \label{Defn:alpha-provable, calP(alpha)} \raggedright 
If $\calP$ is a plausible description, \,$\alpha \e \Alg$, and 
$x$ is either a formula or a finite set of formulas then define \nl 
$x$ is \defn{$\alpha$-provable}\, iff 
\,$\alpha \vminus x$\, iff \,$(\alpha,()) \vminus x$, \ and \nl 
\,$\calP(\alpha) = \{f \e \Fml : \alpha \vminus f\}$\, 
to be the set of all $\alpha$-provable formulas.  
\end{Defn} 

\subsubsection*{A semantic aside}
Subsection \ref{Subsection:Proof Relation and Proof Algorithms}, 
and its culmination in Definition \ref{Defn:|-}, 
can be given a semantic interpretation. 
By Theorem \ref{Thm:Right Weakening}(\ref{Modus Ponens for strict rules}), 
the meaning of the strict rule \,$A \!\strArr\! f$\, is that 
for any proof algorithm $\alpha$, 
if \,$\alpha \vminus A$\, then \,$\alpha \vminus f$. 
The meaning of the defeasible rule \,$A \!\defArr\! f$\, is that 
for any non-factual proof algorithm $\alpha$, 
if \,$\alpha \vminus A$\, and the evidence against $f$ is defeated 
then \,$\alpha \vminus f$. 
Exactly what the evidence against $f$ is and how it is defeated is given by 
I\ref{|-fml}.\ref{|-Against}. 
By I\ref{|-fml}.\ref{|-Against} the warning rule 
\,$s = A \!\warnArr\! \neg f$\, can only be used as evidence against $f$; 
and exactly how $s$ can be defeated is also given by 
I\ref{|-fml}.\ref{|-Against}. 
Thus Definition \ref{Defn:|-} can be seen as giving a meaning to 
each of the three kinds of rules. 

Similarly Definition \ref{Defn:|-}, and the explanations preceding it, 
can be seen as giving a meaning to each of the proof algorithms. 
By I\ref{|-fact}, we see that \,$\varphi \vminus f$\, means \,$\Ax \models f$. 
Each of the non-factual proof algorithms, $\alpha$, regards 
$R^s_d[f]$ as the set of potential evidence for $f$; 
and how $\alpha$ establishes that there is actual evidence for $f$ is 
given by I\ref{|-fml}.\ref{|-For}. 
Given some actual evidence $r$ for $f$, 
the set that $\alpha$ regards as evidence against $f$ is $\Foe(\alpha,f,r)$. 
Exactly how this evidence can be defeated is given by 
I\ref{|-fml}.\ref{|-Against}. 

\subsection{A Proof Theory}
\label{Subsection:Proof Theory}

Definition \ref{Defn:|-} is recursive, 
however it can be iterated to yield a rooted tree --- 
defined in Definition \ref{Defn:EvaluationTree} --- 
that could be regarded as the structure of a proof in PPL. 
The nodes of this tree will have special labels called tags which we now define. 

\begin{Defn} \label{Defn:tag} \raggedright
Suppose $(R,>)$ is a plausible description, \,$\alpha \e \Alg$, 
$F$ is a finite set of formulas, $H$ is an $\alpha$-history, 
$f$ is a formula, \,$r \e R^s_d[f]$, \,$s \e R[\neg f]$, 
and $p$ is a node of a tree. \nl
The \defn{tag}, $t(p)$, of $p$ is a triple 
\,$t(p) = (\Subj(p), \op(p), \pv(p))$. \nl
The \defn{subject} of $p$, $\Subj(p)$, has one of the following forms: 
$(\alpha,H,F)$, $-(\alpha',H,F)$, $(\alpha,H,f)$, $(\alpha,H,f,r)$, 
or $(\alpha,H,f,r,s)$. \nl
The \defn{operation} of $p$, $\op(p)$, is either $\min$ (for minimum), 
$\max$ (for maximum), or $-$. 
If $\op(p)$ is min [resp. max, $-$] then $p$ is referred to as a min [resp. max, minus] node. \nl
The \defn{proof value} of $p$, $\pv(p)$, is either $+1$ or $-1$. 
\end{Defn} 

The arithmetic properties of the proof values are defined below. 
These are as expected, but note that 
\,$\max\{\} = -1$\, and \,$\min\{\} = +1$. 

\begin{Defn} \label{Defn:operations} \raggedright
Suppose \,$S \!\subseteq\! \{+1, -1\}$. 
\nl
\begin{tabular}{@{}lll}
1) \,$\min S = -1$\, iff \,$-1 \e S$. & \hspace{2em} 
	(3) \,$\max S = +1$\, iff \,$+1 \e S$. & \hspace{2em} 
	(5) \,$-$\,$-1 = +1$. \\
2) \,$\min S = +1$\, iff \,$-1 \nte S$. & \hspace{2em} 
	(4) \,$\max S = -1$\, iff \,$+1 \nte S$. & \hspace{2em} 
	(6) \,$-$\,$+1 = -1$. \\ 
\end{tabular} 
\end{Defn} 

So $\min$ and $\max$ act like quantifiers when applied to a set of proof values. 
That is, \nl 
$\min S = -1$\, iff \,there exists $v$ in $S$ such that \,$v = -1$; \nl 
$\max S = +1$\, iff \,there exists $v$ in $S$ such that \,$v = +1$; \nl 
$\min S = +1$\, iff \,for all $v$ in $S$, \,$v = +1$; \quad and \nl
$\max S = -1$\, iff \,for all $v$ in $S$, \,$v = -1$. 

\begin{Defn} \label{Defn:EvaluationTree} \raggedright
Let \,$\calP = (R,>)$\, be a plausible description. 
Then $T$ is an \defn{evaluation tree of} $\calP$ iff 
$T$ is a rooted tree constructed as follows. 
Each node, $p$, of $T$ has exactly one tag, $t(p)$. 
For each node $p$ of $T$ there is exactly one number, $\#_p$, in 
$[\ref{Tset}..\ref{Tminus}]$ such that $p$ satisfies T$\#_p$ and T\ref{Tvalue}. 
\begin{compactenum}[T1)]
\item \label{Tset} 
$\Subj(p) = (\alpha,H,F)$, \,$\alpha \e \Alg$, $H$ is an $\alpha$-history, 
and $F$ is a finite set of formulas. 
Define \,$S(p) = \{(\alpha,H,f) : f \e F\}$. 
Then \,$\op(p) = \min$, $p$ has $|S(p)|$ children, and 
each element of $S(p)$ is the subject of exactly one child of $p$. 
If \,$S(p) = \{\}$\, then \,$\pv(p) = +1$. 

\item \label{Tfact} 
$\Subj(p) = (\alpha,H,f)$, \,$\alpha \e \Alg$, $H$ is an $\alpha$-history, 
$f$ is a formula, and \,$\Ax \models f$. \nl
Then $p$ has no children and \,$t(p) = ((\alpha,H,f), \min, +1)$. 

\item \label{Tfml} 
$\Subj(p) = (\alpha,H,f)$, \,$\alpha \in \Alg \!-\!\{\varphi\}$, 
$H$ is an $\alpha$-history, $f$ is a formula, and 
\,$\Ax \not\models f$. 
Define 
\,$S(p)$ = $\{(\alpha,H,f,r) : \alpha r \nte H$\, and \,$r \e R^s_d[f]\}$. 
Then \,$\op(p) = \max$, $p$ has $|S(p)|$ children, and 
each element of $S(p)$ is the subject of exactly one child of $p$. 
If \,$S(p) = \{\}$\, then \,$\pv(p) = -1$. 

\item \label{TFor} 
$\Subj(p) = (\alpha,H,f,r)$, \,$\alpha \in \Alg \!-\!\{\varphi\}$, 
$H$ is an $\alpha$-history, $f$ is a formula, \,$\Ax \not\models f$, 
\,$\alpha r \nte H$, and \,$r \e R^s_d[f]$. 
Define \,$S(p) = \{(\alpha,H$+$\alpha r,A(r))\}$ $\cup$ 
$\{(\alpha,H,f,r,s) : s \e \Foe(\alpha,f,r)\}$. 
Then \,$\op(p) = \min$, $p$ has $|S(p)|$ children, and 
each element of $S(p)$ is the subject of exactly one child of $p$. 

\item \label{TDftd} 
$\Subj(p) = (\alpha,H,f,r,s)$, \,$\alpha \in \Alg \!-\! \{\varphi,\pi'\}$, 
$H$ is an $\alpha$-history, $f$ is a formula, \,$\Ax \not\models f$, 
\,$\alpha r \nte H$, \,$r \e R^s_d[f]$, and \,$s \e \Foe(\alpha,f,r)$. 
Define \,$S(p)$ = $\{(\alpha,H$+$\alpha t,A(t)) : 
\alpha t \nte H$\, and \,$t \e R^s_d[f;s]\}$ $\cup$ 
$\{-(\alpha',H$+$\alpha's,A(s)) : \alpha's \nte H\}$. 
Then \,$\op(p) = \max$, $p$ has $|S(p)|$ children, and 
each element of $S(p)$ is the subject of exactly one child of $p$. 
If \,$S(p) = \{\}$\, then \,$\pv(p) = -1$. 

\item \label{Tminus} 
$\Subj(p) = -(\alpha',H,F)$, 
\,$\alpha \in \{\pi,\psi,\beta,\beta'\} \!\cup\! \{\psi' :\ >$ is not empty$\}$, 
$H$ is an $\alpha$-history, and $F$ is a finite set of formulas. 
Then \,$\op(p) = -$; $p$ has exactly one child, say $p_1$; and 
\,$\Subj(p_1) = (\alpha',H,F)$. 

\item \label{Tvalue} 
If \,$\op(p) = \min$\, then \,$\pv(p) = \min\{\pv(c) : c$ is a child of $p\}$. \nl
If \,$\op(p) = \max$\, then \,$\pv(p) = \max\{\pv(c) : c$ is a child of $p\}$. \nl
If \,$\op(p) = -$\, and $c$ is the child of $p$ then \,$\pv(p) = -\pv(c)$. 
\end{compactenum}
\end{Defn} 

It is possible for an evaluation tree to be infinite. 
Although this can be prevented by insisting that 
the set of rules in a plausible description is finite, it is not necessary. 

\begin{Defn} \label{Defn:PlausibleTheory, PlausibleLogic} \raggedright 
A plausible description $\calP$ is a \defn{plausible theory} iff 
every evaluation tree of $\calP$ is finite. 
A \defn{Propositional Plausible Logic} consists of 
a plausible theory and its proof relation. 
\end{Defn} 

Both the proof relation $\vminus$ and evaluation trees are 
cumbersome to use for derivations by hand. 
So we shall define a proof function $P$, which is easier to use and 
is a straightforward translation of the proof relation $\vminus$ of 
Definition \ref{Defn:|-} into the function $P$ 
such that \,$(\alpha,H) \vminus x$ \ iff \ $P(\alpha,H,x) = +1$ \ and 
\ $(\alpha,H) \not\vminus x$ \ iff \ $P(\alpha,H,x) = -1$. 
The auxiliary functions: $\For$ (evidence for), and $\Dftd$ (defeated), 
are used in the definition of $P$. 

\begin{Defn} \label{Defn:P(...)} \raggedright
Suppose \,$\calP = (R,>)$\, is a plausible theory, \,$\alpha \e \Alg$, 
$H$ is an $\alpha$-history, and $f$ is a formula. 
The \defn{proof function for} $\calP$, $P$, and its auxiliary functions 
$\For$ and $\Dftd$ are defined by P\ref{Pset} to P\ref{PDftd}. 
\begin{compactenum}[P1)]
\item \label{Pset} 
If $F$ is a finite set of formulas, then 
\,$P(\alpha,H,F)$ = $\min\{P(\alpha,H,f) : f \e F\}$. 

\item \label{Pfact} 
If \,$\Ax \models f$\, then \,$P(\alpha,H,f) = +1$. \ 
Also \,$P(\varphi,H,f) = +1$\, iff \,$\Ax \models f$. 

\item \label{Pfml} 
If \,$\Ax \not\models f$\, and \,$\alpha \neq \varphi$\, then \nl
\,$P(\alpha,H,f)$ = $\max\{\For(\alpha,H,f,r) : 
\alpha r \nte H$\, and \,$r \e R^s_d[f]\}$. 

\item \label{PFor} 
If \,$\Ax \not\models f$\, and \,$\alpha \neq \varphi$\, and 
\,$\alpha r \nte H$\, and \,$r \e R^s_d[f]$\, then \nl
\,$\For(\alpha,H,f,r)$ = 
$\min[\{P(\alpha,H$+$\alpha r,A(r))\}$ $\cup$ 
$\{\Dftd(\alpha,H,f,r,s) : s \e \Foe(\alpha,f,r)\}]$. 

\item \label{PDftd} 
If \,$\Ax \not\models f$\, and \,$\alpha \in \Alg \!-\! \{\varphi, \pi'\}$\, 
and \,$\alpha r \nte H$\, and \,$r \e R^s_d[f]$\, and 
\,$s \e \Foe(\alpha,f,r)$\, then \nl 
$\Dftd(\alpha,H,f,r,s)$ = $\max[\{P(\alpha,H$+$\alpha t,A(t)) : 
\alpha t \nte H$\, and \,$t \e R^s_d[f;s]\}$ $\cup$ 
$\{-P(\alpha', H$+$\alpha's, A(s)) : \alpha's \nte H\}]$. 
\end{compactenum}
\end{Defn} 

We end this subsection by stating the relationship between 
proof relations (Definition \ref{Defn:|-}), 
evaluation trees (Definition \ref{Defn:EvaluationTree}), and 
proof functions (Definition \ref{Defn:P(...)}). 
But before we can do this we need the following notation. 

\begin{Defn} \label{Defn:T[alpha,H,x],T(alpha,H,x)} \raggedright 
Suppose \,$\calP = (R,>)$\, is a plausible theory, \,$\alpha \e \Alg$, 
$H$ is an $\alpha$-history, and 
$x$ is either a formula or finite set of formulas. 
Let $T[\alpha,H,x]$ denote the evaluation tree of $\calP$ 
whose root node has the subject $(\alpha,H,x)$. 
Let $T(\alpha,H,x)$ denote the proof value of the root node of $T[\alpha,H,x]$. 
\end{Defn} 

\begin{Thm}[Notational Equivalence] 
\label{Thm:P iff |- iff T} \raggedright
Suppose $\calP$ is a plausible theory, \,$\alpha \e \Alg$, 
$H$ is an $\alpha$-history, and 
$x$ is either a formula or a finite set of formulas. \nl
Then \,$P(\alpha,H,x) = +1$ \ iff \ $(\alpha,H) \vminus x$ \ iff \ $T(\alpha,H,x) = +1$. 
\end{Thm} 

The idea of `logical consequence' in PPL is defined and most easily understood 
by considering the proof relation $\vminus$ of Definition \ref{Defn:|-}. 
The evaluation trees of Definition \ref{Defn:EvaluationTree} are 
mainly used to prove results about PPL. 
Proof functions (Definition \ref{Defn:P(...)}) make hand evaluations easier. 
So the equivalences expressed in 
Theorem \ref{Thm:P iff |- iff T} are essential. 

\subsection{A Truth Theory}
\label{Subsection:Truth Theory}

Logics often have a function from the set of all formulas to 
a set of truth values such that 
(a) the truth value of a formula is related to its proof value, and \nl
(b) the truth value of a formula is related to the truth values of its parts. \nl
Subsection \ref{Subsection:Truth Values} deals with (b), 
while this subsection is concerned with (a). 

Consider the possibilities that could occur when the proof algorithm 
$\alpha$ evaluates the evidence for and against the formula $f$. 
If there is sufficient evidence for both $f$ and $\neg f$ then, 
as far as $\alpha$ is concerned, $f$ and $\neg f$ are ambiguous, 
and so both should be assigned the \textbf{ambiguous} truth value $\tv{a}$. 
If there is sufficient evidence for $f$ but 
insufficient evidence for $\neg f$ then, as far as $\alpha$ is concerned, 
$f$ is usually true and $\neg f$ is usually false, 
so $f$ should be assigned the \textbf{usually true} truth value $\tv{t}$ and 
$\neg f$ should be assigned the \textbf{usually false} truth value $\tv{f}$. 
If there is insufficient evidence for both $f$ and $\neg f$ then 
$\alpha$ does not know enough about $f$ or about $\neg f$, 
and so both should be assigned the \textbf{undetermined} truth value $\tv{u}$. 

Since the truth value of a formula, $f$, 
depends on the proof algorithm, $\alpha$, evaluating its evidence, 
we need a veracity (or truth) function $V$ such that $V(\alpha,f)$ is in 
the set of plausible truth values $\{\tv{a}, \tv{t}, \tv{f}, \tv{u}\}$. 

\begin{Defn} \label{Defn:V(.,.)} \raggedright 
Suppose \,$\calP = (R,>)$\, is a plausible theory, \,$\alpha \e \Alg$, and 
$f$ is any formula. 
The \defn{truth function for} $\calP$, $V$, from \,$\Alg \!\times\! \Fml$\, to 
the \defn{set of plausible truth values} $\{\tv{a}, \tv{t}, \tv{f}, \tv{u}\}$ 
is defined by V\ref{tv{a}} to V\ref{tv{u}}. 
\begin{compactenum}[V1)]
\item \label{tv{a}} 
$V(\alpha,f) = \tv{a}$ \ iff \ $\alpha \vminus f$\, and 
							\,$\alpha \vminus \neg f$. 
\item \label{tv{t}} 
$V(\alpha,f) = \tv{t}$ \ iff \ $\alpha \vminus f$\, and 
							\,$\alpha \not\vminus \neg f$. 
\item \label{tv{f}} 
$V(\alpha,f) = \tv{f}$ \ iff \ $\alpha \not\vminus f$\, and 
							\,$\alpha \vminus \neg f$. 
\item \label{tv{u}} 
$V(\alpha,f) = \tv{u}$ \ iff \ $\alpha \not\vminus f$\, and 
							\,$\alpha \not\vminus \neg f$. 
\end{compactenum}
\end{Defn} 

Now that PPL is defined we need to show that it is well-behaved and 
satisfies all the principles in Section \ref{Section:Principles}. 
But before we do that it is worthwhile to get a better understanding 
of the logic by applying it to some examples. 

\section{Examples} 
\label{Section:Examples}

We shall show how PPL represents and reasons with the first three 
signpost examples in Section \ref{Section:Principles}. 
To save space and effort we shall use some of the theorems in 
Section \ref{Section:Properties of PPL}; 
this will also illustrate some of the utility of these theorems. 
In some of the following examples we shall use the following equations 
denoted by $\dagger$ and $\Box$. \nl
$\dagger$) $P(\alpha,H,\{f\}) = P(\alpha,H,f)$, by P\ref{Pset}. \nl
$\Box$) $P(\alpha,H,\{\}) = \min\{\} = +1$, by P\ref{Pset}.

\subsection{The Non-Monotonicity Example} 
\label{SubSection:Non-Monotonicity}

Recall the following from Example \ref{Eg:Non-Monotonicity}. 
\begin{compactenum}[1)]
\item $a$ is probably true. 
\item $\neg a$ is (definitely) true. 
\end{compactenum}
We show that from (1) the conclusion is `$a$ is plausible'; and 
from (1) and (2), that `$a$ is plausible' cannot be deduced, 
but `$\neg a$ is true' can be. 

The plausible theory $(R,>)$ which models (1) is defined as follows. 
The priority relation $>$ is empty, and 
\,$R = \{r_a\}$, where \ $r_a \mbox{ is \,} \{\} \!\defArr\! a$. 
So \,$R_s = \{\} = \Ax(R) = \Ax$, \ 
$R[a] = \{r_a\}$, \ $R[\neg a] = \{\}$. 
Also if \,$l \e \{a,\neg a\}$\, and \,$s \e R$\, then \,$R[l;s] = \{\}$. 

\begin{Eval}$\alpha \e \{\pi,\psi,\beta\}$\, and \,$\alpha \vminus a$ 
\label{Eval:Non-Monotonicity1} \raggedright
\begin{compactenum}[1$\alpha$)] \raggedright
\item 
	$P(\alpha,(),a)$ = $\For(\alpha,(),a,r_a)$, by P\ref{Pfml}
\item 
	= $P(\alpha,(\alpha r_a),\{\})$, by P\ref{PFor}, and 
	\,$\Foe(\alpha,a,r_a) = \{\}$
\item 
	= $+1$, by $\Box$
\end{compactenum}
\end{Eval} 

Some evaluations can be parameterised by the proof algorithm. 
The range of such a parameter is given after the number of the evaluation. 
If an evaluation proves or disproves something then this is given 
after the number of the evaluation. 

Evaluation \ref{Eval:Non-Monotonicity1} and 
Theorem \ref{Thm:P iff |- iff T}(Notational Equivalence), shows that 
$\pi$, $\psi$, and $\beta$ can prove $a$ using only (1). 

The plausible theory $(R,>)$ which models (1) and (2) is defined as follows. 
The priority relation $>$ is empty, and 
\,$R = \{r_a, r^s_{na}\}$, where \ 
$r_a \mbox{ is \,} \{\} \!\defArr\! a$, \ and \ 
$r^s_{na} \mbox{ is \,} \{\} \!\strArr\! \neg a$. 
Since \,$R_s = \{r^s_{na}\}$, \,$\Ax(R) = \Ax = \{\neg a\}$. 
So by P\ref{Pfact}, if \,$\alpha \e \{\varphi, \pi, \psi, \beta\}$\, 
then \,$P(\alpha, (), \neg a) = +1$. 

Hence using only (1) and (2), $\neg a$ is certain, 
and by Theorem \ref{Thm:Consistency}(\ref{phi,pi,psi,beta,beta'})(Consistency), 
$\pi$, $\psi$, and $\beta$ cannot prove $a$.

\subsection{The Ambiguity Puzzle} 
\label{SubSection:Ambiguity}

We show that the $\pi$ and $\psi$ proof algorithms are ambiguity propagating 
and that the $\beta$ proof algorithm is ambiguity blocking. 

The plausible theory $(R,>)$ which models the Ambiguity Puzzle 
(Example \ref{Eg:Ambiguity}) is defined as follows. 
The priority relation $>$ is empty, and 
\,$R = \{r_a, r_{na}, r_b, r_{anb}\}$, where \ 
$r_a \mbox{ is \,} \{\} \!\defArr\! a$, \ 
$r_{na} \mbox{ is \,} \{\} \!\defArr\! \neg a$, \ 
$r_b \mbox{ is \,} \{\} \!\defArr\! b$, \ and \ 
$r_{anb} \mbox{ is \,} \{a\} \!\defArr\! \neg b$. 

Since \,$R_s = \{\}$, \,$\Ax(R) = \Ax = \{\}$. 
So \,$R[a] = \{r_a\}$, \ $R[b] = \{r_b\}$, 
\ $R[\neg a] = \{r_{na}\}$, and \,$R[\neg b] = \{r_{anb}\}$. 
If \,$l \e \{a,\neg a,b,\neg b\}$\, and \,$s \e R$\, then \,$R[l;s] = \{\}$. 

\begin{Eval}$\alpha \e \{\pi,\psi,\beta\}$ \ 
\label{Eval:Ambiguity} \raggedright
\begin{compactenum}[1$\alpha$)] \raggedright
\item 
	$P(\alpha,(),b)$ = $\For(\alpha,(),b,r_b)$, by P\ref{Pfml}
\item 
	= $\min\{P(\alpha,(\alpha r_b),\{\})$, $\Dftd(\alpha,(),b,r_b,r_{anb})\}$, 
	by P\ref{PFor}
\item 
	= $\Dftd(\alpha,(),b,r_b,r_{anb})$, by $\Box$
\item 
	= $-P(\alpha',(\alpha' r_{anb}),a)$, by P\ref{PDftd}, $\dagger$
\item 
	= $-\For(\alpha',(\alpha' r_{anb}),a,r_a)$, by P\ref{Pfml}
\end{compactenum}
\end{Eval} 

\begin{Eval} $\alpha \e \{\pi,\psi\}$\, and \,$\alpha \not\vminus b$ 
\label{Eval:AmbiguityProp} \raggedright
\begin{compactenum}[1$\alpha$)] \raggedright
\addtocounter{enumi}{4}
\item 
	$P(\alpha,(),b)$ = $-\For(\alpha',(\alpha' r_{anb}),a,r_a)$, 
	by Evaluation \ref{Eval:Ambiguity}
\item 
	= $-P(\alpha',(\alpha' r_{anb},\alpha' r_a),\{\})$, by P\ref{PFor}
\item 
	= $-1$, by $\Box$.
\end{compactenum}
\end{Eval} 

\begin{Eval} $\beta \vminus b$ 
\label{Eval:AmbiguityBlock} \raggedright
\begin{compactenum}[1$\beta$)] \raggedright
\addtocounter{enumi}{4}
\item 
	$P(\beta,(),b)$ = $-\For(\beta',(\beta' r_{anb}),a,r_a)$, 
	by Evaluation \ref{Eval:Ambiguity}
\item 
	= $-\min\{P(\beta',(\beta'r_{anb},\beta'r_a),\{\})$, 
	$\Dftd(\beta',(\beta'r_{anb}),a,r_a,r_{na})\}$, by P\ref{PFor}
\item 
	= $-\Dftd(\beta',(\beta'r_{anb}),a,r_a,r_{na})$, by $\Box$
\item 
	= $--P(\beta,(\beta'r_{anb},\beta r_{na}),\{\})$, by P\ref{PDftd}
\item 
	= $+1$, by $\Box$.
\end{compactenum}
\end{Eval} 

By Evaluation \ref{Eval:AmbiguityProp} and 
Theorems \ref{Thm:P iff |- iff T}(Notational Equivalence) 
and \ref{Thm:Decisiveness}(Decisiveness), 
$\pi$ and $\psi$ cannot prove $b$ and so they are ambiguity propagating. 
By Evaluation \ref{Eval:AmbiguityBlock} and 
Theorem \ref{Thm:P iff |- iff T}(Notational Equivalence), 
$\beta$ proves $b$ and so is ambiguity blocking. 

\subsection{The 3-lottery Example} 
\label{SubSection:3-lottery}

Recall the following from Example \ref{Eg:3-lottery}. 
\begin{compactenum}[1)] \raggedright
\item Exactly one element of $\{s_1, s_2, s_3\}$ is true. 
\item Each element of $\{\neg s_1, \neg s_2, \neg s_3\}$ is usually true. 
\item The disjunction of any pair of elements of $\{s_1, s_2, s_3\}$ 
	is usually true.  
\end{compactenum}

From (2) we get $r_{11}$ to $r_{13}$ below. 
From (3) we get $r_{14}$ to $r_{16}$ below. 
From (1) we have \,$\OR\{s_1, s_2, s_3\}$, \,$\neg \AND\{s_1, s_2\}$, 
\,$\neg \AND\{s_1, s_3\}$, and \,$\neg \AND\{s_2, s_3\}$.  
Converting these facts to clauses gives: 
$\Ax$ = $\{\OR\{s_1,s_2,s_3\}$, $\OR\{\neg s_1, \neg s_2\}$, 
$\OR\{\neg s_1, \neg s_3\}$, $\OR\{\neg s_2, \neg s_3\}\}$. 

The plausible theory $(R,>)$ which models this situation is defined as follows. 
The priority relation $>$ is empty, and 
\,$R = \{r_1, r_2, ..., r_{16}\}$, where \ \ 
$r_1$: $\{\} \strArr \AND \Ax$, \nl
\begin{tabular}{@{}l@{\hspace{3em}}l@{\hspace{3em}}l} 
$r_2$: $\{\neg s_1\} \strArr \OR\{s_2, s_3\}$, & 
$r_5$: $\{\AND\{\neg s_2, \neg s_3\}\} \strArr s_1$, &
$r_8$: $\{s_1\} \strArr \AND\{\neg s_2, \neg s_3\}$, \\

$r_3$: $\{\neg s_2\} \strArr \OR\{s_1, s_3\}$, & 
$r_6$: $\{\AND\{\neg s_1, \neg s_3\}\} \strArr s_2$, &
$r_9$: $\{s_2\} \strArr \AND\{\neg s_1, \neg s_3\}$, \\

$r_4$: $\{\neg s_3\} \strArr \OR\{s_1, s_2\}$, & 
$r_7$: $\{\AND\{\neg s_1, \neg s_2\}\} \strArr s_3$, &
$r_{10}$: $\{s_3\} \strArr \AND\{\neg s_1, \neg s_2\}$, \smallskip \\

$r_{11}$: $\{\} \defArr \neg s_1$, & 
$r_{14}$: $\{\} \defArr \OR\{s_1, s_2\}$, & \\

$r_{12}$: $\{\} \defArr \neg s_2$, & 
$r_{15}$: $\{\} \defArr \OR\{s_1, s_3\}$, & \\

$r_{13}$: $\{\} \defArr \neg s_3$, & 
$r_{16}$: $\{\} \defArr \OR\{s_2, s_3\}$. & 
\end{tabular} 

Let \,$U = \{\neg s_1, \neg s_2, \OR\{s_1, s_2\}\}$. 
If \,$\alpha \e \{\pi, \psi, \beta\}$\, then 
we show $\alpha$ proves each element of $U$, 
$\alpha$ cannot prove the negation of each element of $U$, 
and $\alpha$ cannot prove $\AND\{\neg s_1, \neg s_2\}$. 

Note \ $R^s_d[\neg s_1] = \{r_2,r_6,r_7,r_9,r_{10},r_{11},r_{16}\}$, 
\ and \ $R^s_d[s_1] = \{r_5,r_8\} = R^s_d[\AND\{\neg s_2, \neg s_3\}]$. 

\begin{Eval} $\pi \vminus \neg s_1$ 
\label{Eval:3-lottery pi|- neg s_1} \raggedright
\begin{compactenum}[ \,1)] \raggedright
\item 
	$P(\pi,(),\neg s_1)$ = $\max\{\For(\pi,(),\neg s_1,r_i) : 
	i \e \{2,6,7,9,10,11,16\}\}$, by P\ref{Pfml}
\item 
	$\For(\pi,(),\neg s_1,r_{11})$ = $\min\{P(\pi,(\pi r_{11}),\{\})$, 
	$\Dftd(\pi,(),\neg s_1,r_{11},r_5)$, \nl \hspace*{18.6em}
	$\Dftd(\pi,(),\neg s_1,r_{11},r_8)\}$, by P\ref{PFor}
\item 
	= $\min\{-P(\pi',(\pi' r_5),\AND\{\neg s_2, \neg s_3\})$, 
	$-P(\pi',(\pi' r_8),s_1)\}$, by $\Box$, P\ref{PDftd}, $\dagger$
\item 
	$P(\pi',(\pi' r_5),\AND\{\neg s_2, \neg s_3\})$ = 
	$\For(\pi',(\pi' r_5),\AND\{\neg s_2, \neg s_3\},r_8)$, by P\ref{Pfml}
\item 
	= $P(\pi',(\pi' r_5, \pi' r_8),s_1)$, by P\ref{PFor}, $\dagger$
\item 
	= $\max\{\}$, by P\ref{Pfml}
\item 
	= $-1$. 
\item 
	$\therefore \For(\pi,(),\neg s_1,r_{11})$ = $-P(\pi',(\pi' r_8),s_1)$, 
	by (7) to (2)
\item 
	= $-\For(\pi',(\pi' r_8),s_1,r_5)$, by P\ref{Pfml}
\end{compactenum}
\begin{compactenum}[1)] \raggedright
\addtocounter{enumi}{9}
\item 
	= $-P(\pi',(\pi' r_8,\pi' r_5),\AND\{\neg s_2, \neg s_3\})$, 
	by P\ref{PFor}, $\dagger$
\item 
	= $-\max\{\}$, by P\ref{Pfml}
\item 
	= $+1$.
\item 
	$\therefore P(\pi,(),\neg s_1)$ = $+1$, by (12) to (8), and (1). 
\end{compactenum}
\end{Eval} 

Because the 3-lottery example is symmetric 
in $s_1$, $s_2$, and $s_3$, a very similar evaluation gives 
\,$P(\pi,(),\neg s_2)$ = $+1$\, and \,$P(\pi,(),\neg s_3)$ = $+1$. 
Hence by $\dagger$, \,$P(\pi,(),\{\neg s_3\})$ = $+1$. \nl
By Theorem \ref{Thm:P iff |- iff T}(Notational Equivalence), 
\,$\pi \vminus \neg s_1$, \,$\pi \vminus \neg s_2$, and 
\,$\pi \vminus \{\neg s_3\}$. \nl
By Theorem \ref{Thm:Right Weakening}(\ref{Modus Ponens for strict rules})(Modus Ponens for strict rules), 
using $r_4$, we get \,$\pi \vminus \OR\{s_1, s_2\}$. \nl
Thus $\pi$ proves each element in 
\,$U$ = $\{\neg s_1, \neg s_2, \OR\{s_1, s_2\}\}$. 

Suppose \,$\alpha \e \{\pi, \psi, \beta\}$. 
Then by Theorem \ref{Thm:Hierarchy}(The proof algorithm hierarchy), 
$\alpha$ proves each element of $U$. 
By Theorem \ref{Thm:Consistency}(\ref{phi,pi,psi,beta,beta'})(Consistency), 
the negation of each element of $U$ cannot be proved by $\alpha$. 
Hence by 
Theorem \ref{Thm:Right Weakening}(\ref{Right Weakening})(Right Weakening), 
$\alpha$ cannot prove $\AND\{\neg s_1, \neg s_2\}$. 


\section{Properties of Propositional Plausible Logic (PPL)} 
\label{Section:Properties of PPL}

We shall show that PPL is well-behaved and 
satisfies all the principles in Section \ref{Section:Principles}. 

\begin{Thm}[Decisiveness] 
\label{Thm:Decisiveness} \raggedright
Suppose $\calP$ is a plausible theory, \,$\alpha \e \Alg$, 
$H$ is an $\alpha$-history, and 
$x$ is either a formula or a finite set of formulas. \nl
Then either \,$T(\alpha,H,x) = +1$\, or \,$T(\alpha,H,x) = -1$, but not both. 
\end{Thm} 

Knowing that every evaluation will terminate is very comforting. 
So at the end of each evaluation either we will have a proof or we will not. 
If we do not have a proof then maybe there is a proof but we missed it. 
Fortunately decisiveness assures us that an evaluation will always terminate, 
and when it does we will have either a proof or a disproof --- 
that is a demonstration that there is no proof.

\begin{Thm} [Plausible Conjunction] 
\label{Thm:Plausible Conjunction}  \raggedright 
Suppose $(R,>)$ is a plausible description, \,$\Ax = \Ax(R)$, 
\,$\alpha \e \Alg$, $H$ is an $\alpha$-history, and 
$f$ and $g$ are both formulas. \nl
If \,$\Ax \models f$\, and \,$(\alpha,H) \vminus g$\, 
then \,$(\alpha,H) \vminus \AND \{f,g\}$. 
\end{Thm}  

The Plausible Conjunction theorem shows that each $\alpha$ satisfies the 
Plausible Conjunction Principle 
(Principle \ref{Prin:Conjunction}.\ref{Prin:Plausible Conjunction Principle}). 

\begin{Thm}[Right Weakening] 
\label{Thm:Right Weakening}
Suppose $(R,>)$ is a plausible description, \,$\Ax = \Ax(R)$, 
\,$\alpha \e \Alg$, $H$ is an $\alpha$-history, and 
$f$ and $g$ are both formulas. 
\begin{compactenum}[1)]
\item \label{Strong Right Weakening} 
	If \,$(\alpha,H) \vminus f$\, and \,$\Ax\!\cup\!\{f\} \models g$\, 
	then \,$(\alpha,H) \vminus g$. [Strong Right Weakening]
\item \label{Right Weakening} 
	If \,$(\alpha,H) \vminus f$\, and \,$f \models g$\, 
	then \,$(\alpha,H) \vminus g$. [Right Weakening]
\item \label{Modus Ponens for strict rules} 
	If \,$A \!\strArr\! g \in R_s$\, and \,$(\alpha,H) \vminus A$\, 
	then \,$(\alpha,H) \vminus g$. [Modus Ponens for strict rules]
\end{compactenum}
\end{Thm} 

Theorem \ref{Thm:Right Weakening}(\ref{Strong Right Weakening}) shows that 
$\alpha$ has the strong right weakening property, and hence has all the right weakening properties mentioned in Subsection \ref{Subsection:Right Weakening}. 

\begin{Thm}[Consistency] 
\label{Thm:Consistency} \raggedright
Suppose $(R,>)$ is a plausible theory, \,$\Ax = \Ax(R)$, 
\,$\alpha \in \{\varphi,\pi,\psi,\beta,\beta'\}$, and 
both $f$ and $g$ are any formulas. 
\begin{compactenum}[1)]
\item \label{phi,pi,psi,beta,beta'} 
	If \,$\alpha \vminus f$\, and \,$\alpha \vminus g$\, 
	then \,$\Ax \!\cup\! \{f,g\}$\, is satisfiable. 
\item \label{psi-psi'} 
	If \,$(\psi,H) \vminus f$\, then \,$(\psi',H) \not\vminus \neg f$. 
\item \label{pi-pi'} 
	Suppose that whenever \,$s \e R^s_d[\neg f]$\, and 
	\,$(\pi',H$+$\pi' s) \vminus A(s)$\, then \,$R^s_d[f;s] = \{\}$. \nl 
	If \,$(\pi,H) \vminus f$\, then \,$(\pi',H) \not\vminus \neg f$. 
\end{compactenum}
\end{Thm} 

Part \ref{phi,pi,psi,beta,beta'} of Theorem \ref{Thm:Consistency} says 
that PPL is strongly 2-consistent. 
Theorem \ref{Thm:Consistency}(\ref{psi-psi'}) says that 
if there is sufficient evidence for $\psi$ to prove $f$ 
then the evidence for $\neg f$ is too weak for $\psi'$ to register. 
Theorem \ref{Thm:Consistency}(\ref{pi-pi'}) gives conditions under which 
a similar statement can be said about $\pi$ and $\pi'$. 
In particular when either $R^s_d[\neg f]$ or $>$ is empty. 

\begin{Thm}[Truth Values] 
\label{Thm:Truth values} 
Suppose $(R,>)$ is a plausible theory, \,$\alpha \e \Alg$, 
$F$ is a finite set of formulas, and $f$ is a formula. 
\begin{compactenum}[ \,1)]
\item \label{V(alpha,notnotf) = V(alpha,f)} 
	$V(\alpha,\neg \neg f) = V(\alpha,f)$. 
\item \label{V(alpha,f) = tv{t} iff V(alpha,not f) = tv{f}} 
	$V(\alpha,f) = \tv{t}$\, iff \,$V(\alpha,\neg f) = \tv{f}$. 
\item \label{V(alpha,f) = tv{f} iff V(alpha,not f) = tv{t}} 
	$V(\alpha,f) = \tv{f}$\, iff \,$V(\alpha,\neg f) = \tv{t}$. 
\item \label{V(alpha,f) = tv{a} iff V(alpha,not f) = tv{a}} 
	$V(\alpha,f) = \tv{a}$\, iff \,$V(\alpha,\neg f) = \tv{a}$. 
\item \label{V(alpha,f) = tv{u} iff V(alpha,not f) = tv{u}} 
	$V(\alpha,f) = \tv{u}$\, iff \,$V(\alpha,\neg f) = \tv{u}$. 
\item \label{If V(alpha,AND F) = tv{t} then V(alpha,f) = tv{t}} 
	If \,$V(\alpha,\AND F) = \tv{t}$\, then 
	for each $f$ in $F$, \,$V(\alpha,f) = \tv{t}$. 
\item \label{If V(alpha,f) = tv{t} then V(alpha,OR F) = tv{t}} 
	If \,$f \e F$\, and \,$V(\alpha,f) = \tv{t}$\, 
	then \,$V(\alpha,\OR F) = \tv{t}$. 
\item \label{If alpha isin {phi,pi,psi,beta,beta'} then V(alpha,f) isin {t,f,u}.} 
	If \,$\alpha \e \{\varphi,\pi,\psi,\beta,\beta'\}$\, 
	then \,$V(\alpha,f) \e \{\tv{t},\tv{f},\tv{u}\}$. 
\item \label{If V(alpha,f) = a then alpha isin {psi',pi'}.} 
	If \,$V(\alpha,f) = \tv{a}$\, then \,$\alpha \e \{\psi',\pi'\}$. 
\end{compactenum}
\begin{compactenum}[1)] \raggedright
\addtocounter{enumi}{9}
\item \label{alpha is complete} 
	If \,$V(\alpha,f) = \tv{t}$\, then \,$\alpha \vminus f$. (completeness)
\item \label{alpha is sound} 
	If \,$\alpha \e \{\varphi,\pi,\psi,\beta,\beta'\}$\, and 
	\,$\alpha \vminus f$\, then \,$V(\alpha,f) = \tv{t}$. (soundness)
\end{compactenum}
\end{Thm} 

Parts \ref{V(alpha,notnotf) = V(alpha,f)} to 
\ref{V(alpha,f) = tv{u} iff V(alpha,not f) = tv{u}} of 
Theorem \ref{Thm:Truth values} show that negation is truth-functional 
with desirable properties. 
In Subsection \ref{Subsection:Truth Values} the desired relation 
between the truth values of a conjunction and its conjuncts 
is given by statement (4), and the desired relation between 
the truth values of a disjunction and its disjuncts is given by statement (5). 
Parts \ref{If V(alpha,AND F) = tv{t} then V(alpha,f) = tv{t}} and 
\ref{If V(alpha,f) = tv{t} then V(alpha,OR F) = tv{t}} of 
Theorem \ref{Thm:Truth values} show that PPL satisfies these relationships. 

The primed algorithms $\beta'$, $\psi'$, and $\pi'$, 
assess the significance of evidence against a formula. 
Theorem \ref{Thm:Truth values}(\ref{If V(alpha,f) = a then alpha isin {psi',pi'}.}) 
shows that the threshold of significance for $\psi'$ and $\pi'$ is so low that 
they can assess the evidence against both $f$ and $\neg f$ as significant. 
However Theorem \ref{Thm:Truth values}(\ref{If alpha isin {phi,pi,psi,beta,beta'} then V(alpha,f) isin {t,f,u}.}) 
shows that the other algorithms have a 3-valued truth system.
The expected completeness and soundness results are given by parts 
\ref{alpha is complete} and \ref{alpha is sound} of 
Theorem \ref{Thm:Truth values}. 

The final result shows the relationships between the various proof algorithms. 
Recall Definition \ref{Defn:alpha-provable, calP(alpha)} defines 
$\calP(\alpha)$ to be the set of all formulas provable from $\calP$ using 
the proof algorithm $\alpha$. 

\begin{Thm} [The proof algorithm hierarchy]
\label{Thm:Hierarchy} 
Suppose \,$\calP = (R,>)$\, is a plausible theory. 
\begin{compactenum}[1)] \raggedright
\item \label{hierarchy} 
	$\calP(\varphi) \subseteq \calP(\pi) \subseteq \calP(\psi) \subseteq 
	\calP(\beta) = \calP(\beta') \subseteq \calP(\psi') \subseteq \calP(\pi')$. 
\item \label{hierarchy with empty >} 
	If $>$ is empty then 
	\,$\calP(\varphi) \subseteq \calP(\pi) = \calP(\psi) \subseteq 
	\calP(\beta) = \calP(\beta') \subseteq \calP(\psi') = \calP(\pi')$. 
\end{compactenum}
\end{Thm} 

So $\beta'$ proves exactly the same formulas as $\beta$. 
Also if $>$ is empty then $\pi$ and $\psi$ prove exactly the same formulas,
as do $\pi'$ and $\psi'$. 
The set of formulas proved by $\pi'$ is very similar to 
the union of all extensions of an extension based logic, like Default Logic. 

The hierarchy shown in Theorem \ref{Thm:Hierarchy} is consistent with 
the intuition that ambiguity propagating proof algorithms 
are more cautious than ambiguity blocking algorithms. 
A similar hierarchy for a Defeasible Logic, 
also consistent with this intuition, 
is given in Section 5 of \cite{BAGM:2010}. 

When a logic has several proof algorithms 
it is important to determine how they relate, 
because this gives a greater theoretical understanding of the logic. 
Theorem \ref{Thm:Hierarchy} shows that the proof algorithms of PPL 
are totally ordered according to reliability or level of confidence. 
Suppose that the plausible-reasoning situation gives no information 
concerning whether ambiguity should be blocked or propagated. 
If the proof algorithm hierarchy is not totally ordered 
then we could have two incomparable algorithms 
only one of which proved the formula of interest. 
In such circumstances it is not clear what should be concluded. 
By Theorem \ref{Thm:Hierarchy} no such dilemma can occur in PPL. 

\bigskip

We shall now check that PPL satisfies all the principles in 
Section \ref{Section:Principles}. 

PPL has strict and defeasible rules, 
and so can distinguish between factual and plausible statements. 
Moreover PPL does not use numbers, like probabilities, 
that could lead to a proved formula 
being more precise than the information used to derive it. 
So the Representation Principle (Principle \ref{Prin:Representation}) 
is satisfied. 

The correspondence between the general `plausible-structure' notation of 
Sections \ref{Section:Introduction} and \ref{Section:Principles} and 
the particular notation of PPL is as follows. 
The plausible-structure $\mathcal{S}$ corresponds to the 
plausible description \,$\calP = (R,>)$. 
If we let \,$\Ax = \Ax(R)$, then $\Fact(\mathcal{S})$ corresponds to $\Ax$, 
$\Thms(\Fact(\mathcal{S}))$ corresponds to \,$\{f : \Ax \models f\}$, and 
$\Thms(\mathcal{L},\alpha,\mathcal{S})$ corresponds to $\calP(\alpha)$. 
For the rest of this section suppose $\alpha$ is in $\{\pi,\psi,\beta\}$. 

As explained in the paragraph above Refinement \ref{Refine:5.1}, 
the Evidence Principle (Principle 
\ref{Prin:Evidence, Non-monotonicity}.\ref{Prin:Evidence}) is satisfied. 
In Subsection \ref{SubSection:Non-Monotonicity} we showed that 
$\alpha$ satisfies the Non-Monotonicity Principle (Principle 
\ref{Prin:Evidence, Non-monotonicity}.\ref{Prin:Non-Monotonicity}). 
Hence $\alpha$ satisfies Principle \ref{Prin:Evidence, Non-monotonicity}.

In Subsection \ref{SubSection:3-lottery} we showed that all three elements of 
\,$U = \{\neg s_1, \neg s_2, \OR\{s_1, s_2\}\}$\, were $\alpha$-provable; 
but that the conjunction \,$\AND\{\neg s_1, \neg s_2\}$\, 
was not $\alpha$-provable. 
Thus $\alpha$ satisfies the Non-Conjunction Principle 
(Principle \ref{Prin:Conjunction}.\ref{Prin:Non-Conjunction}). 
Theorem \ref{Thm:Plausible Conjunction} shows that $\alpha$ satisfies the 
Plausible Conjunction Principle 
(Principle \ref{Prin:Conjunction}.\ref{Prin:Plausible Conjunction Principle}). 
By Theorem \ref{Thm:Consistency}(\ref{phi,pi,psi,beta,beta'}), 
both $s_1$ and $s_2$ are not $\alpha$-provable. 
Thus $\alpha$ satisfies the Non-Disjunction Principle 
(Principle \ref{Prin:Non-Disjunction}). 
Because $U$ is not satisfiable $\alpha$ satisfies the 
Non-3-Consistency Principle 
(Principle \ref{Prin:Consistency}.\ref{Prin:Non-3-Consistency}). 
Theorem \ref{Thm:Consistency}(\ref{phi,pi,psi,beta,beta'}) shows that 
$\alpha$ satisfies the Strong 2-Consistency Principle 
(Principle \ref{Prin:Consistency}.\ref{Prin:Strong 2-Consistency}) 
and so satisfies the 1-Consistency Principle 
(Principle \ref{Prin:Consistency}.\ref{Prin:1-Consistency}). 

By I\ref{|-fact} of Definition \ref{Defn:|-}, 
$\varphi$ and $\alpha$ are supraclassical. 
So by the remark after Principle \ref{Prin:Plausible Supraclassicality}, 
they satisfy the Plausible Supraclassicality Principle 
(Principle \ref{Prin:Plausible Supraclassicality}). 
Theorem \ref{Thm:Right Weakening}(\ref{Strong Right Weakening}) shows that 
$\alpha$ has the Strong Right Weakening property. 
So by the remark after Principle \ref{Prin:Plausible Right Weakening}, 
$\alpha$ satisfies the Plausible Right Weakening Principle 
(Principle \ref{Prin:Plausible Right Weakening}). 

By I\ref{|-fact} of Definition \ref{Defn:|-}, 
$\varphi$ is a factual proof algorithm. 
Subsection \ref{SubSection:Ambiguity} shows that 
$\pi$ and $\psi$ are ambiguity propagating proof algorithms, and 
$\beta$ is an ambiguity blocking proof algorithm. 
Also PPL makes the proof algorithm used explicit. 
Hence PPL satisfies the Many Proof Algorithms Principle 
(Principle \ref{Prin:Many Proof Algorithms}). 
Theorems \ref{Thm:Decisiveness} and \ref{Thm:P iff |- iff T} show that 
$\varphi$ and $\alpha$ satisfy the Decisiveness Principle 
(Principle \ref{Prin:Decisiveness}). 
The truth-value system given in Subsection \ref{Subsection:Truth Theory} 
and Theorem \ref{Thm:Truth values} 
shows that $\alpha$ satisfies the Included Middle Principle 
(Principle \ref{Prin:Included Middle}). 

Thus PPL satisfies all the principles in Section \ref{Section:Principles}. 

\section{Conclusion} 
\label{Section:Conclusion}

We have tried to characterise those propositional logics that 
do plausible reasoning by suggesting some principles 
that such logics should satisfy. 
Four important examples of plausible reasoning are presented, 
and several principles are derived from these examples. 

Propositional Plausible Logic (PPL) has been defined. 
It satisfies all the principles, and deals with negation, conjunction, 
and disjunction. 
PPL has been applied to the first three examples, 
and several theorems about PPL are proved in the appendices. 
PPL has been implemented by George Wilson 
under the direction of Dr.~Andrew Rock, 
who has implemented other Defeasible Logics. 
As far as we know, PPL is the only non-numeric non-monotonic logic 
that satisfies all the principles in Section \ref{Section:Principles} 
and also correctly reasons with all the examples in 
Section \ref{Section:Principles}. 

Future research could make PPL significantly more useful and powerful by 
incorporating variables in a similar way to the programming language Prolog. 

\acks{The author thanks Michael J. Maher for comments on an earlier version of this paper. 
The author also thanks Ren\'e Hexel for helpful discussions about the contents of Section \ref{Section:Principles} Principles of Plausible Reasoning. }


\section*{The Appendices} 

\appendix
\section{Proof of Theorems \ref{Thm:Decisiveness} and \ref{Thm:P iff |- iff T}}

\begin{Lem} \label{Lem:[dual-]clauses, implication} 
Let $L$ and $M$ be any two sets of literals. 
\begin{compactenum}[1)]
\item \label{clauses, implication} 
	$\OR L \models \OR M$\, iff 
	either \,$L \!\subseteq\! M$\, or $\OR M$ is a tautology. 
\item \label{dual-clauses, implication} 
	$\AND M \models \AND L$\, iff 
	either \,$L \!\subseteq\! M$\, or $\AND M$ is a contradiction. 
\end{compactenum}

\noindent \pf{Proof} 

Let $L$ and $M$ be any two sets of literals. 

(\ref{clauses, implication}) 
If \,$L \!\subseteq\! M$\, or $\OR M$ is a tautology 
then \,$\OR L \models \OR M$. 

Conversely suppose \,$\OR L \models \OR M$. 
If \,$L \!-\! M = \{\}$\, then \,$L \!\subseteq\! M$. 
So suppose there is a literal $l$ such that \,$l \in L \!-\! M$. 
If $\OR M$ is not a tautology then there is a valuation $v$ such that 
\,$v(\OR M) = \F$\, and \,$v(l) = \T$. 
But that contradicts \,$\OR L \models \OR M$, 
so $\OR M$ is must be a tautology. 

(\ref{dual-clauses, implication}) 
If \,$L \!\subseteq\! M$\, or $\AND M$ is a contradiction 
then \,$\AND M \models \AND L$. 

Conversely suppose \,$\AND M \models \AND L$. 
If \,$L \!-\! M = \{\}$\, then \,$L \!\subseteq\! M$. 
So suppose there is a literal $l$ such that \,$l \in L \!-\! M$. 
If $\AND M$ is not a contradiction then there is a valuation $v$ such that 
\,$v(\AND M) = \T$\, and \,$v(l) = \F$. 
But that contradicts \,$\AND M \models \AND L$, 
so $\AND M$ is must be a contradiction. 

\noindent \eop{EndProofLem\ref{Lem:[dual-]clauses, implication}} 
\end{Lem} 
\begin{Lem} \label{Lem:Err,Sat} 
Let $C$ be a set of clauses. 
\begin{compactenum}[1)]
\item \label{Err(C) is closed under nnot.} 
	$l \e \Err(C)$\, iff \,$\nnot l \e \Err(C)$. 
\item \label{Sat(C) subseteq C} 
	$\Sat(C) \subseteq C$. 
\item \label{The empty clause is not in Sat(C).} 
	$\OR\{\} \notin \Sat(C)$. 
\item \label{The empty clause is not in Res(Sat(C)).} 
	$\OR\{\} \notin \Res(\Sat(C))$. 
\item \label{C is satisfiable iff the empty clause is not in Res(C).} 
	$C$ is satisfiable iff \,$\OR\{\} \nte \Res(C)$. 
\item \label{Sat(C) is satisfiable.} 
	$\Sat(C)$ is satisfiable, $\Res(\Sat(C))$ is satisfiable, and 
	$\CorRes(\Sat(C))$ is satisfiable. 
\end{compactenum}

\noindent \pf{Proof} 

Let $C$ be a set of clauses. 

(\ref{Err(C) is closed under nnot.}, \ref{Sat(C) subseteq C}, 
\ref{The empty clause is not in Sat(C).}) 
These parts follow immediately from Definition \ref{Defn:Err,Sat}. 

(\ref{The empty clause is not in Res(Sat(C)).}) 
Assume \,$\OR\{\} \in \Res(\Sat(C))$. 
Since \,$\OR\{\} \notin \Sat(C)$, there is a literal $l$ such that 
\,$\OR\{l\} \e \Res(\Sat(C))$\, and \,$\OR\{\nnot l\} \e \Res(\Sat(C))$. 
Since \,$\Sat(C) \subseteq C$, \,$\Res(\Sat(C)) \subseteq \Res(C)$. 
So \,$\OR\{l\} \e \Res(C)$\, and \,$\OR\{\nnot l\} \e \Res(C)$. 
Hence \,$l \e \Err(C)$\, and \,$\nnot l \e \Err(C)$. 
So by Definition \ref{Defn:Err,Sat}, for all $c$ in $\Sat(C)$, 
\,$l \nte \Lit(c)$. 
But \,$\OR\{l\} \e \Res(\Sat(C))$, 
so there exists $c$ in $\Sat(C)$ such that \,$l \e \Lit(c)$. 
This contradiction shows that \,$\OR\{\} \notin \Res(\Sat(C))$. 

(\ref{C is satisfiable iff the empty clause is not in Res(C).}) 
This is well known from classical propositional logic. 

(\ref{Sat(C) is satisfiable.}) 
By parts (\ref{The empty clause is not in Res(Sat(C)).}) and 
(\ref{C is satisfiable iff the empty clause is not in Res(C).}) 
of this lemma, $\Sat(C)$ is satisfiable. 
Hence $\Res(\Sat(C))$ is satisfiable and so $\CorRes(\Sat(C))$ is satisfiable. 

\noindent \eop{EndProofLem\ref{Lem:Err,Sat}} 
\end{Lem} 

We say that a set $L$ of literals is \defn{contingent} iff $L$ is not empty and 
if $a$ is any atom then \,$\{a, \neg a\} \!\not\subseteq\! L$. 

\begin{Lem} \label{Lem:Ax} 
Suppose $(R,>)$ is a plausible description, \,$\Ax = \Ax(R)$, 
\,$\alpha \e \Alg$, $H$ is an $\alpha$-history, and $f$ is a formula. 
\begin{compactenum}[1)]
\item \label{Ax is satisfiable.} 
	$\Ax$ is satisfiable. 
\item \label{Axioms are contingent.} 
	Each axiom in $\Ax$ is contingent. 
\item \label{Axioms are simplified.} 
	Each axiom in $\Ax$ is either a literal \nl or $\OR L$ 
	where $L$ is a finite set of literals such that \,$|L| \!\geq\! 2$.  
\item \label{If R[f] is nonempty then Ax+f is satisfiable.} 
	If \,$R[f] \neq \{\}$\, then \,$\Ax \!\cup\! \{f\}$\, is satisfiable and 
	\,$\Ax \not\models \neg f$. 
\item \label{If (alpha,H) |- f then Ax+f is satisfiable.} 
	If \,$(\alpha,H) \vminus f$\, then \,$\Ax \!\cup\! \{f\}$\, is satisfiable 
	and \,$\Ax \not\models \neg f$. 
\end{compactenum}

\noindent \pf{Proof} 

Suppose $(R,>)$ is a plausible description, \,$\Ax = \Ax(R)$, 
\,$\alpha \e \Alg$, $H$ is an $\alpha$-history, and $f$ is a formula. 

(\ref{Ax is satisfiable.}) 
By Definition \ref{Defn:PlausibleDescription}(PD\ref{Ax=}), 
\,$\Ax = \CorRes(\Sat(\Ax))$. 
By Lemma \ref{Lem:Err,Sat}(\ref{Sat(C) is satisfiable.}), \nl
$\CorRes(\Sat(\Ax))$ is satisfiable. 
Hence $\Ax$ is satisfiable. 

(\ref{Axioms are contingent.}) 
By the definitions, 
\,$\Ax$ = $\CorRes(\Sat(\Ax))$ = $\SmpMinCtge(\Res(\Sat(\Ax)))$; 
so each axiom is either contingent or empty. 
But $\Ax$ is satisfiable, so each axiom is contingent. 

(\ref{Axioms are simplified.}) 
This follows from part (\ref{Axioms are contingent.}) and 
\,$\Ax = \Smp(C)$\, where $C$ is a set of clauses. 

(\ref{If R[f] is nonempty then Ax+f is satisfiable.}) 
Suppose \,$R[f] \neq \{\}$. 
By Definition \ref{Defn:subsets of R}(\ref{R'[f]}), 
there exists $r$ in $R$ such that \,$\Ax\!\cup\!\{c(r)\}$ is satisfiable 
and \,$\Ax\!\cup\!\{c(r)\} \models f$. 
Hence \,$\Ax \!\cup\! \{f\}$\, is satisfiable, 
and so \,$\Ax \not\models \neg f$. 

(\ref{If (alpha,H) |- f then Ax+f is satisfiable.}) 
Suppose \,$(\alpha,H) \vminus f$. 
By Definition \ref{Defn:|-}(I\ref{|-fact},I\ref{|-fml}), 
either \,$\Ax \models f$\, or \,$R^s_d[f] \neq \{\}$. 
If \,$\Ax \models f$\, then by part (\ref{Ax is satisfiable.}), 
\,$\Ax \!\cup\! \{f\}$\, is satisfiable. 
If \,$R^s_d[f] \neq \{\}$\, then by 
part (\ref{If R[f] is nonempty then Ax+f is satisfiable.}), 
\,$\Ax \!\cup\! \{f\}$\, is satisfiable. 
But \,$\Ax \!\cup\! \{f\}$\, is satisfiable implies \,$\Ax \not\models \neg f$. 

\noindent \eop{EndProofLem\ref{Lem:Ax}} 
\end{Lem} 
\begin{Lem} \label{Lem:Strict rules} 
Suppose $(R,>)$ is a plausible description, and \,$\Ax = \Ax(R)$. 
\begin{compactenum}[1)]
\item \label{A(r)=} 
	If \,$r \e R_s$\, then either \,$A(r) = \{\}$; \,or 
	\,$A(r) = \{l\}$, where $l$ is a literal; \nl 
	or \,$A(r) = \{\AND L\}$, where \,$|L| \!\geq\! 2$\, and $L$ is contingent. 
\item \label{If r is strict then Ax U A(r) |= c(r).} 
	If \,$r \e R_s$\, then \,$\Ax \!\cup\! A(r) \models c(r)$. 
\end{compactenum}

\noindent \pf{Proof} 

Suppose $(R,>)$ is a plausible description, and \,$\Ax = \Ax(R)$. 

(\ref{A(r)=}) 
By Lemma \ref{Lem:Ax}(\ref{Axioms are contingent.}), 
if \,$\OR L \e \Ax$\, then $\OR L$ is contingent and so $L$ is contingent. 
Hence, if \,$\{\} \!\subset\! K \!\subset\! L$\, 
then \,$L\!-\!K$\, is contingent. 
So \,$\nnot (L\!-\!K)$\, is contingent. 
Therefore the result holds for all $r$ in $\Rul(\Ax)$. 
So by Definition \ref{Defn:PlausibleDescription}(PD\ref{Rs=}), 
the result holds for all $r$ in $R_s$. 

(\ref{If r is strict then Ax U A(r) |= c(r).})
Take any $r$ in $\Rul(\Ax)$, and suppose $v$ is a valuation such that 
$v(\Ax \!\cup\! A(r))$ is the true truth value; that is, 
\,$v \models \Ax \!\cup\! A(r)$. 
Then either $r$ is \,$\{\} \!\strArr\! c$\, where \,$c \e \Ax$, 
or $r$ is \,$\{\smp(\AND\nnot(L\!-\!K))\} \!\strArr\! \smp(\OR K)$, where 
\,$\{\} \!\subset\! K \!\subset\! L$\, and \,$\OR L \e \Ax$. 
Now \,$v \models c$, so in the first case \,$v \models c(r)$. 
In the second case \,$v \models \OR L$, and \,$v \models \AND\nnot(L\!-\!K)$. 
Then for all \,$l \in L\!-\!K$, \,$v \models \nnot l$\, and so 
\,$v \not\models l$. 
But \,$L = K \cup (L\!-\!K)$. 
So \,$v \models \OR (K \cup (L\!-\!K))$\, and hence \,$v \models \OR K$. 
Thus \,$v \models c(r)$\, in the second case too. 
So the lemma holds for all $r$ in $\Rul(\Ax)$. 

Take any $r_0$ in $R_s \!-\! \Rul(\Ax)$, and suppose 
\,$v \models \Ax \!\cup\! A(r_0)$. 
Then $r_0$ is \,$A(r_0) \strArr \AND c(\Rul(\Ax,A(r_0)))$. 
For each $r$ in $\Rul(\Ax,A(r_0))$, $r$ is \,$A(r_0) \strArr c(r)$. 
By the previous paragraph, \,$v \models c(r)$. 
But this is true for every $r$ in $\Rul(\Ax,A(r_0))$ and 
so \,$v \models \AND c(\Rul(\Ax,A(r_0)))$. 

\noindent \eop{EndProofLem\ref{Lem:Strict rules}} 
\end{Lem} 
\begin{Lem} 
\label{Lem:R[.]} 
Suppose $(R_0,>)$ is a plausible description, $\Ax$ is its set of axioms, 
\,$R \!\subseteq\! R_0$, $f$ and $g$ are formulas, 
\,$\alpha \e \Alg$, and \,$\{r,s\} \!\subseteq\! R_0$. 
\begin{compactenum}[1)]
\item \label{R[.] subset R} 
	$R[f]\!\subseteq\!R$. 
\item \label{If R' subset R then R'[.] subset R[.].} 
	If \,$R'\!\subseteq\!R$\, then \,$R'[f]\!\subseteq\!R[f]$. 
\item \label{If f equiv g then R[f] = R[g].} 
	If \,$f \equiv g$\, then \,$R[f] = R[g]$. 
\item \label{If f is a fact then R[g] = R[f and g] and Foe...} 
	If \,$\Ax \models f$\, then \,$R[\AND\{f,g\}] = R[g]$\, and 
	\,$\Foe(\alpha,\AND \{f,g\},r) \subseteq \Foe(\alpha,g,r)$. 
\item \label{If Ax+f|=g then R[f] subseteq R[g], etc.} 
	If \,$\Ax\!\cup\!\{f\} \models g$\, then 
	(a)~$\Ax\!\cup\!\{\neg g\} \models \neg f$, 
	(b)~$R[f] \subseteq R[g]$, 
	(c)~$R[f;s] \subseteq R[g;s]$, 
	(d)~$R[\neg g] \subseteq R[\neg f]$, 
	(e)~$\Foe(\alpha,g,r) \subseteq \Foe(\alpha,f,r)$. 
\item \label{If Ax+f+g is unsat then R[f] subseteq R[comp g].} 
	If \,$\Ax\!\cup\!\{f,g\}$\, is unsatisfiable then 
	\,$R[f] \subseteq R[\neg g]$\, and \,$R[g] \subseteq R[\neg f]$. 
\end{compactenum}
\noindent \pf{Proof} 

(\ref{R[.] subset R}) 
By Definition 
\ref{Defn:subsets of R}(\ref{R'[f]}), \,$R[f]\!\subseteq\!R$. 

(\ref{If R' subset R then R'[.] subset R[.].}) 
Suppose \,$R'\!\subseteq\!R$. 
Then \,$R'[f] = \{r \e R' : \Ax\!\cup\!\{c(r)\}$ is satisfiable and 
$\Ax\!\cup\!\{c(r)\} \models f\}$ $\subseteq$ 
$\{r \e R : \Ax\!\cup\!\{c(r)\}$ is satisfiable and 
$\Ax\!\cup\!\{c(r)\} \models f\} = R[f]$. 

(\ref{If f equiv g then R[f] = R[g].}) 
Suppose \,$f \equiv g$. 
By Definition~\ref{Defn:subsets of R}(\ref{R'[f]}), 
\,$R[f]$ = $\{r \e R : \Ax\!\cup\!\{c(r)\}$ is satisfiable and 
$\Ax\!\cup\!\{c(r)\} \models f\}$ = 
$\{r \e R : \Ax\!\cup\!\{c(r)\}$ is satisfiable and 
$\Ax\!\cup\!\{c(r)\} \models g\}$ = $R[g]$. 

(\ref{If f is a fact then R[g] = R[f and g] and Foe...}) 
Suppose \,$\Ax \models f$. 
By Definition~\ref{Defn:subsets of R}(\ref{R'[f]}), \,$R[g]$ \nl
$= \{r \e R : \Ax\!\cup\!\{c(r)\}$ is satisfiable and 
	\,$\Ax\!\cup\!\{c(r)\} \models g\}$ \nl
$= \{r \e R : \Ax\!\cup\!\{c(r)\}$ is satisfiable and 
	\,$\Ax\!\cup\!\{c(r)\} \models f$\, and 
	\,$\Ax\!\cup\!\{c(r)\} \models g\}$ \nl
$= \{r \e R : \Ax\!\cup\!\{c(r)\}$ is satisfiable and 
	\,$\Ax\!\cup\!\{c(r)\} \models \AND\{f,g\} \}$ \nl
$= R[\AND\{f,g\}]$. 

Let Claim 1 be: \,$\Foe(\alpha,\AND \{f,g\},r) \subseteq \Foe(\alpha,g,r)$. \nl
If \,$\alpha \e \{\varphi,\pi'\}$\, or \,$r = \rse$\, 
then \,$\Foe(\alpha,\AND \{f,g\},r) = \{\}$. 
Hence Claim 1 holds. 

Suppose \,$\alpha = \psi'$. 
Then \,$\Foe(\psi',\AND \{f,g\},r) = \{s \e R[\neg \AND \{f,g\}] : s > r\}$\, 
and \,$\Foe(\psi',g,r) = \{s \e R[\neg g] : s > r\}$. 
Take any $s$ in $\Foe(\psi',\AND \{f,g\},r)$. 
Then \,$s \e R$, \,$\Ax \!\cup\! \{c(s)\}$\, is satisfiable, 
\,$\Ax \!\cup\! \{c(s)\} \models \neg \AND \{f,g\}$, and \,$s > r$. 
But \,$\Ax \models f$, so \,$\Ax \!\cup\! \{c(s)\} \models \neg g$\, 
and therefore \,$s \e \Foe(\psi',g,r)$. 
Hence Claim 1 holds. 

So suppose \,$\alpha \e \{\pi,\psi,\beta,\beta'\}$\, and \,$r \neq \rse$. 
Then 
\,$\Foe(\alpha,\AND \{f,g\},r) = \{s \e R[\neg \AND \{f,g\}] : s \not< r\}$\, 
and \,$\Foe(\alpha,g,r) = \{s \e R[\neg g] : s \not< r\}$. 
Take any $s$ in $\Foe(\alpha,\AND \{f,g\},r)$. 
Then \,$s \e R$, \,$\Ax \!\cup\! \{c(s)\}$\, is satisfiable, 
\,$\Ax \!\cup\! \{c(s)\} \models \neg \AND \{f,g\}$, and \,$s \not< r$. 
But \,$\Ax \models f$, so \,$\Ax \!\cup\! \{c(s)\} \models \neg g$\, 
and therefore \,$s \e \Foe(\alpha,g,r)$. 
Hence Claim 1 holds. 

Thus Claim 1 is proved. 

(\ref{If Ax+f|=g then R[f] subseteq R[g], etc.})
Suppose \,$\Ax\!\cup\!\{f\} \models g$. 
By Definition \ref{Defn:subsets of R}(\ref{R'[f]},\ref{R'[f;s]}), 
\,$R[f] = \{r \e R : \Ax\!\cup\!\{c(r)\}$ is satisfiable and 
$\Ax\!\cup\!\{c(r)\} \models f\}$\, and 
\,$R[f;s] = \{t \e R[f] : t > s\}$. \nl 
(a) Every valuation satisfies exactly one of $f$ or $\neg f$. 
Hence \,$\Ax\!\cup\!\{\neg g\} \models \neg f$. \nl 
(b) Take any $r$ in $R[f]$. 
Then \,$\Ax\!\cup\!\{c(r)\}$ is satisfiable and 
\,$\Ax\!\cup\!\{c(r)\} \models f$.  
Hence \,$\Ax\!\cup\!\{c(r)\} \models g$\, and so \,$r \e R[g]$. 
Thus \,$R[f] \subseteq R[g]$. \nl 
(c) Take any $t$ in $R[f;s]$. 
Then \,$t \e R[f]$\, and \,$t > s$. 
By part (b), \,$t \e R[g]$\, and so \,$t \e R[g;s]$. 
Thus \,$R[f;s] \subseteq R[g;s]$. \nl 
(d) This follows from parts (a) and (b). \nl 
(e) This follows from parts (a) and (c). 

(\ref{If Ax+f+g is unsat then R[f] subseteq R[comp g].}) 
Suppose \,$\Ax\!\cup\!\{f,g\}$\, is unsatisfiable. 
Take any $r$ in $R[f]$. 
Then \,$\Ax\!\cup\!\{c(r)\}$ is satisfiable and 
\,$\Ax\!\cup\!\{c(r)\} \models f$.  
Hence \,$\Ax\!\cup\!\{c(r)\} \models \neg g$. 
Therefore \,$r \e R[\neg g]$\, and so \,$R[f] \subseteq R[\neg g]$. 
By swapping $f$ and $g$ we get \,$R[g] \subseteq R[\neg f]$. 

\noindent \eop{EndProofLem\ref{Lem:R[.]}} 
\end{Lem} 

We need to extend the notation introduced in 
Definition \ref{Defn:T[alpha,H,x],T(alpha,H,x)}.

\begin{Defn} \label{Defn:T[],T()} 
Suppose \,$\calP = (R,>)$\, is a plausible theory, \,$\alpha \e \Alg$, 
$H$ is an $\alpha$-history, $F$ is a finite set of formulas, 
$f$ is a formula, \,$r \e R^s_d[f]$, and \,$s \e R[\neg f]$. 
\begin{compactenum}[1)]
\item Let $T[\alpha,H,F]$ be the evaluation tree of $\calP$ 
	whose root has the subject $(\alpha,H,F)$; and \nl
	let $T(\alpha,H,F)$ be the proof value of the root of $T[\alpha,H,F]$. 
\item Let $T[\alpha,H,f]$ be the evaluation tree of $\calP$ 
	whose root has the subject $(\alpha,H,f)$; and \nl 
	let $T(\alpha,H,f)$ be the proof value of the root of $T[\alpha,H,f]$. 
\item Let $T[\alpha,H,f,r]$ be the evaluation tree of $\calP$ 
	whose root has the subject $(\alpha,H,f,r)$; and \nl 
	let $T(\alpha,H,f,r)$ be the proof value of the root of $T[\alpha,H,f,r]$. 
\item Let $T[\alpha,H,f,r,s]$ be the evaluation tree of $\calP$ 
	whose root has the subject $(\alpha,H,f,r,s)$; and 
	let $T(\alpha,H,f,r,s)$ be the proof value of the root of 
	$T[\alpha,H,f,r,s]$. 
\item Let $T[-(\alpha,H,F)]$ be the evaluation tree of $\calP$ 
	whose root has the subject $-(\alpha,H,F)$; and 
	let $T(-(\alpha,H,F))$ be the proof value of the root of $T[-(\alpha,H,F)]$. 
\end{compactenum}
\end{Defn} 
\begin{Thm} [Theorem \ref{Thm:Decisiveness} Decisiveness]
\label{AppThm:Decisiveness} 
Suppose $\calP$ is a plausible theory, \,$\alpha \e \Alg$, 
$H$ is an $\alpha$-history, and $x$ is either a formula or a finite set of formulas. 
\begin{compactenum}[1)]
\item \label{T[alpha,H,f] is finite.} 
	$T[\alpha,H,x]$ has finitely many nodes. 
\item \label{T(alpha,H,f) = +1 or -1} 
	Either \,$T(\alpha,H,x) = +1$\, or \,$T(\alpha,H,x) = -1$\, 
	but not both. 
\end{compactenum}

\noindent \pf{Proof} 

(\ref{T[alpha,H,f] is finite.}) follows from 
Definition \ref{Defn:PlausibleTheory, PlausibleLogic}. 

(\ref{T(alpha,H,f) = +1 or -1}) follows from part 
(\ref{T[alpha,H,f] is finite.}) of this lemma and 
Definition \ref{Defn:EvaluationTree}. 

\noindent \eop{EndProofThm\ref{AppThm:Decisiveness}} 
\end{Thm} 
\begin{Thm}[Theorem \ref{Thm:P iff |- iff T} Notational Equivalence] 
\label{Thm:P()=+1 iff |- iff T()=+1} 
Suppose $(R,>)$ is a plausible theory, \,$\alpha \e \Alg$, 
$H$ is an $\alpha$-history, $F$ is a finite set of formulas, 
and $f$ is a formula. 
\begin{compactenum}[1)]
\item \label{equiv set} 
$P(\alpha,H,F) = +1$ \ \ iff \ \ $(\alpha,H) \vminus F$ \ \ 
iff \ \ $T(\alpha,H,F) = +1$. 

\item \label{equiv formula} 
$P(\alpha,H,f) = +1$ \ \ iff \ \ $(\alpha,H) \vminus f$ \ \ 
iff \ \ $T(\alpha,H,f) = +1$. 

\item \label{equiv For} 
Suppose \,$\Ax \not\models f$\, and \,$\alpha \neq \varphi$\, and 
\,$\alpha r \nte H$\, and \,$r \e R^s_d[f]$\, and $f$ is satisfiable. \nl
Then \,$\For(\alpha,H,f,r) = +1$ \nl
iff \,$T(\alpha,H,f,r) = +1$ \nl
iff \,$(\alpha,H$+$\alpha r) \vminus A(r)$\, and 
    \,$\forall s \e \Foe(\alpha,f,r)$, $\Dftd(\alpha,H,f,r,s) = +1$. 

\item \label{equiv Dftd} 
Suppose \,$\Ax \not\models f$\, and 
\,$\alpha \in \Alg \!-\! \{\varphi, \pi'\}$\, and 
\,$\alpha r \nte H$\, and \,$r \e R^s_d[f]$\, and $f$ is satisfiable and 
\,$s \e \Foe(\alpha,f,r)$. 
Then \,$\Dftd(\alpha,H,f,r,s) = +1$ \nl
iff \,$T(\alpha,H,f,r,s) = +1$ \nl
iff either \,$\exists t \e R^s_d[f;s]$\, such that 
 		\,$\alpha t \nte H$\, and \,$(\alpha,H$+$\alpha t) \vminus A(t)$; \nl
\hspace*{2.5em}
or \,$\alpha's \nte H$\, and \,$(\alpha',H$+$\alpha's) \not\vminus A(s)$. 
\end{compactenum}

\noindent \pf{Proof} 

Let \,$\calP = (R,>)$\, be a plausible theory. 
The proof is by induction on the number of nodes in 
an evaluation tree of $\calP$. 
Let $p$ be the only node of an evaluation tree of $\calP$. 

If $p$ satisfies T\ref{Tset} then the subject of $p$ is $(\alpha,H,\{\})$ 
and the proof value of $p$ is $+1$. 
So \,$T(\alpha,H,\{\}) = +1$. 
By P\ref{Pset}, $P(\alpha,H,\{\})$ = $\min\{\}$ = $+1$. 
By I\ref{|-set}, $(\alpha,H) \vminus \{\}$. 

If $p$ satisfies T\ref{Tfact} then the subject of $p$ is $(\alpha,H,f)$ 
and the proof value of $p$ is $+1$. 
So \,$T(\alpha,H,f) = +1$. 
By P\ref{Pfact}, $P(\alpha,H,f) = +1$. 
By I\ref{|-fact}, $(\alpha,H) \vminus f$. 

If $p$ satisfies T\ref{Tfml} then the subject of $p$ is $(\alpha,H,f)$ 
and the proof value of $p$ is $-1$. 
So \,$T(\alpha,H,f) = -1$. 
Let \,$S(p) = \{(\alpha,H,f,r) : \alpha r \nte H$\, and \,$r \e R^s_d[f]\}$. 
By T\ref{Tfml}, $S(p)$ is empty. 
By P\ref{Pfml}, \,$P(\alpha,H,f)$ 
= $\max\{\For(\alpha,H,f,r) : (\alpha,H,f,r) \e S(p)\}$ = $\max\{\}$ = $-1$. 
Since $S(p)$ is empty, for all $r$ in $R^s_d[f]$, \,$\alpha r \e H$. 
Hence I\ref{|-fml}.\ref{|-For} fails and so \,$(\alpha,H) \not\vminus f$. 

Since $p$ has no children, 
$p$ does not satisfy T\ref{TFor} or T\ref{Tminus}. 

If $p$ satisfies T\ref{TDftd} then the subject of $p$ is $(\alpha,H,f,r,s)$ 
and the proof value of $p$ is $-1$. 
So \,$T(\alpha,H,f,r,s) = -1$. 
Let \,$S(p)$ = $\{(\alpha,H$+$\alpha t,A(t)) : 
\alpha t \nte H$\, and \,$t \e R^s_d[f;s]\}$ $\cup$ 
$\{-(\alpha',H$+$\alpha's,A(s)) : \alpha's \nte H\}$. 
Also let 
\,$S' = \{P(\alpha,H$+$\alpha t,A(t)) : 
\alpha t \nte H$\, and \,$t \e R^s_d[f;s]\}$ $\cup$ 
$\{-P(\alpha', H$+$\alpha's, A(s)) : \alpha's \nte H\}$. \nl 
By T\ref{TDftd}, $S(p)$ is empty, and so $S'$ is also empty. 
By P\ref{PDftd}, \,$\Dftd(\alpha,H,f,r,s)$ = $\max S'$ = $\max\{\}$ = $-1$. 
Since \,$S(p) = \{\}$, we have for each $t$ in $R^s_d[f;s]$, $\alpha t \e H$; 
and \,$\alpha's \e H$. 
Hence the last characterisation of \,$\Dftd(\alpha,H,f,r,s) = +1$\, 
in part (\ref{equiv Dftd}) is false. 

Thus the result holds for all evaluation trees of $\calP$ that 
have only the root node. 

Take any positive integer $n$. 
We shall denote the following inductive hypothesis by IndHyp. 
Suppose the result holds for all evaluation trees of $\calP$ that have less than 
$n\!+\!1$ nodes. 
Let $T$ be an evaluation tree of $\calP$ that has $n\!+\!1$ nodes and 
let $p$ be the root of $T$. 
Then $p$ has at least one child. 

If $p$ satisfies T\ref{Tset} then the subject of $p$ is $(\alpha,H,F)$. 
By T\ref{Tset}, IndHyp, P\ref{Pset}, and I\ref{|-set}, 
\begin{tabular}{@{}l@{ \,}ll}
$T(\alpha,H,F) = +1$ & iff \ for all $f$ in $F$, \,$T(\alpha,H,f) = +1$ & \\
 & iff \ for all $f$ in $F$, \,$P(\alpha,H,f) = +1$ & 
	iff \ $P(\alpha,H,F) = +1$ \\ 
 & iff \ for all $f$ in $F$, \,$(\alpha,H) \vminus f$ & 
	iff \ $(\alpha,H) \vminus F$. \\
\end{tabular}  

Since $p$ has a child, $p$ does not satisfy T\ref{Tfact}. 

If $p$ satisfies T\ref{Tfml} then the subject of $p$ is $(\alpha,H,f)$. 
Let 
$S(p) = \{(\alpha,H,f,r) : \alpha r \nte H$\, and \,$r \e R^s_d[f]\}$. 
By T\ref{Tfml}, IndHyp, P\ref{Pfml}, and I\ref{|-fml}, 
\,$T(\alpha,H,f) = +1$\, 
\begin{compactitem}[iff]
\item there exists $(\alpha,H,f,r)$ in $S(p)$ such that 
\,$T(\alpha,H,f,r) = +1$\, 
 
\item there exists $(\alpha,H,f,r)$ in $S(p)$ such that 
\,$\For(\alpha,H,f,r) = +1$ 

\item $P(\alpha,H,f) = +1$.   

\item there exists $(\alpha,H,f,r)$ in $S(p)$ such that 
\,$\For(\alpha,H,f,r) = +1$ 

\item there exists $(\alpha,H,f,r)$ in $S(p)$ such that 
\,$(\alpha,H$+$\alpha r) \vminus A(r)$\, and 
\,$\forall s \e \Foe(\alpha,f,r)$, $\Dftd(\alpha,H,f,r,s) = +1$

\item there exists $(\alpha,H,f,r)$ in $S(p)$ such that 
\,$(\alpha,H$+$\alpha r) \vminus A(r)$\, and 
\,$\forall s \e \Foe(\alpha,f,r)$, \nl 
either \,$\exists t \e R^s_d[f;s]$\, such that \,$\alpha t \nte H$\, and 
\,$(\alpha,H$+$\alpha t) \vminus A(t)$; \nl 
or \,$\alpha's \nte H$\, and \,$(\alpha',H$+$\alpha's) \not\vminus A(s)$

\item $\exists r \e R^s_d[f]$\, such that 
I\ref{|-fml}.\ref{|-For} and I\ref{|-fml}.\ref{|-Against} 

\item $(\alpha,H)\vminus f$. 
\end{compactitem} 

\smallskip

If $p$ satisfies T\ref{TFor} then the subject of $p$ is $(\alpha,H,f,r)$. 
By T\ref{TFor}, IndHyp, and P\ref{PFor}, \nl
\,$T(\alpha,H,f,r) = +1$\, 
\begin{compactitem}[iff]
\item $T(\alpha,H$+$\alpha r,A(r)) = +1$\, and 
	\,$\forall s \e \Foe(\alpha,f,r)$, \,$T(\alpha,H,f,r,s) = +1$\,  
\item $(\alpha,H$+$\alpha r) \vminus A(r)$\, and 
	\,$\forall s \e \Foe(\alpha,f,r)$, \,$\Dftd(\alpha,H,f,r,s) = +1$ 
\item $P(\alpha,H$+$\alpha r,A(r)) = +1$\, and 
	\,$\forall s \e \Foe(\alpha,f,r)$, \,$\Dftd(\alpha,H,f,r,s) = +1$ 
\item $\For(\alpha,H,f,r) = +1$. 
\end{compactitem}

\smallskip

If $p$ satisfies T\ref{TDftd} then the subject of $p$ is $(\alpha,H,f,r,s)$. 
By T\ref{TDftd}, IndHyp, T\ref{Tminus}, T\ref{Tvalue}, and P\ref{PDftd}, 
\,$T(\alpha,H,f,r,s) = +1$\, 
\begin{compactitem}[iff]
\item either \,$\exists t \e R^s_d[f;s]$\, such that 
\,$\alpha t \nte H$\, and \,$T(\alpha,H$+$\alpha t,A(t)) = +1$; \nl
or \,$\alpha's \nte H$\, and \,$T(-(\alpha',H$+$\alpha's,A(s))) = +1$ 

\item either \,$\exists t \e R^s_d[f;s]$\, such that 
\,$\alpha t \nte H$\, and \,$P(\alpha,H$+$\alpha t,A(t)) = +1$; \nl
or \,$\alpha's \nte H$\, and \,$T(\alpha',H$+$\alpha's,A(s)) = -1$ 

\item either \,$\exists t \e R^s_d[f;s]$\, such that 
\,$\alpha t \nte H$\, and \,$P(\alpha,H$+$\alpha t,A(t)) = +1$; \nl
or \,$\alpha's \nte H$\, and \,$P(\alpha',H$+$\alpha's, A(s)) = -1$ 

\item $\Dftd(\alpha,H,f,r,s) = +1$. 
\end{compactitem} 

\smallskip

If $p$ satisfies T\ref{Tminus} then the subject of $p$ is $-(\alpha',H,F)$. 
By T\ref{Tminus}, T\ref{Tvalue}, and IndHyp, \nl
\ $T(-(\alpha',H,F)) = +1$ \ iff \ $T(\alpha',H,F) = -1$ \ iff 
\ $P(\alpha',H,F) = -1$ \ iff \ $(\alpha',H) \not\vminus F$. 

Thus the theorem is proved by induction. 

\noindent \eop{EndProofThm\ref{Thm:P()=+1 iff |- iff T()=+1}} 
\end{Thm} 

\section{Proof of Theorems \ref{Thm:Plausible Conjunction}, 
\ref{Thm:Right Weakening}, \ref{Thm:Consistency}, and \ref{Thm:Truth values}}

\begin{Thm} [Theorem \ref{Thm:Plausible Conjunction} Plausible Conjunction] 
\label{AppThm:Plausible Conjunction} \raggedright 

Suppose $(R,>)$ is a plausible description, \,$\Ax = \Ax(R)$, 
\,$\alpha \e \Alg$, $H$ is an $\alpha$-history, and 
$f$ and $g$ are both formulas. 
If \,$\Ax \models f$\, and \,$(\alpha,H) \vminus g$\, 
then \,$(\alpha,H) \vminus \AND \{f,g\}$. 

\noindent \pf{Proof} 

Suppose $(R,>)$ is a plausible description, \,$\Ax = \Ax(R)$, 
\,$\alpha \e \Alg$, $H$ is an $\alpha$-history, and 
$f$ and $g$ are both formulas. 
Further suppose that \,$\Ax \models f$\, and \,$(\alpha,H) \vminus g$. 
We shall use Definition \ref{Defn:|-}. 

Since \,$(\alpha,H) \vminus g$, by Definition \ref{Defn:|-}(I\ref{|-fml}), 
there exists $r$ in $R^s_d[g]$ such that I\ref{|-fml}.\ref{|-For} and 
I\ref{|-fml}.\ref{|-Against} both hold. 
By Lemma~\ref{Lem:R[.]}(\ref{If f is a fact then R[g] = R[f and g] and Foe...}), 
there exists $r$ in $R^s_d[\AND \{f,g\}]$ such that 
I\ref{|-fml}.\ref{|-For} and I\ref{|-fml}.\ref{|-Against} both hold. 
Thus \,$(\alpha,H) \vminus \AND \{f,g\}$. 

\noindent \eop{EndProofTheorem\ref{AppThm:Plausible Conjunction}} 
\end{Thm}  
\begin{Thm} [Theorem \ref{Thm:Right Weakening} Right Weakening] 
\label{AppThm:Right Weakening}
Suppose $(R,>)$ is a plausible description, \,$\Ax = \Ax(R)$, 
\,$\alpha \e \Alg$, $H$ is an $\alpha$-history, and 
$f$ and $g$ are both formulas. 
\begin{compactenum}[1)]
\item \label{App:Strong Right Weakening} 
	If \,$(\alpha,H) \vminus f$\, and \,$\Ax\!\cup\!\{f\} \models g$\, 
	then \,$(\alpha,H) \vminus g$. [Strong Right Weakening]
\item \label{App:Right Weakening} 
	If \,$(\alpha,H) \vminus f$\, and \,$f \models g$\, 
	then \,$(\alpha,H) \vminus g$. [Right Weakening]
\item \label{App:Modus Ponens for strict rules} 
	If \,$A \!\strArr\! g \in R_s$\, and \,$(\alpha,H) \vminus A$\, 
	then \,$(\alpha,H) \vminus g$. [Modus Ponens for strict rules]
\end{compactenum}

\noindent \pf{Proof} 

Suppose $(R,>)$ is a plausible description, \,$\Ax = \Ax(R)$, 
\,$\alpha \e \Alg$, $H$ is an $\alpha$-history, and 
$f$ and $g$ are both formulas. 
Further suppose that \,$(\alpha,H) \vminus f$. 
We shall use Definition \ref{Defn:|-}. 

(\ref{App:Strong Right Weakening}) 
Suppose \,$\Ax\!\cup\!\{f\} \models g$. \nl 
If \,$\Ax \models f$\, then \,$\Ax \models g$\, and 
so by I\ref{|-fact}, \,$(\alpha,H) \vminus g$. 
So suppose \,$\Ax \not\models f$. 
Then \,$\alpha \neq \varphi$. 

Since \,$(\alpha,H) \vminus f$, by I\ref{|-fml}.\ref{|-For} for $f$, 
\,$\exists r_0 \e R^s_d[f]$\, such that 
\,$\alpha r_0 \nte H$\, and \,$(\alpha,H$+$\alpha r_0) \vminus A(r_0)$. 
By Lemma \ref{Lem:R[.]}(\ref{If Ax+f|=g then R[f] subseteq R[g], etc.})(b), 
\,$R^s_d[f] \subseteq R^s_d[g]$. 
Hence \,$r_0 \e R^s_d[g]$. 
So I\ref{|-fml}.\ref{|-For} holds for $g$. 

By Lemma \ref{Lem:R[.]}(\ref{If Ax+f|=g then R[f] subseteq R[g], etc.})(d,e), 
\,$R[\neg g] \subseteq R[\neg f]$\, and 
\,$\Foe(\alpha,g,r_0) \subseteq \Foe(\alpha,f,r_0)$. 
By Lemma \ref{Lem:R[.]}(\ref{If Ax+f|=g then R[f] subseteq R[g], etc.})(b,c), 
\,$R^s_d[f] \subseteq R^s_d[g]$\, and 
\,$R^s_d[f;s] \subseteq R^s_d[g;s]$. 

Now take any $s_0$ in $\Foe(\alpha,g,r_0)$. 
Then \,$s_0 \e \Foe(\alpha,f,r_0)$. 
If I\ref{|-fml}.\ref{|-Against}.\ref{TeamDefeat} holds for $f$ then 
\,$\exists t_0 \e R^s_d[f;s_0]$\, such that \,$\alpha t_0 \nte H$\, and 
\,$(\alpha,H$+$\alpha t_0) \vminus A(t_0)$. 
Hence \,$t_0 \e R^s_d[g;s_0]$\, and so 
I\ref{|-fml}.\ref{|-Against}.\ref{TeamDefeat} holds for $g$. 
If I\ref{|-fml}.\ref{|-Against}.\ref{DisableByNoProof} holds for $f$ then 
\,$\alpha' s_0 \nte H$\, and \,$(\alpha',H$+$\alpha' s_0) \not\vminus A(s_0)$. 
Hence I\ref{|-fml}.\ref{|-Against}.\ref{DisableByNoProof} holds for $g$. 

Thus I\ref{|-fml}.\ref{|-Against} holds for $g$ and so \,$(\alpha,H) \vminus g$. 

(\ref{App:Right Weakening}) Suppose \,$f \models g$. \nl 
Since \,$f \models g$, we have \,$\Ax\!\cup\!\{f\} \models g$. 
So by part (\ref{Strong Right Weakening}), \,$(\alpha,H) \vminus g$. 

(\ref{App:Modus Ponens for strict rules}) 
Suppose \,$A \!\strArr\! g \in R_s$\, and \,$(\alpha,H) \vminus A$. \nl
By Lemma \ref{Lem:Strict rules}(\ref{A(r)=}), either \,$A = \{\}$, or 
\,$A = \{a\}$\, where $a$ is formula. 

Case 1: \,$A = \{\}$. \nl 
By Definition \ref{Defn:PlausibleDescription}, \,$g = \AND \Ax$\, and 
so \,$\Ax \models g$. 
By Definition \ref{Defn:|-}(I\ref{|-fact}), \,$(\alpha,H) \vminus g$. 

Case 2: \,$A = \{a\}$\, where $a$ is formula. \nl 
By Definition \ref{Defn:|-}(I\ref{|-set}), \,$(\alpha,H) \vminus a$. 
By Lemma \ref{Lem:Strict rules}(\ref{If r is strict then Ax U A(r) |= c(r).}), 
\,$\Ax\!\cup\!\{a\} \models g$. \nl
So by part (\ref{Strong Right Weakening}), 
\,$(\alpha,H) \vminus g$. 

\noindent \eop{EndProofThm\ref{AppThm:Right Weakening}} 
\end{Thm}  
\begin{Defn} \label{Defn:H(alpha := pi')} \raggedright 
Suppose $H$ is an $\alpha$-history. 
If \,$\alpha = \varphi$\, then define \,$H(\varphi\!:=\!\pi')$\, 
to be the sequence formed from $H$ by just replacing each $\varphi$ by $\pi'$. 
If \,$\alpha \e \{\pi,\pi',\psi,\psi',\beta,\beta'\}$\, 
then define \,$H(\alpha\!:=\!\pi')$\, to be the sequence formed from $H$ 
by just replacing each $\alpha$ by $\pi'$, and each $\alpha'$ by $\pi$. 
\end{Defn} 

It is clear that \,$H(\alpha\!:=\!\pi')$\, is a $\pi'$-history. 
\begin{Lem} 
\label{Lem:If alpha|-x then pi'|-x.} 
\raggedright \parindent = 1.5em
Suppose \,$\calP = (R,>)$\, is a plausible theory, $\alpha \e \Alg$, 
$H$ is an $\alpha$-history, and 
$x$ is either a formula or a finite set of formulas. 
If \,$(\alpha,H) \vminus x$\, then \,$(\pi',H(\alpha\!:=\!\pi')) \vminus x$. 
Hence \,$\calP(\alpha) \!\subseteq\! \calP(\pi')$. 

\noindent \pf{Proof} 

Suppose $(R,>)$ is a plausible theory. 
Let $Y(n)$ denote the following conditional statement. \nl 
``If \,$\alpha \e \Alg$, $H$ is an $\alpha$-history, $f$ is a formula, 
$F$ is a finite set of formulas, \,$x \e \{F, f\}$, \,$T(\alpha,H,x) = +1$, and 
\,$|T[\alpha,H,x]| \leq n$\, then \,$T(\pi',H(\alpha\!:=\!\pi'),x) = +1$." 

By Theorem \ref{Thm:P()=+1 iff |- iff T()=+1}(\ref{equiv set},\ref{equiv formula}), and 
Theorem \ref{AppThm:Decisiveness}(\ref{T[alpha,H,f] is finite.}), 
it suffices to prove $Y(n)$ by induction on $n$. 

Suppose \,$n = 1$. 
Let the antecedent of $Y(1)$ hold. 
Let the only node of $T[\alpha,H,x]$ be $p_0$. 
Let the root of $T[\pi',H(\alpha\!:=\!\pi'),x]$ be $q_0$. 

If $p_0$ satisfies T\ref{Tset} then \,$x = \{\}$\, and so 
$q_0$ satisfies T\ref{Tset}. 
So by T\ref{Tset}, $T[\pi',H(\alpha\!:=\!\pi'),\{\}]$ has only one node and 
\,$T(\pi',H(\alpha\!:=\!\pi'),\{\}) = +1$. 
If $p_0$ satisfies T\ref{Tfact} then \,$x = f$\, and \,$\Ax \models f$. 
So $q_0$ satisfies T\ref{Tfact} and hence 
\,$T(\pi',H(\alpha\!:=\!\pi'),f) = +1$. 
Since $p_0$ has no children and the proof value of $p_0$ is $+1$, 
$p_0$ does not satisfy T\ref{Tfml}. 
Since the subject of $p_0$ is $(\alpha,H,x)$, $p_0$ does not satisfy 
T\ref{TFor}, or T\ref{TDftd}, or T\ref{Tminus}. 
Thus the base case holds. 

Take any positive integer $n$. 
Suppose $Y(n)$ is true. 
We shall prove $Y(n\!+\!1)$. 

Suppose the antecedent of $Y(n\!+\!1)$ holds and that 
\,$|T[\alpha,H,x]| = n\!+\!1$. 
Then \,$T(\alpha,H,x) = +1$. 
Let $p_0$ be the root of $T[\alpha,H,x]$ 
and $q_0$ be the root of $T[\pi',H(\alpha\!:=\!\pi'),x]$. 

If $p_0$ satisfies T\ref{Tset} then \,$x = F$. 
We see that $q_0$ also satisfies T\ref{Tset}. 
So \,$t(p_0) = ((\alpha,H,F), \min, +1)$\, and 
\,$t(q_0) = ((\pi',H(\alpha\!:=\!\pi'),F), \min, w_0)$, where 
\,$w_0 = T(\pi',H(\alpha\!:=\!\pi'),F) \in \{+1,-1\}$. 
Let $\{p_{\!f} : f \e F\}$ be the set of children of $p_0$ and 
$\{q_{\!f} : f \e F\}$ be the set of children of $q_0$. 
Let $f$ be any formula in $F$. 
Then the subject of $p_{\!f}$ is $(\alpha,H,f)$, and the subject of 
$q_{\!f}$ is $(\pi',H(\alpha\!:=\!\pi'),f)$. 
Also \,$|T[\alpha,H,f]| \leq n$\, and the proof value of $p_{\!f}$ is $+1$ 
because $p_0$ is a min node with proof value $+1$. 
So by $Y(n)$ the proof value of $q_{\!f}$ is $+1$. 
But this is true for each $f$, so \,$w_0 = +1$, as required. 

Since $p_0$ has a child, $p_0$ does not satisfy T\ref{Tfact}. 

If $p_0$ satisfies T\ref{Tfml} then \,$x = f$. 
We see that $q_0$ also satisfies T\ref{Tfml}. 
So \,$t(p_0) = ((\alpha,H,f), \max, +1)$\, and 
\,$t(q_0) = ((\pi',H(\alpha\!:=\!\pi'),f), \max, w_0)$, where 
\,$w_0 = T(\pi',H(\alpha\!:=\!\pi'),f) \in \{+1, -1\}$. 

We shall adopt the following naming conventions. 
Each non-root node of $T[\alpha,H,f]$ is denoted by $p_l(\#,y)$ where 
$l$ is the level of the node, $\#$ is the number in 
$[\ref{Tset}..\ref{Tminus}]$ such that the node satisfies T\#, and 
$y$ is a rule, or a formula, or a set, which distinguishes siblings. 
The proof value of $p_l(\#,y)$ will be denoted by $v_l(\#,y)$. 
For non-root nodes in $T[\pi',H(\alpha\!:=\!\pi'),f]$ we shall use 
$q_l(\#,y)$, and its proof value will be denoted by $w_l(\#,y)$. 

Let the set of children of $p_0$ be 
\,$\{p_1(\ref{TFor},r) : \alpha r \nte H$\, and \,$r \e R^s_d[f]\}$, 
where the tags of these children are: 
\,$t(p_1(\ref{TFor},r))$ = $((\alpha,H,f,r),\min,v_1(\ref{TFor},r))$. 
So \,$+1 = \max\{v_1(\ref{TFor},r) : \alpha r \nte H$\, and 
\,$r \e R^s_d[f]\}$. 

Let the set of children of $q_0$ be 
\,$\{q_1(\ref{TFor},r) : \pi' r \nte H(\alpha\!:=\!\pi')$\, and 
\,$r \e R^s_d[f]\}$, where the tags of these children are: 
\,$t(q_1(\ref{TFor},r))$ $=$ 
$((\pi',H(\alpha\!:=\!\pi'),f,r),\min,w_1(\ref{TFor},r))$. 
So \,$w_0 = \max\{w_1(\ref{TFor},r) : \pi' r \nte H(\alpha\!:=\!\pi')$\, and 
\,$r \e R^s_d[f]\}$. 

From above there exists $r_0$ in $R^s_d[f]$ such that 
\,$\alpha r_0 \nte H$\, and \,$v_1(\ref{TFor},r_0) = +1$. 
By Definition \ref{Defn:H(alpha := pi')}, 
if \,$\pi' r_0 \e H(\alpha\!:=\!\pi')$\, then \,$\alpha r_0 \e H$. 
So if \,$\alpha r_0 \nte H$\, then \,$\pi' r_0 \nte H(\alpha\!:=\!\pi')$. 
Hence $q_1(\ref{TFor},r_0)$ exists. 
We shall show that \,$w_1(\ref{TFor},r_0) = +1$\, and hence that \,$w_0 = +1$, as required. 

Because $p_1(\ref{TFor},r_0)$ is a min node whose proof value is $+1$, 
the proof value of every child of $p_1(\ref{TFor},r_0)$ must be $+1$. 
$p_2(\ref{Tset},r_0)$ is a child of $p_1(\ref{TFor},r_0)$ such that 
\,$t(p_2(\ref{Tset},r_0)) = ((\alpha,H$+$\alpha r_0,A(r_0)),\min,+1)$. 

The only child of $q_1(\ref{TFor},r_0)$ is $q_2(\ref{Tset},r_0)$ where 
\,$t(q_2(\ref{Tset},r_0)) = ((\pi',H(\alpha\!:=\!\pi')$+$\pi' r_0,A(r_0)),\min,w_2(\ref{Tset},r_0))$. 
So \,$w_1(\ref{TFor},r_0)$ = $w_2(\ref{Tset},r_0)$. 

Since \,$|T[\alpha,H$+$\alpha r_0,A(r_0)]| \leq n$\, and 
\,$T(\alpha,H$+$\alpha r_0,A(r_0)) = +1$, by $Y(n)$ we have 
\,$T(\pi',H(\alpha\!:=\!\pi')$+$\pi' r_0,A(r_0)) = w_2(\ref{Tset},r_0) = +1$. 
Hence \,$w_1(\ref{TFor},r_0) = +1$\, and so \,$w_0 = +1$, as required. 

Since the subject of $p_0$ is $(\alpha,H,x)$, $p_0$ does not satisfy 
T\ref{TFor}, or T\ref{TDftd}, or T\ref{Tminus}. 

Thus $Y(n)$, and hence the lemma, is proved by induction. 

\noindent 
\eop{EndProofLem\ref{Lem:If alpha|-x then pi'|-x.}} 
\end{Lem} 
\begin{Defn} \label{Defn:History of a node} \raggedright 
Suppose \,$\calPd = (R,>)$\, is a plausible description, \,$\alpha \e \Alg$, 
$H$ is an $\alpha$-history, $F$ is a finite set of formulas, $f$ is a formula, 
$r$ and $s$ are any rules, $T$ is an evaluation tree of $\calPd$, and 
$p$ is any node of $T$. 
If \,$\Subj(p)$ $\in$ $\{(\alpha,H,F), (\alpha,H,f), (\alpha,H,f,r), 
(\alpha,H,f,r,s), -(\alpha,H,F)\}$\, 
then the \defn{history of $p$}, $\Hist(p)$, is defined by \,$\Hist(p) = H$, and 
the \defn{algorithm of $p$}, $\alg(p)$, is defined by \,$\alg(p) = \alpha$. 
\end{Defn} 
\begin{Defn} \label{Defn:(alpha:alpha')}  
Suppose \,$\{\alpha, \lambda\} \!\subseteq\! \Alg$, $H$ is a $\lambda$-history, 
and $T$ is an evaluation tree of some plausible theory. 
If \,$\lambda \nte \{\alpha,\alpha'\}$\, then define 
\,$\lambda(\alpha\!:\!\alpha') = \lambda$; else define 
\,$\alpha(\alpha\!:\!\alpha') = \alpha'$\, and 
\,$\alpha'(\alpha\!:\!\alpha') = \alpha$. 
If \,$H = (\lambda_1r_1, ..., \lambda_nr_n)$\, then define 
\,$H(\alpha\!:\!\alpha') = 
(\lambda_1(\alpha\!:\!\alpha')r_1, ..., \lambda_n(\alpha\!:\!\alpha')r_n)$. 
Define $T(\alpha\!:\!\alpha')$ to be the tree formed from $T$ by 
only changing the subject of each node as follows. 
For each node $p$ of $T$ replace $\alg(p)$ by $\alg(p)(\alpha\!:\!\alpha')$, 
and replace $\Hist(p)$ by $\Hist(p)(\alpha\!:\!\alpha')$. 
\end{Defn} 
\begin{Defn} \label{Defn:equivalent,isomorphic algorithms}  
Suppose \,$\alpha \e \Alg$. 
$\alpha$ is \defn{isomorphic} to $\alpha'$, \,$\alpha \simeq \alpha'$, 
iff for each plausible theory $\calP$, 
if $T$ is an evaluation tree of $\calP$ then 
$T(\alpha\!:\!\alpha')$ is an evaluation tree of $\calP$. 
\end{Defn} 

It should be clear that if \,$\alpha \simeq \alpha'$\, 
then for each plausible theory $\calP$, \,$\calP(\alpha) = \calP(\alpha')$. 

\begin{Lem} 
\label{Lem:beta is isomorphic to beta'}
$\beta \simeq \beta'$. 
Hence for each plausible theory $\calP$, \,$\calP(\beta) = \calP(\beta')$. 

\noindent \pf{Proof}

Suppose \,$\alpha \e \{\beta,\beta'\}$\, and 
\,$\calP = (R,>)$\, is a plausible theory. 
Let $Y(n)$ denote the following conditional statement. 
``If $T$ is an evaluation tree of $\calP$ and \,$|T| \leq n$\, 
then \,$T(\alpha\!:\!\alpha')$ is an evaluation tree of $\calP$." 
By Definition \ref{Defn:PlausibleTheory, PlausibleLogic}, 
it suffices to prove $Y(n)$ by induction on $n$. 
By Definitions \ref{Defn:EvaluationTree} and \ref{Defn:T[],T()}, 
if $T$ is an evaluation tree of $\calP$ then 
either \,$T = T[\alpha,H, ...]$\, or \,$T = T[-(\alpha',H,F)]$. 
So if \,$T(\alpha\!:\!\alpha')$ is an evaluation tree of $\calP$ then 
either \,$T(\alpha\!:\!\alpha') = T[\alpha',H(\alpha\!:\!\alpha'), ...]$\, 
or \,$T(\alpha\!:\!\alpha') = T[-(\alpha,H(\alpha\!:\!\alpha'),F)]$. 

Suppose \,$n=1$\, and the antecedent of $Y(1)$ holds. 
Let the root of $T$ be $p_0$. 
Since \,$\alpha \e \{\beta,\beta'\}$, 
without loss of generality we can suppose \,$\alg(p_0) = \alpha$. 
Let the root of $T(\alpha\!:\!\alpha')$ be $q_0$. 
Since $p_0$ has no children, $q_0$ has no children. \nl
If $p_0$ satisfies T\ref{Tset} then let \,$\Subj(p_0) = (\alpha,H,\{\})$. 
Hence \,$\Subj(q_0) = (\alpha',H(\alpha\!:\!\alpha'),\{\})$\, 
and so $q_0$ satisfies T\ref{Tset}. 
Thus $T(\alpha\!:\!\alpha')$ is an evaluation tree of $\calP$. \nl
If $p_0$ satisfies T\ref{Tfact} then let \,$\Subj(p_0) = (\alpha,H,f)$. 
Hence \,$\Subj(q_0) = (\alpha',H(\alpha\!:\!\alpha'),f)$\, 
and so $q_0$ satisfies T\ref{Tfact}. 
Thus $T(\alpha\!:\!\alpha')$ is an evaluation tree of $\calP$. \nl
If $p_0$ satisfies T\ref{Tfml} then let \,$\Subj(p_0) = (\alpha,H,f)$. 
Hence \,$\Subj(q_0) = (\alpha',H(\alpha\!:\!\alpha'),f)$. 
Since $p_0$ has no children, \,$S(p_0) = \{\}$. 
So if \,$r \e R^s_d[f]$\, then \,$\alpha r \e H$. 
Now \,$\alpha r \e H$\, iff \,$\alpha' r \e H(\alpha\!:\!\alpha')$. 
Hence \,$S(q_0) = \{\}$\, and so $q_0$ satisfies T\ref{Tfml}. 
Thus $T(\alpha\!:\!\alpha')$ is an evaluation tree of $\calP$. \nl
Since $p_0$ and $q_0$ have no children, 
$p_0$ and $q_0$ satisfy neither T\ref{TFor} nor T\ref{Tminus}. \nl 
If $p_0$ satisfies T\ref{TDftd} then let \,$\Subj(p_0) = (\alpha,H,f,r,s)$. 
Hence \,$\Subj(q_0) = (\alpha',H(\alpha\!:\!\alpha'),f,r,s)$. 
Since $p_0$ has no children, \,$S(p_0) = \{\}$\, and so 
\,$S(p_0,\alpha) = \{\}$. 
Hence if \,$t \e R^s_d[f;s]$\, then \,$\alpha t \e H$. 
Now \,$\alpha t \e H$\, iff \,$\alpha' t \e H(\alpha\!:\!\alpha')$. \nl
Also \,$\alpha's \e H$. 
Hence \,$\alpha s \e H(\alpha\!:\!\alpha')$. 
So \,$S(q_0,\alpha') = \{\}$\, and so \,$S(q_0) = \{\}$. 
Thus $q_0$ satisfies T\ref{TDftd} and so 
$T(\alpha\!:\!\alpha')$ is an evaluation tree of $\calP$. 

All cases have been considered and so $Y(1)$ holds. 

If $T$ is any tree and $p$ is any node of $T$ for which $\Subj(p)$ is defined, 
then define the set $S(p,T)$ of subjects of the children of $p$ in $T$ by 
\,$S(p,T) = \{\Subj(c) : c$ is a child of $p$ in $T\}$. 

Take any integer $n$ such that \,$n \geq 1$. 
Suppose that $Y(n)$ is true. 
We shall prove $Y(n\!+\!1)$. 
Suppose the antecedent of $Y(n\!+\!1)$ holds and that \,$|T| = n\!+\!1$. 
Let $p_0$ be the root of $T$. 
If \,$\alg(p_0) \nte \{\alpha,\alpha'\}$\, then $T(\alpha\!:\!\alpha')$ is $T$ 
and so $T(\alpha\!:\!\alpha')$ is an evaluation tree of $\calP$. 
So suppose \,$\alg(p_0) \e \{\alpha,\alpha'\}$. 
Let $q_0$ be the root of $T(\alpha\!:\!\alpha')$. 

If $p_0$ satisfies T\ref{Tset} then let \,$\Subj(p_0) = (\alpha,H,F)$. 
Hence \,$\Subj(q_0) = (\alpha',H(\alpha\!:\!\alpha'),F)$. 
So $q_0$ satisfies T\ref{Tset}. 
Recall that \,$S(p_0,T[\alpha,H,F])$ = $\{(\alpha,H,f) : f \e F\}$. 
So \,$S(q_0,T[\alpha,H,F](\alpha\!:\!\alpha'))$ 
= $\{(\alpha',H(\alpha\!:\!\alpha'),f) : f \e F\}$ 
= $S(q_0,T[\alpha',H(\alpha\!:\!\alpha'),F])$, by T\ref{Tset}. 
But for each $(\alpha,H,f)$ in $S(p_0,T[\alpha,H,F])$, $T[\alpha,H,f]$ is an 
evaluation tree of $\calP$ and \,$|T[\alpha,H,f]| \leq n$. 
So by $Y(n)$, $T[\alpha,H,f](\alpha\!:\!\alpha')$ 
is an evaluation tree of $\calP$. 
Hence \,$T[\alpha,H,f](\alpha\!:\!\alpha')$ 
= $T[\alpha',H(\alpha\!:\!\alpha'),f]$. 
Thus \,$T(\alpha\!:\!\alpha')$ = $T[\alpha,H,F](\alpha\!:\!\alpha')$ 
= $T[\alpha',H(\alpha\!:\!\alpha'),F]$\, 
which is an evaluation tree of $\calP$. 

Since $p_0$ has a child, $p_0$ does not satisfy T\ref{Tfact}. 

If $p_0$ satisfies T\ref{Tfml} then let \,$\Subj(p_0) = (\alpha,H,f)$. 
Hence \,$\Subj(q_0) = (\alpha',H(\alpha\!:\!\alpha'),f)$. 
So $q_0$ satisfies T\ref{Tfml}. 
Recall that \,$S(p_0,T[\alpha,H,f])$ = 
$\{(\alpha,H,f,r) : \alpha r \nte H$\, and \,$r \e R^s_d[f]\}$. 
Since \,$\alpha r \e H$\, iff \,$\alpha' r \e H(\alpha\!:\!\alpha')$\, we have 
\,$\alpha r \nte H$\, iff \,$\alpha' r \nte H(\alpha\!:\!\alpha')$. 
So \,$S(q_0,T[\alpha,H,f](\alpha\!:\!\alpha'))$ 
= $\{(\alpha',H(\alpha\!:\!\alpha'),f,r) : 
	\alpha r \nte H$\, and \,$r \e R^s_d[f]\}$ 
= $\{(\alpha',H(\alpha\!:\!\alpha'),f,r) : 
	\alpha' r \nte H(\alpha\!:\!\alpha')$\, and \,$r \e R^s_d[f]\}$ 
= $S(q_0,T[\alpha',H(\alpha\!:\!\alpha'),f])$, by T\ref{Tfml}. 
But for each $(\alpha,H,f,r)$ in $S(p_0,T[\alpha,H,f])$, $T[\alpha,H,f,r]$ is an 
evaluation tree of $\calP$ and \,$|T[\alpha,H,f,r]| \leq n$. 
So by $Y(n)$, $T[\alpha,H,f,r](\alpha\!:\!\alpha')$ 
is an evaluation tree of $\calP$. 
Hence \,$T[\alpha,H,f,r](\alpha\!:\!\alpha')$ 
= $T[\alpha',H(\alpha\!:\!\alpha'),f,r]$. 
Thus \,$T(\alpha\!:\!\alpha')$ = $T[\alpha,H,f](\alpha\!:\!\alpha')$ 
= $T[\alpha',H(\alpha\!:\!\alpha'),f]$\, 
which is an evaluation tree of $\calP$. 

If $p_0$ satisfies T\ref{TFor} then let \,$\Subj(p_0) = (\alpha,H,f,r)$. 
Hence \,$\Subj(q_0) = (\alpha',H(\alpha\!:\!\alpha'),f,r)$. 
Since \,$\alpha r \e H$\, iff \,$\alpha' r \e H(\alpha\!:\!\alpha')$\, we have 
\,$\alpha r \nte H$\, iff \,$\alpha' r \nte H(\alpha\!:\!\alpha')$. 
So $q_0$ satisfies T\ref{TFor}. 
Recall that \,$S(p_0,T[\alpha,H,f,r])$ = $\{(\alpha,H\!+\!\alpha r,A(r))\}$ 
$\!\cup\!$ $\{(\alpha,H,f,r,s) : s \e \Foe(\alpha,f,r)\}$. 
Since \,$\Foe(\alpha,f,r) = \Foe(\alpha',f,r)$, 
\,$S(q_0,T[\alpha,H,f,r](\alpha\!:\!\alpha'))$ 
= $\{(\alpha',H(\alpha\!:\!\alpha')\!+\!\alpha' r,A(r))\}$ $\!\cup\!$ 
	$\{(\alpha',H(\alpha\!:\!\alpha'),f,r,s) : s \e \Foe(\alpha,f,r)\}$ 
= $S(q_0,T[\alpha',H(\alpha\!:\!\alpha'),f,r])$, by T\ref{TFor}. 
But $T[\alpha,H\!+\!\alpha r,A(r)]$ is an evaluation tree of $\calP$ and 
\,$|T[\alpha,H\!+\!\alpha r,A(r)]| \leq n$. 
So by $Y(n)$, $T[\alpha,H\!+\!\alpha r,A(r)](\alpha\!:\!\alpha')$ 
is an evaluation tree of $\calP$. 
Hence \,$T[\alpha,H\!+\!\alpha r,A(r)](\alpha\!:\!\alpha')$ 
= $T[\alpha',H(\alpha\!:\!\alpha')\!+\!\alpha' r,A(r)]$. 
Also for each $(\alpha,H,f,r,s)$ in $S(p_0,T[\alpha,H,f,r])$, 
$T[\alpha,H,f,r,s]$ is an evaluation tree of $\calP$ and 
\,$|T[\alpha,H,f,r,s]| \leq n$. 
So by $Y(n)$, $T[\alpha,H,f,r,s](\alpha\!:\!\alpha')$ 
is an evaluation tree of $\calP$. 
Hence \,$T[\alpha,H,f,r,s](\alpha\!:\!\alpha')$ 
= $T[\alpha',H(\alpha\!:\!\alpha'),f,r,s]$. 
Thus \,$T(\alpha\!:\!\alpha')$ = $T[\alpha,H,f,r](\alpha\!:\!\alpha')$ 
= $T[\alpha',H(\alpha\!:\!\alpha'),f,r]$\, 
which is an evaluation tree of $\calP$. 

If $p_0$ satisfies T\ref{TDftd} then let \,$\Subj(p_0) = (\alpha,H,f,r,s)$. 
Hence \,$\Subj(q_0) = (\alpha',H(\alpha\!:\!\alpha'),f,r,s)$. 
Since \,$\alpha r \e H$\, iff \,$\alpha' r \e H(\alpha\!:\!\alpha')$\, we have 
\,$\alpha r \nte H$\, iff \,$\alpha' r \nte H(\alpha\!:\!\alpha')$. 
So $q_0$ satisfies T\ref{TDftd}. 
Recall that \,$S(p_0,T[\alpha,H,f,r,s])$ = $\{(\alpha,H\!+\!\alpha t,A(t)) : 
\alpha t \nte H$\, and \,$t \e R^s_d[f;s]\}$ $\!\cup\!$ 
$\{-(\alpha',H\!+\!\alpha's,A(s)) : \alpha's \nte H\}$. 
Also since \,$\alpha' r \e H$\, iff \,$\alpha r \e H(\alpha\!:\!\alpha')$\, 
we have \,$\alpha' r \nte H$\, iff \,$\alpha r \nte H(\alpha\!:\!\alpha')$. 
So \,$S(q_0,T[\alpha,H,f,r,s](\alpha\!:\!\alpha'))$ 
= $\{(\alpha',H(\alpha\!:\!\alpha')\!+\!\alpha' t,A(t)) : 
\alpha t \nte H$\, and \,$t \e R^s_d[f;s]\}$ $\!\cup\!$ 
$\{-(\alpha,H(\alpha\!:\!\alpha')\!+\!\alpha s,A(s)) : 
\alpha' s \nte H\}$ \nl 
= $\{(\alpha',H(\alpha\!:\!\alpha')\!+\!\alpha' t,A(t)) : 
\alpha' t \nte H(\alpha\!:\!\alpha')$\, and \,$t \e R^s_d[f;s]\}$ $\!\cup\!$ 
$\{-(\alpha,H(\alpha\!:\!\alpha')\!+\!\alpha s,A(s)) : 
\alpha s \nte H(\alpha\!:\!\alpha')\}$ \nl 
= $S(q_0,T[\alpha',H(\alpha\!:\!\alpha'),f,r,s])$, by T\ref{TDftd}. 

But for each $(\alpha,H\!+\!\alpha t,A(t))$ in $S(p_0,T[\alpha,H,f,r,s])$, 
$T[\alpha,H\!+\!\alpha t,A(t)]$ is an evaluation tree of $\calP$ and 
\,$|T[\alpha,H\!+\!\alpha t,A(t)]| \leq n$. 
So by $Y(n)$, $T[\alpha,H\!+\!\alpha t,A(t)](\alpha\!:\!\alpha')$ 
is an evaluation tree of $\calP$. 
Hence \,$T[\alpha,H\!+\!\alpha t,A(t)](\alpha\!:\!\alpha')$ 
= $T[\alpha',H(\alpha\!:\!\alpha')\!+\!\alpha t,A(t)]$. 
Also if \,$-(\alpha',H\!+\!\alpha's,A(s)) \in S(p_0,T[\alpha,H,f,r,s])$, 
then $T[-(\alpha',H\!+\!\alpha's,A(s))]$ is an evaluation tree of $\calP$ and 
\,$|T[-(\alpha',H\!+\!\alpha's,A(s))]| \leq n$. 
So by $Y(n)$, $T[-(\alpha',H\!+\!\alpha's,A(s))](\alpha\!:\!\alpha')$ 
is an evaluation tree of $\calP$. 
Hence \,$T[-(\alpha',H\!+\!\alpha's,A(s))](\alpha\!:\!\alpha')$ 
= $T[-(\alpha,H(\alpha\!:\!\alpha')\!+\!\alpha s,A(s))]$. 

Thus \,$T(\alpha\!:\!\alpha')$ = $T[\alpha,H,f,r,s](\alpha\!:\!\alpha')$ 
= $T[\alpha',H(\alpha\!:\!\alpha'),f,r,s]$\, 
which is an evaluation tree of $\calP$. 

If $p_0$ satisfies T\ref{Tminus} then let \,$\Subj(p_0) = -(\alpha',H,F)$. 
Hence \,$\Subj(q_0) = -(\alpha,H(\alpha\!:\!\alpha'),F)$. 
So $q_0$ satisfies T\ref{Tminus}. 
Recall that \,$S(p_0,T[-(\alpha',H,F)])$ = $\{(\alpha',H,F)\}$. 
So \nl
\,$S(q_0,T[-(\alpha',H,F)](\alpha\!:\!\alpha'))$ 
= $\{(\alpha,H(\alpha\!:\!\alpha'),F)\}$ 
= $S(q_0,T[-(\alpha,H(\alpha\!:\!\alpha'),F)])$, by T\ref{Tminus}. 
But $T[\alpha',H,F]$ is an evaluation tree of $\calP$ and 
\,$|T[\alpha',H,F]| \leq n$. 
So by $Y(n)$, $T[\alpha',H,F](\alpha\!:\!\alpha')$ 
is an evaluation tree of $\calP$. 
Hence \,$T[\alpha',H,F](\alpha\!:\!\alpha')$ 
= $T[\alpha,H(\alpha\!:\!\alpha'),F]$. 
Thus \,$T(\alpha\!:\!\alpha')$ = $T[-(\alpha',H,F)](\alpha\!:\!\alpha')$ 
= $T[-(\alpha,H(\alpha\!:\!\alpha'),F)]$\, 
which is an evaluation tree of $\calP$. 

Therefore $Y(n\!+\!1)$, and hence the lemma, is proved by induction. 

\noindent \eop{EndProofLem\ref{Lem:beta is isomorphic to beta'}} 
\end{Lem}  
\begin{Lem} 
\label{Lem:If (psi,H)|- x then (psi',H(psi:psi'))|- x.} 
Suppose \,$\calP = (R,>)$\, is a plausible theory, $H$ is a $\psi$-history, 
and $x$ is either a formula or a finite set of formulas. 
If \,$(\psi,H) \vminus x$\, then \,$(\psi',H(\psi\!:\!\psi')) \vminus x$. \nl
Hence \,$\calP(\psi) \!\subseteq\! \calP(\psi')$. 

\noindent \pf{Proof} 

Suppose $(R,>)$ is a plausible theory. 
Let $Y(n)$ denote the following conditional statement. 
``If $H$ is a $\psi$-history, $F$ is a finite set of formulas, $f$ is a formula, 
\,$x \e \{F, f\}$, \,$T(\psi,H,x) = +1$, and \,$|T[\psi,H,x]| \leq n$\, 
then \,$T(\psi',H(\psi\!:\!\psi'),x) = +1$." 

By Theorem \ref{Thm:P()=+1 iff |- iff T()=+1}(\ref{equiv set},\ref{equiv formula}), and 
Theorem \ref{Thm:Decisiveness}(\ref{T[alpha,H,f] is finite.}), 
it suffices to prove $Y(n)$ by induction on $n$. 

Suppose \,$n = 1$. 
Let the antecedent of $Y(1)$ hold. 
Let the only node of $T[\psi,H,x]$ be $p_0$. 
Let the root of $T[\psi',H(\psi\!:\!\psi'),x]$ be $q_0$. 

If $p_0$ satisfies T\ref{Tset} then \,$x = \{\}$\, and so 
$q_0$ satisfies T\ref{Tset}. 
So by T\ref{Tset}, $T[\psi',H(\psi\!:\!\psi'),\{\}]$ has only one node and 
\,$T(\psi',H(\psi\!:\!\psi'),\{\}) = +1$. 
If $p_0$ satisfies T\ref{Tfact} then \,$x = f$\, and \,$\Ax \models f$. 
So $q_0$ satisfies T\ref{Tfact} and hence \,$T(\psi',H(\psi\!:\!\psi'),f) = +1$. 
Since $p_0$ has no children and the proof value of $p_0$ is $+1$, $p_0$ does not satisfy T\ref{Tfml}. 
Since the subject of $p_0$ is $(\psi,H,x)$, $p_0$ does not satisfy 
T\ref{TFor}, or T\ref{TDftd}, or T\ref{Tminus}. 
Thus the base case holds. 

Take any positive integer $n$. 
Suppose $Y(n)$ is true. 
We shall prove $Y(n\!+\!1)$. 

Suppose the antecedent of $Y(n\!+\!1)$ holds and that 
\,$|T[\psi,H,x]| = n\!+\!1$. 
Then \,$T(\psi,H,x) = +1$. 
Let $p_0$ be the root of $T[\psi,H,x]$ 
and $q_0$ be the root of $T[\psi',H(\psi\!:\!\psi'),x]$. 

If $p_0$ satisfies T\ref{Tset} then \,$x = F$. 
We see that $q_0$ also satisfies T\ref{Tset}. 
So \nl
\,$t(p_0) = ((\psi,H,F), \min, +1)$\, and 
\,$t(q_0) = ((\psi',H(\psi\!:\!\psi'),F), \min, w_0)$, where \nl
\,$w_0 = T(\psi',H(\psi\!:\!\psi'),F) \in \{+1,-1\}$. 
Let $\{p_{\!f} : f \e F\}$ be the set of children of $p_0$ and 
$\{q_{\!f} : f \e F\}$ be the set of children of $q_0$. 
Let $f$ be any formula in $F$. 
Then the subject of $p_{\!f}$ is $(\psi,H,f)$, and the subject of 
$q_{\!f}$ is $(\psi',H(\psi\!:\!\psi'),f)$. 
Also \,$|T[\psi,H,f]| \leq n$\, and the proof value of $p_{\!f}$ is $+1$ because $p_0$ is a min node with proof value $+1$. 
So by $Y(n)$ the proof value of $q_{\!f}$ is $+1$. 
But this is true for each $f$, so \,$w_0 = +1$, as required. 

Since $p_0$ has a child, $p_0$ does not satisfy T\ref{Tfact}. 

If $p_0$ satisfies T\ref{Tfml} then \,$x = f$. 
We see that $q_0$ also satisfies T\ref{Tfml}. 
So \nl
\,$t(p_0) = ((\psi,H,f), \max, +1)$\, and 
\,$t(q_0) = ((\psi',H(\psi\!:\!\psi'),f), \max, w_0)$, where \nl
\,$w_0 = T(\psi',H(\psi\!:\!\psi'),f) \in \{+1, -1\}$. 

We shall adopt the following naming conventions. 
Each non-root node of $T$ is denoted by $p_l(\#,y)$ where 
$l$ is the level of the node, 
$\#$ is the number in $[\ref{Tset}..\ref{Tminus}]$ such that 
the node satisfies T\#, and $y$ is a rule, or a formula, or a set, 
which distinguishes siblings. 
The proof value of $p_l(\#,y)$ will be denoted by $v_l(\#,y)$. 
For non-root nodes in $T[\psi',H(\psi\!:\!\psi'),f]$ we shall use $q_l(\#,y)$, 
and its proof value will be denoted by $w_l(\#,y)$. 

Let the set of children of $p_0$ be 
\,$\{p_1(\ref{TFor},r) : \psi r \nte H$\, and \,$r \e R^s_d[f]\}$, 
where the tags of these children are: \nl 
\,$t(p_1(\ref{TFor},r))$ = $((\psi,H,f,r),\min,v_1(\ref{TFor},r))$. 
So \,$+1 = 
\max\{v_1(\ref{TFor},r) : \psi r \nte H$\, and \,$r \e R^s_d[f]\}$. 

Let the set of children of $q_0$ be \,$\{q_1(\ref{TFor},r) : 
\psi' r \nte H(\psi\!:\!\psi')$\, and \,$r \e R^s_d[f]\}$, 
where the tags of these children are: \,$t(q_1(\ref{TFor},r))$ = 
$((\psi',H(\psi\!:\!\psi'),f,r),\min,w_1(\ref{TFor},r))$. 
So \,$w_0 = \max\{w_1(\ref{TFor},r) : \psi' r \nte H(\psi\!:\!\psi')$\, 
and \,$r \e R^s_d[f]\}$. 

From above there exists $r_0$ in $R^s_d[f]$ such that 
\,$\psi r_0 \nte H$\, and \,$v_1(\ref{TFor},r_0) = +1$. 
By Definition \ref{Defn:(alpha:alpha')}, 
if \,$\psi' r_0 \e H(\psi\!:\!\psi')$\, then \,$\psi r_0 \e H$. 
So if \,$\psi r_0 \nte H$\, then \,$\psi' r_0 \nte H(\psi\!:\!\psi')$. 
Hence $q_1(\ref{TFor},r_0)$ exists. 
We shall show that \,$w_1(\ref{TFor},r_0) = +1$\, 
and hence that \,$w_0 = +1$, as required. 

Let the set of children of $p_1(\ref{TFor},r_0)$ be 
\,$\{p_2(\ref{Tset},r_0)\}$ $\cup$ 
$\{p_2(\ref{TDftd},s) : s \e \Foe(\psi,f,r_0)\}$, 
where the tags of these children are: \,$t(p_2(\ref{Tset},r_0)) 
= ((\psi,H$+$\psi r_0,A(r_0)),\min,v_2(\ref{Tset},r_0))$; and \nl
\,$t(p_2(\ref{TDftd},s)) = ((\psi,H,f,r_0,s),\max,v_2(\ref{TDftd},s))$. \nl 
So \,$+1 = v_1(\ref{TFor},r_0)$ = $\min[\{v_2(\ref{Tset},r_0)\}$ 
$\cup$ $\{v_2(\ref{TDftd},s) : s \e \Foe(\psi,f,r_0)\}]$. \nl 
Hence \,$v_2(\ref{Tset},r_0) = +1$; and for each $s$ in $\Foe(\psi,f,r_0)$, 
\,$v_2(\ref{TDftd},s) = +1$. 

Let the set of children of $q_1(\ref{TFor},r_0)$ be \,$\{q_2(\ref{Tset},r_0)\}$ 
$\cup$ $\{q_2(\ref{TDftd},s) : s \e R[\neg f;r_0]\}$, 
where the tags of these children are: 
\,$t(q_2(\ref{Tset},r_0)) = 
((\psi',H(\psi\!:\!\psi')$+$\psi' r_0,A(r_0)),\min,w_2(\ref{Tset},r_0))$; and  
\,$t(q_2(\ref{TDftd},s)) = 
((\psi',H(\psi\!:\!\psi'),f,r_0,s),\max,w_2(\ref{TDftd},s))$. \nl 
So \,$w_1(\ref{TFor},r_0) = \min[\{w_2(\ref{Tset},r_0)\}$ $\cup$ 
$\{w_2(\ref{TDftd},s) : s \e R[\neg f;r_0]\}]$. 

Since \,$v_2(\ref{Tset},r_0) = +1$, 
\,$T(\psi,H$+$\psi r_0,A(r_0)) = +1$\, and 
\,$|T[\psi,H$+$\psi r_0,A(r_0)]| \leq n$. 
So by $Y(n)$, \,$w_2(\ref{Tset},r_0) = 
T(\psi',H(\psi\!:\!\psi')$+$\psi' r_0,A(r_0)) = +1$. 
Therefore \,$w_1(\ref{TFor},r_0) = 
\min\{w_2(\ref{TDftd},s) : s \e R[\neg f;r_0]\}$. 
If \,$R[\neg f;r_0] = \{\}$\, then 
\,$w_1(\ref{TFor},r_0) = \min\{\} = +1$\, as desired. 
So suppose \,$R[\neg f;r_0] \neq \{\}$. 

Since \,$R[\neg f;r_0] \!\subseteq\! \Foe(\psi,f,r_0)$, 
if $q_2(\ref{TDftd},s)$ exists then $p_2(\ref{TDftd},s)$ exists. 

For each node, $p_2(\ref{TDftd},s)$, 
let the set of children of $p_2(\ref{TDftd},s)$ be 
$\{p_3(\ref{Tset},t) : \psi t \nte H$ and $t \e R^s_d[f;s]\}$ $\cup$ 
$\{p_3(\ref{Tminus},s) : \psi' s \nte H\}$, 
where the tags of these children are: 
\,$t(p_3(\ref{Tset},t)) = 
((\psi,H$+$\psi t,A(t)),\min,v_3(\ref{Tset},t))$; and 
\,$t(p_3(\ref{Tminus},s)) = 
(-(\psi',H$+$\psi' s,A(s)), - ,v_3(\ref{Tminus},s))$. \nl
So \,$+1 = v_2(\ref{TDftd},s)$ = 
$\max[\{v_3(\ref{Tset},t) : \psi t \nte H$ and $t \e R^s_d[f;s]\}$ 
$\cup$ $\{v_3(\ref{Tminus},s) : \psi' s \nte H\}]$. 

For each node, $q_2(\ref{TDftd},s)$, 
let the set of children of $q_2(\ref{TDftd},s)$ be 
$\{q_3(\ref{Tset},t) : \psi' t \nte H(\psi\!:\!\psi')$ and 
$t \e R^s_d[f;s]\}$ $\cup$ 
$\{q_3(\ref{Tminus},s) : \psi s \nte H(\psi\!:\!\psi')\}$, 
where the tags of these children are: 
\,$t(q_3(\ref{Tset},t)) = 
((\psi',H(\psi\!:\!\psi')$+$\psi' t,A(t)),\min,w_3(\ref{Tset},t))$; and 
\,$t(q_3(\ref{Tminus},s)) = 
(-(\psi,H(\psi\!:\!\psi')$+$\psi s,A(s)), - ,w_3(\ref{Tminus},s))$. 
So \,$w_2(\ref{TDftd},s)$ = 
$\max[\{w_3(\ref{Tset},t) : \psi' t \nte H(\psi\!:\!\psi')$ and 
$t \e R^s_d[f;s]\}$ $\cup$ 
$\{w_3(\ref{Tminus},s) : \psi s \nte H(\psi\!:\!\psi')\}]$. 

By Definition \ref{Defn:(alpha:alpha')}, 
\,$\psi t \e H$\, iff \,$\psi' t \e H(\psi\!:\!\psi')$. 
So \,$\psi t \nte H$\, iff \,$\psi' t \nte H(\psi\!:\!\psi')$. 
Therefore $p_3(\ref{Tset},t)$ exists iff $q_3(\ref{Tset},t)$ exists. 
If there exists \,$t_0 \e R^s_d[f;s]$\, such that 
\,$v_3(\ref{Tset},t_0) = +1$\, then 
\,$T(\psi,H$+$\psi t_0,A(t_0)) = +1$\, and 
\,$|T[\psi,H$+$\psi t_0,A(t_0)]| \leq n$\, so by $Y(n)$, 
\,$w_3(\ref{Tset},t_0) = 
T(\psi',H(\psi\!:\!\psi')$+$\psi' t_0,A(t_0)) = +1$. 
Hence \,$w_2(\ref{TDftd},s) = +1$. 

So suppose no such \,$t_0 \e R^s_d[f;s]$\, exists. 
Then $p_3(\ref{Tminus},s)$ exists such that \,$\psi' s \nte H$\, and 
\,$v_3(\ref{Tminus},s) = +1$. 
By Definition \ref{Defn:(alpha:alpha')}, 
\,$\psi' s \e H$\, iff \,$\psi s \e H(\psi\!:\!\psi')$. 
So \,$\psi' s \nte H$\, iff \,$\psi s \nte H(\psi\!:\!\psi')$. 
Hence $q_3(\ref{Tminus},s)$ exists. 

Let the child of $p_3(\ref{Tminus},s)$ be $p_4(\ref{Tset},s)$ where 
\,$t(p_4(\ref{Tset},s)) = 
((\psi',H$+$\psi' s,A(s)),\min,v_4(\ref{Tset},s))$\, and 
\,$v_3(\ref{Tminus},s) = -v_4(\ref{Tset},s)$. 
So \,$v_4(\ref{Tset},s) = -1$. 

Let the child of $q_3(\ref{Tminus},s)$ be $q_4(\ref{Tset},s)$ where \nl
\,$t(q_4(\ref{Tset},s)) = 
((\psi,H(\psi\!:\!\psi')$+$\psi s,A(s)),\min,w_4(\ref{Tset},s))$\, 
and \,$w_3(\ref{Tminus},s) = -w_4(\ref{Tset},s)$. 

Assume \,$w_4(\ref{Tset},s) = +1$. 
Then \,$T(\psi,H(\psi\!:\!\psi')$+$\psi s,A(s)) = +1$\, and \nl
\,$|T[\psi,H(\psi\!:\!\psi')$+$\psi s,A(s)]| \leq n$\, so by $Y(n)$, 
\,$T(\psi',H(\psi\!:\!\psi')(\psi\!:\!\psi')$+$\psi's,A(s)) = +1$. 
But \,$H(\psi\!:\!\psi')(\psi\!:\!\psi') = H$. 
So \,$T(\psi',H$+$\psi' s,A(s)) = +1$. 
From above \,$-1 = v_4(\ref{Tset},s) = T(\psi',H$+$\psi' s,A(s)) = +1$. 
This contradiction shows that \,$w_4(\ref{Tset},s) = -1$. 
Hence \,$w_3(\ref{Tminus},s) = +1$\, and so \,$w_2(\ref{TDftd},s) = +1$. 

\smallskip 

So in both cases for all $s$ in $R[\neg f;r_0]$, 
\,$w_2(\ref{TDftd},s) = +1$\, and so \,$w_1(\ref{TFor},r_0) = +1$. 
Hence \,$w_0 = +1$, as required. 

Since the subject of $p_0$ is $(\psi,H,x)$, $p_0$ does not satisfy 
T\ref{TFor}, or T\ref{TDftd}, or T\ref{Tminus}. 

Thus $Y(n)$, and hence the lemma, is proved by induction. 

\noindent \eop{EndProofLem\ref{Lem:If (psi,H)|- x then (psi',H(psi:psi'))|- x.}} 
\end{Lem} 
\begin{Lem} 
\label{Lem:If f is satisfiable and Ax notmodels f then R^s_d[f] is finite.} 
Suppose $(R,>)$ is a plausible theory, \,$\Ax = \Ax(R)$, and $f$ is a formula. 
If $f$ is satisfiable and \,$\Ax \not\models f$\, then $R^s_d[f]$ is finite. 

\noindent \pf{Proof} 

Suppose $(R,>)$ is a plausible theory, \,$\Ax = \Ax(R)$, and $f$ is a formula. 
Also suppose $f$ is satisfiable and \,$\Ax \not\models f$. 
Take any $\alpha$ in $\Alg\!-\!\{\varphi\}$. 
By Definition \ref{Defn:PlausibleTheory, PlausibleLogic}, 
$T[\alpha,(),f]$ is finite, and so its root has only finitely many children. 
The root of $T[\alpha,(),f]$ satisfies 
T\ref{Tfml} of Definition \ref{Defn:EvaluationTree}. 
Therefore $R^s_d[f]$ is finite. 

\noindent \eop{EndProofLem\ref{Lem:If f is satisfiable and Ax notmodels f then R^s_d[f] is finite.}} 
\end{Lem} 
\begin{Thm} [Theorem \ref{Thm:Consistency} Consistency] 
\label{AppThm:Consistency} 
\raggedright \parindent = 1.5em

Suppose $(R,>)$ is a plausible theory, $\Ax = \Ax(R)$, 
$\alpha \e \{\varphi,\pi,\psi,\beta,\beta'\}$, 
and both $f$ and $g$ are any formulas. 
\begin{compactenum}[1)]
\item \label{App:phi,pi,psi,beta,beta'} 
	If \,$\alpha \vminus f$\, and \,$\alpha \vminus g$\, 
	then \,$\Ax \!\cup\! \{f,g\}$\, is satisfiable. 
\item \label{App:psi-psi'} 
	If \,$(\psi,H) \vminus f$\, then \,$(\psi',H) \not\vminus \neg f$. 
\item \label{App:pi-pi'} 
	Suppose that whenever \,$s \e R^s_d[\neg f]$\, and 
	\,$(\pi',H$+$\pi' s) \vminus A(s)$\, then \,$R^s_d[f;s] = \{\}$. \nl 
	If \,$(\pi,H) \vminus f$\, then \,$(\pi',H) \not\vminus \neg f$. 
\end{compactenum}

\noindent \pf{Proof}

Suppose $(R,>)$ is a plausible theory, $\Ax = \Ax(R)$, 
$\alpha \e \{\varphi,\pi,\psi,\beta,\beta'\}$, 
and both $f$ and $g$ are any formulas. 

(\ref{App:phi,pi,psi,beta,beta'}) 
Suppose \,$\alpha \vminus f$\, and \,$\alpha \vminus g$. 
So by Theorem \ref{Thm:P()=+1 iff |- iff T()=+1}(\ref{equiv formula}), 
\,$T(\alpha,(),f) = +1$\, and \,$T(\alpha,(),g) = +1$. 

Let $p_0$ be the root of $T[\alpha,(),f]$ and 
$q_0$ be the root of $T[\alpha,(),g]$. 
Since the subject of $p_0$ is $(\alpha,(),f)$, 
$p_0$ does not satisfy T\ref{Tset}, or T\ref{TFor}, or T\ref{TDftd}, 
or T\ref{Tminus}. 
Since the subject of $q_0$ is $(\alpha,(),g)$, 
$q_0$ does not satisfy T\ref{Tset}, or T\ref{TFor}, or T\ref{TDftd}, 
or T\ref{Tminus}. 
Therefore $p_0$ satisfies T\ref{Tfact} or T\ref{Tfml}, 
and $q_0$ satisfies T\ref{Tfact} or T\ref{Tfml}. 
So there are four cases to consider. 

Case 1: $p_0$ satisfies T\ref{Tfact} and $q_0$ satisfies T\ref{Tfact}. \nl 
Then \,$\Ax \models f$\, and \,$\Ax \models g$. 
By Lemma \ref{Lem:Ax}(\ref{Ax is satisfiable.}), 
$\Ax$ is satisfiable. 
Therefore \,$\Ax \!\cup\! \{f,g\}$\, is satisfiable. 

Case 2: $p_0$ satisfies T\ref{Tfact} and $q_0$ satisfies T\ref{Tfml}. \nl 
Then \,$\Ax \models f$, \,$\rse \e R[f]$, \,$\Ax \not\models g$, and 
\,$\alpha \neq \varphi$. 
So \,$\alpha \e \{\pi,\psi,\beta,\beta'\}$. 
Assume \,$\Ax \!\cup\! \{f,g\}$\, is unsatisfiable. 
By Lemma \ref{Lem:R[.]}(\ref{If Ax+f+g is unsat then R[f] subseteq R[comp g].}), 
\,$R[f] \subseteq R[\neg g]$. 
So \,$\rse \e R[\neg g]$. 
By T\ref{Tfml}, $q_0$ has a child, $q_1$, in $T[\alpha,(),g]$ such that 
\,$t(q_1) = ((\alpha,(),g,r_{\!g}),\min,+1)$\, and \,$r_{\!g} \e R^s_d[g]$. 
By T\ref{TFor}, $q_1$ has a child, $q_2$, in $T[\alpha,(),g]$ such that 
\,$t(q_2) = ((\alpha,(),g,r_{\!g},\rse),\max,+1)$. 
By T\ref{TDftd}, $q_2$ has a child, $q_3$, in $T[\alpha,(),g]$ such that 
\,$\pv(q_3) = +1$. 
By T\ref{TDftd}, \,$\Subj(q_3) \in S(q_2)$. 
By Definition \ref{Defn:subsets of R}(\ref{R'[f;s]}), 
$R^s_d[g;\rse] = \{\}$\, and so \,$S(q_2) = S(q_2,\alpha)$. 
However, \,$A(\rse) = \{\}$.
So \,$t(q_3) = (-(\alpha',(\alpha'\rse),\{\}),-,+1)$. 
By T\ref{Tminus}, $q_3$ has a child, $q_4$ in $T[\alpha,(),g]$\, such that 
\,$\Subj(q_4) = (\alpha',(\alpha'\rse),\{\})$. 
So by T\ref{Tset}, \,$\pv(q_4) = +1$. 
But by T\ref{Tvalue}, \,$+1 = \pv(q_3) = -\pv(q_4) = -1$. 
This contradiction shows that \,$\Ax\!\cup\!\{f,g\}$\, is satisfiable. 

Case 3: $p_0$ satisfies T\ref{Tfml} and $q_0$ satisfies T\ref{Tfact}. \nl 
This case is the same as Case 2 but with $p$ and $q$ interchanged and 
with $f$ and $g$ interchanged. 
So by doing the indicated interchanges 
the proof for Case 2 becomes a proof for Case 3. 

Case 4: $p_0$ satisfies T\ref{Tfml} and $q_0$ satisfies T\ref{Tfml}. \nl 
Then \,$\Ax \not\models f$, \,$\Ax \not\models g$, and \,$\alpha \neq \varphi$. 
So \,$\alpha \e \{\pi,\psi,\beta,\beta'\}$. 
By T\ref{Tfml}, $p_0$ has a child, $p_1$, in $T[\alpha,(),f]$ such that 
\,$t(p_1) = ((\alpha,(),f,r_{\!f}),\min,+1)$\, and \,$r_{\!f} \e R^s_d[f]$. 
By T\ref{Tfml}, $q_0$ has a child, $q_1$, in $T[\alpha,(),g]$ such that 
\,$t(q_1) = ((\alpha,(),g,r_{\!g}),\min,+1)$\, and \,$r_{\!g} \e R^s_d[g]$. 

Assume \,$\Ax \!\cup\! \{f,g\}$\, is unsatisfiable. 
By Lemma \ref{Lem:R[.]}(\ref{If Ax+f+g is unsat then R[f] subseteq R[comp g].}), 
\,$R^s_d[f] \subseteq R[f] \subseteq R[\neg g]$\, and 
\,$R^s_d[g] \subseteq R[g] \subseteq R[\neg f]$. 
So \,$r_{\!f} \e R[\neg g]$\, and \,$r_{\!g} \e R[\neg f]$. 

By T\ref{TFor}, either \,$r_{\!f} \!>\! r_{\!g}$; or 
$p_1$ has a child, $p_2(r_{\!g})$, in $T[\alpha,(),f]$ 
such that \,$\Subj(p_2(r_{\!g})) = (\alpha,(),f,r_{\!f},r_{\!g})$\, and 
\,$\pv(p_2(r_{\!g})) = +1$. 
By T\ref{TDftd}, $p_2(r_{\!g})$ has a child, 
$p_3(r_{\!g})$ in $T[\alpha,(),f]$ such that 
\,$\pv(p_3(r_{\!g})) = +1$\, and 
\,$\Subj(p_3(r_{\!g})) \in S(p_2(r_{\!g}))$. 

Similarly by T\ref{TFor}, either \,$r_{\!g} \!>\! r_{\!f}$; or 
$q_1$ has a child, $q_2(r_{\!f})$, 
in $T[\alpha,(),g]$ such that 
\,$\Subj(q_2(r_{\!f})) = (\alpha,(),g,r_{\!g},r_{\!f})$\, and 
\,$\pv(q_2(r_{\!f})) = +1$. 
By T\ref{TDftd}, $q_2(r_{\!f})$ has a child, 
$q_3(r_{\!f})$ in $T[\alpha,(),g]$ such that 
\,$\pv(q_3(r_{\!f})) = +1$\, and 
\,$\Subj(q_3(r_{\!f})) \in S(q_2(r_{\!f}))$. 

Case 4.1: $\Subj(p_3(r_{\!g})) = -(\alpha',(\alpha' r_{\!g}),A(r_{\!g}))$. \nl 
Since \,$\pv(p_3(r_{\!g})) = +1$, 
$T(\alpha',(\alpha' r_{\!g}),A(r_{\!g})) = -1$. 
So by Theorem \ref{Thm:P()=+1 iff |- iff T()=+1}(\ref{equiv set}), 
\,$(\alpha',(\alpha' r_{\!g})) \not\vminus A(r_{\!g})$. 
But \,$\Subj(q_2(r_{\!g})) = (\alpha,(\alpha r_{\!g}),A(r_{\!g}))$\, 
and \,$\pv(q_2(r_{\!g})) = +1$. 
So \,$T(\alpha,(\alpha r_{\!g}),A(r_{\!g})) = +1$. 
By Theorem \ref{Thm:P()=+1 iff |- iff T()=+1}(\ref{equiv set}), 
\,$(\alpha,(\alpha r_{\!g})) \vminus A(r_{\!g})$. 
By Lemmas \ref{Lem:If alpha|-x then pi'|-x.}, 
\ref{Lem:If (psi,H)|- x then (psi',H(psi:psi'))|- x.}, 
and \ref{Lem:beta is isomorphic to beta'}, 
\,$(\alpha',(\alpha' r_{\!g})) \vminus A(r_{\!g})$. 
This contradiction shows that Case 4.1 cannot occur. 
Thus \,$\Subj(p_3(r_{\!g})) = (\alpha,(\alpha t),A(t))$\, 
where \,$t \e R^s_d[f;r_{\!g}]$. 

Case 4.2: $\Subj(q_3(r_{\!f})) = -(\alpha',(\alpha' r_{\!f}),A(r_{\!f}))$. \nl 
This case is the same as Case 4.1 but with $p$ and $q$ interchanged and with $f$ and $g$ interchanged. 
So by doing the indicated interchanges the proof that Case 4.1 cannot occur becomes a proof that Case 4.2 cannot occur. 
Thus \,$\Subj(q_3(r_{\!f})) = (\alpha,(\alpha t),A(t))$\, 
where \,$t \e R^s_d[g;r_{\!f}]$. 

\smallskip 

In summary Cases 4.1 and 4.2 have shown that we have \nl
either \,$r_{\!f} \!>\! r_{\!g}$\, or there is a $t$ in $R^s_d[f;r_{\!g}]$; 
and also \nl 
either \,$r_{\!g} \!>\! r_{\!f}$\, or there is a $t$ in $R^s_d[g;r_{\!f}]$. 

So there exists $t_{\!f}(1)$ in $R^s_d[f;r_{\!g}] \subseteq R^s_d[f] 
\subseteq R[f] \subseteq R[\neg g]$. 
Hence \,$t_{\!f}(1) > r_{\!g}$\, and \,$t_{\!f}(1) \in R[\neg g]$. 
So $q_2(r_{\!f})$ can be replaced by $q_2(t_{\!f}(1))$, and 
$q_3(r_{\!f})$ can be replaced by $q_3(t_{\!f}(1))$. 

Also there exists $t_{\!g}(1)$ in $R^s_d[g;r_{\!f}] \subseteq R^s_d[g] 
\subseteq R[g] \subseteq R[\neg f]$. 
Hence \,$t_{\!g}(1) > r_{\!f}$\, and \,$t_{\!g}(1) \in R[\neg f]$. 
So $p_2(r_{\!g})$ can be replaced by $p_2(t_{\!g}(1))$, and 
$p_3(r_{\!g})$ can be replaced by $p_3(t_{\!g}(1))$. 

Similarly, the arguments in Cases 4.1 and 4.2 for these new nodes yield rules $t_{\!f}(2)$ and $t_{\!g}(2)$ with the following properties: 
$t_{\!f}(2) \in R^s_d[f;t_{\!g}(1)] \subseteq R^s_d[f] 
\subseteq R[f] \subseteq R[\neg g]$; and 
$t_{\!g}(2) \in R^s_d[g;t_{\!f}(1)] \subseteq R^s_d[g] 
\subseteq R[g] \subseteq R[\neg f]$. 
So \,$t_{\!f}(2) > t_{\!g}(1)$\, and \,$t_{\!f}(2) \e R[\neg g]$\, 
and \,$t_{\!g}(2) > t_{\!f}(1)$\, and \,$t_{\!g}(2) \e R[\neg f]$. 
Hence \,$t_{\!f}(2) > t_{\!g}(1) > r_{\!f}$\, and 
\,$t_{\!g}(2) > t_{\!f}(1) > r_{\!g}$. 

This process can be continued indefinitely to yield the following sequences of rules. 
\,$r_{\!f} < t_{\!g}(1) < t_{\!f}(2) < t_{\!g}(3) < t_{\!f}(4) < ...$\, and 
\,$r_{\!g} < t_{\!f}(1) < t_{\!g}(2) < t_{\!f}(3) < t_{\!g}(4) < ...$ 
Now each \,$t_{\!f}(i) \e R^s_d[f]$\, and each \,$t_{\!g}(i) \e R^s_d[g]$. 
Since \,$\alpha \vminus f$\, and \,$\alpha \vminus g$, by 
Lemma \ref{Lem:Ax}(\ref{If (alpha,H) |- f then Ax+f is satisfiable.}), 
both $f$ and $g$ are satisfiable. 
But \,$\Ax \not\models f$\, and \,$\Ax \not\models g$, so by 
Lemma \ref{Lem:If f is satisfiable and Ax notmodels f then R^s_d[f] is finite.}, 
both $R^s_d[f]$ and $R^s_d[g]$ are finite. 
So for some $i$ and some $j\!>\!i$, \,$t_{\!f}(2i) = t_{\!f}(2j)$. 
Hence $>$ is cyclic, which contradicts the definition of $>$ as being acyclic. 
This contradiction shows that \,$\Ax \!\cup\! \{f,g\}$\, is satisfiable. 

(\ref{App:psi-psi'}) 
If $H$ is a $\psi$-history then $H$ is also a $\psi'$-history. 
Suppose \,$(\psi,H) \vminus f$. 
We shall use Definition \ref{Defn:|-}. 
Assume \,$(\psi',H) \vminus \neg f$. 

By I\ref{|-fml}.\ref{|-For} for $\neg f$, 
(*1)\,$\exists s_1 \e R^s_d[\neg f]$\, such that 
\,$(\psi',H$+$\psi' s_1) \vminus A(s_1)$. 
By I\ref{|-fml}.\ref{|-Against} for $f$ either 
\,$(\psi',H$+$\psi' s_1) \not\vminus A(s_1)$, which contradicts (*1), or 
(*2)\,$\exists r_2 \e R^s_d[f;s_1]$\, such that 
\,$(\psi,H$+$\psi r_2) \vminus A(r_2)$. 
By I\ref{|-fml}.\ref{|-Against} for $\neg f$ either 
\,$(\psi,H$+$\psi r_2) \not\vminus A(r_2)$, which contradicts (*2), or 
(*3)\,$\exists s_3 \e R^s_d[\neg f;r_2]$\, such that 
\,$(\psi',H$+$\psi' s_3) \vminus A(s_3)$. 
By I\ref{|-fml}.\ref{|-Against} for $f$ either 
\,$(\psi',H$+$\psi' s_3) \not\vminus A(s_3)$, which contradicts (*3), or 
(*4)\,$\exists r_4 \e R^s_d[f;s_3]$\, such that 
\,$(\psi,H$+$\psi r_4) \vminus A(r_4)$. 
So we have \,$r_4 > s_3 > r_2 > s_1$. 

We can continue the reasoning in the above paragraph to create two arbitrarily 
long sequences \,$s_1, s_3, ..., s_{2i-1}, ...$\, and 
\,$r_2, r_4, ..., r_{2i}, ...$\, such that 
each \,$s_{2i-1} \e R^s_d[\neg f]$\, and each \,$r_{2i} \e R^s_d[f]$. 
Moreover for each odd $i$, \,$s_{i+2} > r_{i+1} > s_i$. 
Since \,$(\psi,H) \vminus f$, by 
Lemma \ref{Lem:Ax}(\ref{If (alpha,H) |- f then Ax+f is satisfiable.}), 
$f$ is satisfiable and \,$\Ax \not\models \neg f$. 
Since \,$(\psi',H) \vminus \neg f$, by 
Lemma \ref{Lem:Ax}(\ref{If (alpha,H) |- f then Ax+f is satisfiable.}), 
$\neg f$ is satisfiable and \,$\Ax \not\models f$. 
So by 
Lemma \ref{Lem:If f is satisfiable and Ax notmodels f then R^s_d[f] is finite.}, 
both $R^s_d[f]$ and $R^s_d[\neg f]$ are finite. 
So there is an even $j$ and an even $k$ such that \,$j < k$\, and \,$r_j = r_k$. 
Hence $>$ is cyclic, contradicting its acyclicity. 
Thus (\ref{psi-psi'}) is proved. 

(\ref{App:pi-pi'}) 
Suppose that (*) whenever \,$s \e R^s_d[\neg f]$\, and 
\,$(\pi',H$+$\pi' s) \vminus A(s)$\, then \,$R^s_d[f;s] = \{\}$. 

If $H$ is a $\pi$-history then $H$ is also a $\pi'$-history. 
Suppose \,$(\pi,H) \vminus f$. 
We shall use Definition \ref{Defn:|-}. 
Assume \,$(\pi',H) \vminus \neg f$. 

By I\ref{|-fml}.\ref{|-For} for $\neg f$, 
(**)\,$\exists s_1 \e R^s_d[\neg f]$\, such that 
\,$(\pi',H$+$\pi' s_1) \vminus A(s_1)$. 
By I\ref{|-fml}.\ref{|-Against} for $f$ either 
\,$(\pi',H$+$\pi' s_1) \not\vminus A(s_1)$, which contradicts (**), or 
\,$\exists r_2 \e R^s_d[f;s_1]$\, such that 
\,$(\pi,H$+$\pi r_2) \vminus A(r_2)$, which contradicts (*). 
Thus (\ref{pi-pi'}) is proved. 

\noindent \eop{EndProofThm\ref{AppThm:Consistency}} 
\end{Thm}  
\begin{Thm} [Theorem \ref{Thm:Truth values} Truth values]
\label{AppThm:Truth values} 
\raggedright \parindent = 1.5em

Suppose $(R,>)$ is a plausible theory, \,$\alpha \e \Alg$, 
$F$ is a finite set of formulas, and $f$ is a formula. 
\begin{compactenum}[ \,1)]
\item \label{App:V(alpha,notnotf) = V(alpha,f)} 
	$V(\alpha,\neg \neg f) = V(\alpha,f)$. 
\item \label{App:V(alpha,f) = tv{t} iff V(alpha,not f) = tv{f}} 
	$V(\alpha,f) = \tv{t}$\, iff \,$V(\alpha,\neg f) = \tv{f}$. 
\item \label{App:V(alpha,f) = tv{f} iff V(alpha,not f) = tv{t}} 
	$V(\alpha,f) = \tv{f}$\, iff \,$V(\alpha,\neg f) = \tv{t}$. 
\item \label{App:V(alpha,f) = tv{a} iff V(alpha,not f) = tv{a}} 
	$V(\alpha,f) = \tv{a}$\, iff \,$V(\alpha,\neg f) = \tv{a}$. 
\item \label{App:V(alpha,f) = tv{u} iff V(alpha,not f) = tv{u}} 
	$V(\alpha,f) = \tv{u}$\, iff \,$V(\alpha,\neg f) = \tv{u}$. 
\item \label{App:If V(alpha,AND F) = tv{t} then V(alpha,f) = tv{t}} 
	If \,$V(\alpha,\AND F) = \tv{t}$\, then 
	for each $f$ in $F$, \,$V(\alpha,f) = \tv{t}$. 
\item \label{App:If V(alpha,f) = tv{t} then V(alpha,OR F) = tv{t}} 
	If \,$f \e F$\, and \,$V(\alpha,f) = \tv{t}$\, 
	then \,$V(\alpha,\OR F) = \tv{t}$. 
\item \label{App:If alpha isin {phi,pi,psi,beta,beta'} then V(alpha,f) isin {t,f,u}.} 
	If \,$\alpha \e \{\varphi,\pi,\psi,\beta,\beta'\}$\, 
	then \,$V(\alpha,f) \e \{\tv{t},\tv{f},\tv{u}\}$. 
\item \label{App:If V(alpha,f) = a then alpha isin {psi',pi'}.} 
	If \,$V(\alpha,f) = \tv{a}$\, then \,$\alpha \e \{\psi',\pi'\}$. 
\end{compactenum}
\begin{compactenum}[1)] \raggedright
\addtocounter{enumi}{9}
\item \label{App:alpha is complete} 
	If \,$V(\alpha,f) = \tv{t}$\, then \,$\alpha \vminus f$. (completeness)
\item \label{App:alpha is sound} 
	If \,$\alpha \e \{\varphi,\pi,\psi,\beta,\beta'\}$\, and 
	\,$\alpha \vminus f$\, then \,$V(\alpha,f) = \tv{t}$. (soundness)
\end{compactenum}
\noindent \pf{Proof} 

Suppose $(R,>)$ is a plausible theory, \,$\alpha \e \Alg$, 
$F$ is a finite set of formulas, and $f$ is a formula. 
By Theorem \ref{Thm:Right Weakening}(\ref{Right Weakening}), 
\,$\alpha \vminus f$\, iff \,$\alpha \vminus \neg \neg f$; and so 
\,$\alpha \not\vminus f$\, iff \,$\alpha \not\vminus \neg \neg f$. 

(\ref{App:V(alpha,notnotf) = V(alpha,f)})
This follows from Definition \ref{Defn:V(.,.)} and the equivalences noted above. 

(\ref{App:V(alpha,f) = tv{t} iff V(alpha,not f) = tv{f}}) 
\,$V(\alpha,f) = \tv{t}$\, iff 
\,$\alpha \vminus f$\, and \,$\alpha \not\vminus \neg f$. 
\,$V(\alpha,\neg f) = \tv{f}$\, iff 
\,$\alpha \not\vminus \neg f$\, and \,$\alpha \vminus \neg \neg f$. 
So (\ref{V(alpha,f) = tv{t} iff V(alpha,not f) = tv{f}}) holds.  

(\ref{App:V(alpha,f) = tv{f} iff V(alpha,not f) = tv{t}}) 
\,$V(\alpha,f) = \tv{f}$\, iff 
\,$\alpha \not\vminus f$\, and \,$\alpha \vminus \neg f$. 
\,$V(\alpha,\neg f) = \tv{t}$\, iff 
\,$\alpha \vminus \neg f$\, and \,$\alpha \not\vminus \neg \neg f$. 
So (\ref{V(alpha,f) = tv{f} iff V(alpha,not f) = tv{t}}) holds.  

(\ref{App:V(alpha,f) = tv{a} iff V(alpha,not f) = tv{a}}) 
\,$V(\alpha,f) = \tv{a}$\, iff 
\,$\alpha \vminus f$\, and \,$\alpha \vminus \neg f$. 
\,$V(\alpha,\neg f) = \tv{a}$\, iff 
\,$\alpha \vminus \neg f$\, and \,$\alpha \vminus \neg \neg f$. 
So (\ref{V(alpha,f) = tv{a} iff V(alpha,not f) = tv{a}}) holds.  

(\ref{App:V(alpha,f) = tv{u} iff V(alpha,not f) = tv{u}}) 
\,$V(\alpha,f) = \tv{u}$\, iff 
\,$\alpha \not\vminus f$\, and \,$\alpha \not\vminus \neg f$. 
\,$V(\alpha,\neg f) = \tv{u}$\, iff 
\,$\alpha \not\vminus \neg f$\, and \,$\alpha \not\vminus \neg \neg f$. 
So (\ref{V(alpha,f) = tv{u} iff V(alpha,not f) = tv{u}}) holds.  

(\ref{App:If V(alpha,AND F) = tv{t} then V(alpha,f) = tv{t}}) 
Suppose \,$V(\alpha,\AND F) = \tv{t}$. 
Then \,$\alpha \vminus \AND F$\, and \,$\alpha \not\vminus \neg \AND F$. 
By Theorem \ref{AppThm:Right Weakening}(\ref{App:Right Weakening}), 
for each $f$ in $F$, \,$\alpha \vminus f$. 
Take any $f$ in $F$ and assume \,$\alpha \vminus \neg f$. 
By Theorem \ref{AppThm:Right Weakening}(\ref{App:Right Weakening}), 
\,$\alpha \vminus \OR \neg F$, where \,$\neg F = \{\neg f : f \e F\}$. 
So by Theorem \ref{AppThm:Right Weakening}(\ref{App:Right Weakening}), 
\,$\alpha \vminus \neg \AND F$. 
This contradiction shows that for each $f$ in $F$, 
\,$\alpha \not\vminus \neg f$. 
Thus for each $f$ in $F$, \,$V(\alpha,f) = \tv{t}$. 

(\ref{App:If V(alpha,f) = tv{t} then V(alpha,OR F) = tv{t}}) 
Suppose \,$f \e F$\, and \,$V(\alpha,f) = \tv{t}$. 
Then \,$\alpha \vminus f$\, and \,$\alpha \not\vminus \neg f$. 
By Theorem \ref{AppThm:Right Weakening}(\ref{App:Right Weakening}), 
\,$\alpha \vminus \OR F$. 
Assume \,$\alpha \vminus \neg \OR F$. 
By Theorem \ref{AppThm:Right Weakening}(\ref{App:Right Weakening}), 
\,$\alpha \vminus \AND \neg F$\, and so \,$\alpha \vminus \neg f$. 
This contradiction shows that \,$\alpha \not\vminus \neg \OR F$. 
Thus \,$V(\alpha,\OR F) = \tv{t}$. 

(\ref{App:If alpha isin {phi,pi,psi,beta,beta'} then V(alpha,f) isin {t,f,u}.}) 
Suppose \,$\alpha \e \{\varphi,\pi,\psi,\beta,\beta'\}$. 
Recall \,$V(\alpha,f) = \tv{a}$\, iff 
\,$\alpha \vminus f$\, and \,$\alpha \vminus \neg f$. 
So by Theorem \ref{Thm:Consistency}(\ref{phi,pi,psi,beta,beta'}), 
\,$V(\alpha,f) \neq \tv{a}$. 

(\ref{App:If V(alpha,f) = a then alpha isin {psi',pi'}.}) 
This is just the contrapositive of part 
(\ref{App:If alpha isin {phi,pi,psi,beta,beta'} then V(alpha,f) isin {t,f,u}.}). 

(\ref{App:alpha is complete})
Recall \,$V(\alpha,f) = \tv{t}$\, iff 
\,$\alpha \vminus f$\, and \,$\alpha \not\vminus \neg f$. 

(\ref{App:alpha is sound})
Suppose \,$\alpha \e \{\varphi,\pi,\psi,\beta,\beta'\}$\, and 
\,$\alpha \vminus f$. 
By Definition \ref{Defn:V(.,.)} and \,$\alpha \vminus f$\, we have 
\,$V(\alpha,f) \e \{\tv{a}, \tv{t}\}$. 
So by part (\ref{App:If alpha isin {phi,pi,psi,beta,beta'} then V(alpha,f) isin {t,f,u}.}), 
\,$V(\alpha,f) = \tv{t}$. 

\noindent 
\eop{EndProofThm\ref{AppThm:Truth values}} 
\end{Thm} 

\section{Proof of Theorem \ref{Thm:Hierarchy}}
\begin{Lem} 
\label{Lem:If (phi,I)|- x then (alpha,H)|- x.} 
\raggedright \parindent = 1.5em

Suppose \,$\calP = (R,>)$\, is a plausible theory, $\alpha \e \Alg$, 
$I$ is a $\varphi$-history, $H$ is an $\alpha$-history, and 
$x$ is either a formula or a finite set of formulas. \nl
If \,$(\varphi,I) \vminus x$\, then \,$(\alpha,H) \vminus x$. 
Hence \,$\calP(\varphi) \!\subseteq\! \calP(\alpha)$. 

\noindent \pf{Proof} 

Suppose $(R,>)$ is a plausible theory, \,$\Ax = \Ax(R)$, 
$\alpha \e \Alg$, $H$ is an $\alpha$-history, and 
$x$ is either a formula or a finite set of formulas. 
Let \,$(\varphi,I) \vminus x$. 
We shall use Definition \ref{Defn:|-}. 

Case 1: $x$ is a formula. \nl 
Let \,$x = f$. 
Then \,$(\varphi,I) \vminus f$. 
By Definition \ref{Defn:|-}(I\ref{|-fact}), \,$\Ax \models f$\, and 
\,$(\alpha,H) \vminus f$. 

Case 2: $x$ is a finite set of formulas. \nl 
Let \,$x = F$. 
Then \,$(\varphi,I) \vminus F$. 
By I\ref{|-set}, for all $f$ in $F$, \,$(\varphi,I) \vminus f$. 
By Case 1, \,$(\alpha,H) \vminus f$. \nl
So by I\ref{|-set}, \,$(\alpha,H) \vminus F$. 

\noindent 
\eop{EndProofLem\ref{Lem:If (phi,I)|- x then (alpha,H)|- x.}} 
\end{Lem} 
\begin{Defn} \label{Defn:H(pi := psi)} 
If $H$ is a $\pi$-history then define 
\,$H(\pi\!:=\!\psi)$\, to be the sequence formed from $H$ by just 
replacing each $\pi$ by $\psi$, and each $\pi'$ by $\psi'$. 
\end{Defn} 
\begin{Lem} 
\label{Lem:If (pi,H) |- x then (psi,H(pi:=psi)) |- x.} 
\raggedright \parindent = 1.5em

Suppose \,$\calP = (R,>)$\, is a plausible theory, $H$ is a $\pi$-history, and 
$x$ is either a formula or a finite set of formulas. 
If \,$(\pi,H) \vminus x$\, then \,$(\psi,H(\pi\!:=\!\psi)) \vminus x$. \nl
Hence \,$\calP(\pi) \!\subseteq\! \calP(\psi)$. 

\noindent \pf{Proof} 

Suppose $(R,>)$ is a plausible theory and $\Ax$ is its set of axioms. 
Let $Y(n)$ denote the following conditional statement. \nl 
``If $H$ is a $\pi$-history, 
$x$ is either a formula or a finite set of formulas, 
\,$T(\pi,H,x) = +1$, and \,$|T[\pi,H,x]| \leq n$\, 
then \,$T(\psi,H(\pi\!:=\!\psi),x) = +1$." 

By Theorem \ref{Thm:P()=+1 iff |- iff T()=+1}(\ref{equiv set},\ref{equiv formula}), and 
Theorem \ref{AppThm:Decisiveness}(\ref{T[alpha,H,f] is finite.}), 
it suffices to prove $Y(n)$ by induction on $n$. 

Suppose \,$n = 1$. 
Let the antecedent of $Y(1)$ hold. 
Let $p_0$ be the root of $T[\pi,H,x]$ and 
$q_0$ be the root of $T[\psi,H(\pi\!:=\!\psi),x]$. 
Then $p_0$ has no children. 
If $p_0$ satisfies T\ref{Tset} then \,$x = \{\}$\, and 
so $q_0$ satisfies T\ref{Tset}. 
So by T\ref{Tset}, $T[\psi,H(\pi\!:=\!\psi),\{\}]$ has only one node and 
\,$T(\psi,H(\pi\!:=\!\psi),\{\}) = +1$. 
If $p_0$ satisfies T\ref{Tfact} then \,$x = f$\, and \,$\Ax \models f$. 
So $q_0$ satisfies T\ref{Tfact} and hence \,$T(\psi,H(\pi\!:=\!\psi),f) = +1$. 
Since $p_0$ has no children and the proof value of $p_0$ is $+1$, 
$p_0$ does not satisfy T\ref{Tfml}. 
Since the subject of $p_0$ is $(\pi,H,x)$, $p_0$ does not satisfy 
T\ref{TFor}, or T\ref{TDftd}, or T\ref{Tminus}. 
Thus the base case holds. 

Take any positive integer $n$. 
Suppose that $Y(n)$ is true. 
We shall prove $Y(n\!+\!1)$. 

Suppose the antecedent of $Y(n\!+\!1)$ holds and that 
\,$|T[\pi,H,x]| = n\!+\!1$. 
Let $p_0$ be the root of $T[\pi,H,x]$ 
and $q_0$ be the root of $T[\psi,H(\pi\!:=\!\psi),x]$. 

If $p_0$ satisfies T\ref{Tset} then let $x$ be $F$. 
We see that $q_0$ also satisfies T\ref{Tset}. 
So \,$t(p_0) = ((\pi,H,F), \min, +1)$\, and 
\,$t(q_0) = ((\psi,H(\pi\!:=\!\psi),F), \min, w_0)$, where 
\,$w_0 = T(\psi,H(\pi\!:=\!\psi),F) \in \{+1,-1\}$. 
Let $\{p_{\!f} : f \e F\}$ be the set of children of $p_0$ 
and $\{q_{\!f} : f \e F\}$ be the set of children of $q_0$. 
Let $f$ be any formula in $F$. 
Then the subject of $p_{\!f}$ is $(\pi,H,f)$, 
and the subject of $q_{\!f}$ is $(\psi,H(\pi\!:=\!\psi),f)$. 
Also \,$|T[\pi,H,f]| \leq n$\, and the proof value of $p_{\!f}$ is $+1$ 
because $p_0$ is a min node with proof value $+1$. 
So by $Y(n)$ the proof value of $q_{\!f}$ is $+1$. 
But this is true for each $f$, so \,$w_0 = +1$, as required. 

Since $p_0$ has a child, $p_0$ does not satisfy T\ref{Tfact}. 

If $p_0$ satisfies T\ref{Tfml} then let $x$ be $f$. 
We see that $q_0$ also satisfies T\ref{Tfml}. 
So \,$t(p_0) = ((\pi,H,f), \max, +1)$\, and 
\,$t(q_0) = ((\psi,H(\pi\!:=\!\psi),f), \max, w_0)$, where 
\,$w_0 = T(\psi,H(\pi\!:=\!\psi),f) \in \{+1, -1\}$. 

We shall adopt the following naming conventions. 
Each non-root node of $T[\pi,H,f]$ is denoted by $p_l(\#,y)$ where $l$ is the level of the node, $\#$ is the number in $[\ref{Tset}..\ref{Tminus}]$ such that the node satisfies T\#, and $y$ is a rule, or a formula, or a set, which distinguishes siblings. 
The proof value of $p_l(\#,y)$ will be denoted by $v_l(\#,y)$. 
For non-root nodes in $T[\psi,H(\pi\!:=\!\psi),f]$ we shall use $q_l(\#,y)$, and its proof value will be denoted by $w_l(\#,y)$. 

Let the set of children of $p_0$ be 
\,$\{p_1(\ref{TFor},r) : \pi r \nte H$\, and \,$r \e R^s_d[f]\}$, 
where the tags of these children are: \nl 
\,$t(p_1(\ref{TFor},r))$ = $((\pi,H,f,r),\min,v_1(\ref{TFor},r))$. 
So \,$+1 = \max\{v_1(\ref{TFor},r) : 
\pi r \nte H$\, and \,$r \e R^s_d[f]\}$. 

Let the set of children of $q_0$ be 
\,$\{q_1(\ref{TFor},r) : \psi r \nte H(\pi\!:=\!\psi)$\, and 
\,$r \e R^s_d[f]\}$, 
where the tags of these children are: $t(q_1(\ref{TFor},r))$ = 
$((\psi,H(\pi\!:=\!\psi),f,r),\min,w_1(\ref{TFor},r))$. 
So \,$w_0 = \max\{w_1(\ref{TFor},r) : \psi r \nte H(\pi\!:=\!\psi)$\, and 
\,$r \e R^s_d[f]\}$. 

From above there exists $r_0$ in $R^s_d[f]$ such that 
\,$\pi r_0 \nte H$\, and \,$v_1(\ref{TFor},r_0) = +1$. 
So \,$t(p_1(\ref{TFor},r_0)) = ((\pi,H,f,r_0),\min,+1)$. 

Let the set of children of $p_1(\ref{TFor},r_0)$ be 
\,$\{p_2(\ref{Tset},r_0)\}$ $\cup$ 
$\{p_2(\ref{TDftd},s) : s \e \Foe(\pi,f,r_0)\}$, 
where the tags of these children are: 
\,$t(p_2(\ref{Tset},r_0)) = ((\pi,H$+$\pi r_0,A(r_0)),\min,v_2(\ref{Tset},r_0))$; and  
\,$t(p_2(\ref{TDftd},s)) = ((\pi,H,f,r_0,s),\max,v_2(\ref{TDftd},s))$. 
So \,$+1 = v_1(\ref{TFor},r_0)$ = $\min[\{v_2(\ref{Tset},r_0)\}$ $\cup$ 
$\{v_2(\ref{TDftd},s) : s \e \Foe(\pi,f,r_0)\}]$. 
Hence \,$v_2(\ref{Tset},r_0) = +1$; and for each $s$ in $\Foe(\pi,f,r_0)$, 
\,$v_2(\ref{TDftd},s) = +1$. 

Let the set of children of $q_1(\ref{TFor},r_0)$ be 
\,$\{q_2(\ref{Tset},r_0)\}$ $\cup$ 
$\{q_2(\ref{TDftd},s) : s \e \Foe(\psi,f,r_0)\}$, 
where the tags of these children are: 
\,$t(q_2(\ref{Tset},r_0)) = ((\psi,H(\pi\!:=\!\psi)$+$\psi r_0,A(r_0)),\min,w_2(\ref{Tset},r_0))$; and 
\,$t(q_2(\ref{TDftd},s)) = ((\psi,H(\pi\!:=\!\psi),f,r_0,s),\max,w_2(\ref{TDftd},s))$. \nl
So \,$w_1(\ref{TFor},r_0)$ = $\min[\{w_2(\ref{Tset},r_0)\}$ $\cup$ 
$\{w_2(\ref{TDftd},s) : s \e \Foe(\psi,f,r_0)\}]$. 

Since \,$|T[\pi,H$+$\pi r_0,A(r_0)]| < n$\, and 
\,$T(\pi,H$+$\pi r_0,A(r_0)) = v_2(\ref{Tset},r_0) = +1$, by $Y(n)$ we have 
\,$T(\psi,H(\pi\!:=\!\psi)$+$\psi r_0,A(r_0)) = w_2(\ref{Tset},r_0) = +1$. 
Hence \nl (*) \,$w_1(\ref{TFor},r_0)$ = 
$\min\{w_2(\ref{TDftd},s) : s \e \Foe(\psi,f,r_0)\}$. 

For each $s$ in $\Foe(\pi,f,r_0)$ let the set of children of 
$p_2(\ref{TDftd},s)$ be \nl
$\{p_3(\ref{Tset},t) : \pi t \nte H$ and $t \e R^s_d[f;s]\}$ $\cup$ 
$\{p_3(\ref{Tminus},s) : \pi's \nte H\}$, 
where the tags of these children are: 
\,$t(p_3(\ref{Tset},t)) = ((\pi,H$+$\pi t,A(t)), \min, v_3(\ref{Tset},t))$; 
\ and \ 
\,$t(p_3(\ref{Tminus},s)) = (-(\pi',H$+$\pi's,A(s)), -, v_3(\ref{Tminus},s))$. \nl
So \,$+1 = v_2(\ref{TDftd},s) = 
\max[\{v_3(\ref{Tset},t) : \pi t \nte H$ and $t \e R^s_d[f;s]\}$ $\cup$ 
$\{v_3(\ref{Tminus},s) : \pi's \nte H\}$. 

For each $s$ in $\Foe(\psi,f,r_0)$ let the set of children of 
$q_2(\ref{TDftd},s)$ be $\{q_3(\ref{Tset},t) : 
\psi t \nte H(\pi\!:=\!\psi)$ and $t \e R^s_d[f;s]\}$ $\cup$ 
$\{q_3(\ref{Tminus},s) : \psi's \nte H(\pi\!:=\!\psi)\}$, 
where the tags of these children are: 
\,$t(q_3(\ref{Tset},t)) = ((\psi,H(\pi\!:=\!\psi)$+$\psi t,A(t)), \min, w_3(\ref{Tset},t))$; \ and \ 
\,$t(q_3(\ref{Tminus},s)) = 
(-(\psi',H(\pi\!:=\!\psi)$+$\psi's,A(s)), -, w_3(\ref{Tminus},s))$. \nl
So \,$w_2(\ref{TDftd},s) = \max[\{w_3(\ref{Tset},t) : 
\psi t \nte H(\pi\!:=\!\psi)$ and $t \e R^s_d[f;s]\}$ $\cup$ 
$\{w_3(\ref{Tminus},s) : \psi's \nte H(\pi\!:=\!\psi)\}]$. 

Take any $s$ in $\Foe(\psi,f,r_0)$. 
We shall show that \,$w_2(\ref{TDftd},s) = +1$. 

Suppose there exists $t_0$ such that \,$\pi t_0 \nte H$\, and 
\,$t_0 \e R^s_d[f;s]$\, and 
\,$+1 = v_3(\ref{Tset},t_0) = T(\pi,H$+$\pi t_0,A(t_0))$. 
Then $p_3(\ref{Tset},t_0)$ exists, and \,$\psi t_0 \nte H(\pi\!:=\!\psi)$\, 
and so $q_3(\ref{Tset},t_0)$ exists. 
Since \,$|T[\pi,H$+$\pi t_0,A(t_0)]| < n$, then by $Y(n)$ we have 
\,$T(\psi,H(\pi\!:=\!\psi)$+$\psi t_0,A(t_0)) = w_3(\ref{Tset},t_0) = +1$. 
Hence \,$w_2(\ref{TDftd},s) = +1$. 

So suppose that such a $t_0$ does not exist. 
Then \,$v_3(\ref{Tminus},s) = +1$\, and so $p_3(\ref{Tminus},s)$ exists 
and \,$\pi's \nte H$. 
Hence \,$\psi's \nte H(\pi\!:=\!\psi)$, and so $q_3(\ref{Tminus},s)$ exists. 
Let the child of $p_3(\ref{Tminus},s)$ be $p_4(\ref{Tset},s)$ where 
\,$t(p_4(\ref{Tset},s)) = 
((\pi',H$+$\pi's,A(s)), \min, v_4(\ref{Tset},s))$. 
Then \,$+1 = v_3(\ref{Tminus},s) = -v_4(\ref{Tset},s)$. 
So \,$v_4(\ref{Tset},s) = -1$\, and hence \,$T(\pi',H$+$\pi's,A(s)) = -1$. 
By Theorem \ref{Thm:P()=+1 iff |- iff T()=+1}(\ref{equiv set}), 
\,$(\pi',H$+$\pi's) \not\vminus A(s)$. 
By Definition \ref{Defn:H(alpha := pi')} with \,$\alpha = \psi'$\, and 
Definition \ref{Defn:H(pi := psi)}, 
\,$H(\pi\!:=\!\psi)(\psi'\!:=\!\pi') = H$\, and 
\,$(\psi's)(\psi'\!:=\!\pi') = (\pi's)$. 

Let the child of $q_3(\ref{Tminus},s)$ be $q_4(\ref{Tset},s)$ where 
\,$t(q_4(\ref{Tset},s)) = 
((\psi',H(\pi\!:=\!\psi)$+$\psi's,A(s)), \min, w_4(\ref{Tset},s))$. 
Then \,$w_3(\ref{Tminus},s) = -w_4(\ref{Tset},s)$. 
Assume \,$w_4(\ref{Tset},s) = +1$. 
Then \,$T(\psi',H(\pi\!:=\!\psi)$+$\psi's,A(s)) = +1$\, and so by \nl
Theorem \ref{Thm:P()=+1 iff |- iff T()=+1}(\ref{equiv set}), 
\,$(\psi',H(\pi\!:=\!\psi)$+$\psi's) \vminus A(s)$. 
By Lemma \ref{Lem:If alpha|-x then pi'|-x.} with \,$\alpha = \psi'$, 
\,$(\pi',H(\pi\!:=\!\psi)(\psi'\!:=\!\pi')$++$(\psi's)(\psi'\!:=\!\pi')) \vminus A(s)$. 
But from the previous paragraph, this simplifies to 
\,$(\pi',H$+$\pi's) \vminus A(s)$. 
This contradiction shows that \,$w_4(\ref{Tset},s) = -1$. 
Hence \,$w_3(\ref{Tminus},s) = +1$. 
Therefore \,$w_2(\ref{TDftd},s) = +1$. 

Thus for all $s$ in $\Foe(\psi,f,r_0)$, \,$w_2(\ref{TDftd},s) = +1$. 
So by (*), \,$w_1(\ref{TFor},r_0) = +1$\, and hence \,$w_0 = +1$, as required. 

Since the subject of $p_0$ is $(\pi,H,x)$, $p_0$ does not satisfy 
T\ref{TFor}, or T\ref{TDftd}, or T\ref{Tminus}. 

Thus $Y(n)$, and hence the lemma, is proved by induction. 

\noindent \eop{EndProofLem\ref{Lem:If (pi,H) |- x then (psi,H(pi:=psi)) |- x.}} 
\end{Lem} 
\begin{Defn} \label{Defn:H(psi := beta)}  
If $H$ is a $\psi$-history then define 
\,$H(\psi\!:=\!\beta)$\, to be the sequence formed from $H$ by just 
replacing each $\psi$ by $\beta$, and each $\psi'$ by $\beta'$. 
\end{Defn} 
\begin{Lem} 
\label{Lem:Th(psi) subset Th(beta) subset Th(psi')} 
\raggedright \parindent = 1.5em

Suppose \,$\calP = (R,>)$\, is a plausible theory, $H$ is a $\psi$-history, and 
$x$ is either a formula or a finite set of formulas. 
\begin{compactenum}[1)]
\item \label{Th(psi) subset Th(beta)} 
	If \,$(\psi,H) \vminus x$\, then \,$(\beta,H(\psi\!:=\!\beta)) \vminus x$. 
\item \label{Th(beta) subset Th(psi')} 
	If \,$(\psi',H) \not\vminus x$\, then 
	\,$(\beta',H(\psi\!:=\!\beta)) \not\vminus x$. 
\end{compactenum}
Hence \,$\calP(\psi) \!\subseteq\! \calP(\beta)$\, and 
\,$\calP(\beta') \!\subseteq\! \calP(\psi')$. 

\noindent \pf{Proof} 

Suppose $(R,>)$ is a plausible theory, $\Ax$ is its set of axioms, 
$H$ is a $\psi$-history, and 
$x$ is either a formula or a finite set of formulas. 
Note that $H$ is a $\psi$-history iff $H$ is a $\psi'$-history. 
Let $Y(k)$ and $Z(k)$ denote the following conditional statements.
\begin{compactenum}[$A(k)$:]
\addtocounter{enumi}{24}
\item \label{Y(k)} 
	If $H$ is a $\psi$-history, 
	$x$ is either a formula or a finite set of formulas, 
	\,$T(\psi,H,x) = +1$, and \,$|T[\psi,H,x]| \leq k$\, 
	then \,$T(\beta,H(\psi\!:=\!\beta),x) = +1$. 
\item \label{Z(k)} 
	If $H$ is a $\psi'$-history, 
	$x$ is either a formula or a finite set of formulas, 
	\,$T(\psi',H,x) = -1$, and \,$|T[\psi',H,x]| \leq k$\, 
	then \,$T(\beta',H(\psi\!:=\!\beta),x) = -1$. 
\end{compactenum}
By Theorem \ref{Thm:P()=+1 iff |- iff T()=+1}(\ref{equiv set},\ref{equiv formula}) and 
Theorem \ref{AppThm:Decisiveness}(\ref{T(alpha,H,f) = +1 or -1}) 
it suffices to prove $Y(k)$ and $Z(k)$ by induction on $k$. 

Suppose \,$k = 1$. 

Let the antecedent of $Y(1)$ hold. 
Let $p_0$ be the root of $T[\psi,H,x]$ and 
$q_0$ be the root of $T[\beta,H(\psi\!:=\!\beta),x]$. 
Then $p_0$ has no children. 
If $p_0$ satisfies T\ref{Tset} then \,$x = \{\}$\, and 
so $q_0$ satisfies T\ref{Tset}. 
So by T\ref{Tset}, $T[\beta,H(\psi\!:=\!\beta),\{\}]$ has only one node and 
\,$T(\beta,H(\psi\!:=\!\beta),\{\}) = +1$. 
If $p_0$ satisfies T\ref{Tfact} then \,$x = f$\, and \,$\Ax \models f$. 
So $q_0$ satisfies T\ref{Tfact} and hence 
\,$T(\beta,H(\psi\!:=\!\beta),f) = +1$. 
Since $p_0$ has no children and the proof value of $p_0$ is $+1$, 
$p_0$ does not satisfy T\ref{Tfml}. 
Since the subject of $p_0$ is $(\psi,H,x)$, $p_0$ does not satisfy 
T\ref{TFor} or T\ref{TDftd} or T\ref{Tminus}. 
Thus $Y(1)$ holds. 

Let the antecedent of $Z(1)$ hold. 
Let $m_0$ be the root of $T[\psi',H,x]$ and 
$n_0$ be the root of $T[\beta',H(\psi\!:=\!\beta),x]$. 
Then $m_0$ has no children. 
Since \,$\pv(m_0) = T(\psi',H,x) = -1$, 
$m_0$ does not satisfy T\ref{Tset} or T\ref{Tfact}. 
If $m_0$ satisfies T\ref{Tfml} then \,$x = f$\, and 
for each \,$r \e R^s_d[f]$, \,$\psi'r \e H$. 
So $n_0$ satisfies T\ref{Tfml} and 
for each \,$r \e R^s_d[f]$, \,$\beta' r \e H(\psi\!:=\!\beta)$. 
Hence by T\ref{Tfml}, $n_0$ has no children and so 
\,$-1 = \pv(n_0) = T(\beta',H(\psi\!:=\!\beta),f)$. 
Since the subject of $m_0$ is $(\psi',H,x)$, $m_0$ does not satisfy 
T\ref{TFor} or T\ref{TDftd} or T\ref{Tminus}. 
Thus $Z(1)$ holds. 

Take any positive integer $k$. 
Suppose that both $Y(k)$ and $Z(k)$ are true. 
We shall prove both $Y(k\!+\!1)$ and $Z(k\!+\!1)$. 

Suppose the antecedent of $Y(k\!+\!1)$ holds and 
\,$|T[\psi,H,x]| = k\!+\!1$. 
We must show \,$T(\beta,H(\psi\!:=\!\beta),x) = +1$. 
Let $p_0$ be the root of $T[\psi,H,x]$ 
and $q_0$ be the root of $T[\beta,H(\psi\!:=\!\beta),x]$. 

If $p_0$ satisfies T\ref{Tset} then let $x$ be $F$. 
We see that $q_0$ also satisfies T\ref{Tset}. 
So \,$t(p_0) = ((\psi,H,F), \min, +1)$\, and 
\,$t(q_0) = ((\beta,H(\psi\!:=\!\beta),F), \min, w_0)$, where 
\,$w_0 = T(\beta,H(\psi\!:=\!\beta),F) \in \{+1,-1\}$. 
Let $\{p_{\!f} : f \e F\}$ be the set of children of $p_0$ 
and $\{q_{\!f} : f \e F\}$ be the set of children of $q_0$. 
Let $f$ be any formula in $F$. 
Then the subject of $p_{\!f}$ is $(\psi,H,f)$, 
and the subject of $q_{\!f}$ is $(\beta,H(\psi\!:=\!\beta),f)$. 
Also \,$|T[\psi,H,f]| \leq k$\, and the proof value of $p_{\!f}$ is $+1$ 
because $p_0$ is a min node with proof value $+1$. 
So by $Y(k)$ the proof value of $q_{\!f}$ is $+1$. 
But this is true for each $f$, so by T\ref{Tset}, \,$w_0 = +1$, as required. 

Since $p_0$ has a child, $p_0$ does not satisfy T\ref{Tfact}. 

If $p_0$ satisfies T\ref{Tfml} then let $x$ be $f$. 
We see that $q_0$ also satisfies T\ref{Tfml}. 
So \,$t(p_0) = ((\psi,H,f), \max, +1)$\, and 
\,$t(q_0) = ((\beta,H(\psi\!:=\!\beta),f), \max, w_0)$, where 
\,$w_0 = T(\beta,H(\psi\!:=\!\beta),f) \in \{+1, -1\}$. 

We shall adopt the following naming conventions. 
Each non-root node of $T[\psi,H,f]$ is denoted by $p_l(\#,y)$ where $l$ is the level of the node, $\#$ is the number in $[\ref{Tset}..\ref{Tminus}]$ such that the node satisfies T\#, and $y$ is a rule, or a formula, or a set, which distinguishes siblings. 
The proof value of $p_l(\#,y)$ will be denoted by $v_l(\#,y)$. 
For non-root nodes in $T[\beta,H(\psi\!:=\!\beta),f]$ we shall use $q_l(\#,y)$, and its proof value will be denoted by $w_l(\#,y)$. 

Let the set of children of $p_0$ be 
\,$\{p_1(\ref{TFor},r) : \psi r \nte H$\, and \,$r \e R^s_d[f]\}$, 
where the tags of these children are: \nl 
\,$t(p_1(\ref{TFor},r))$ = $((\psi,H,f,r),\min,v_1(\ref{TFor},r))$. 
So \,$+1 = \max\{v_1(\ref{TFor},r) : 
\psi r \nte H$\, and \,$r \e R^s_d[f]\}$. 
Hence there exists $r_0$ in $R^s_d[f]$ such that 
\,$\psi r_0 \nte H$\, and \,$v_1(\ref{TFor},r_0) = +1$. 
So \,$t(p_1(\ref{TFor},r_0)) = ((\psi,H,f,r_0),\min,+1)$. 

Let the set of children of $q_0$ be 
\,$\{q_1(\ref{TFor},r) : \beta r \nte H(\psi\!:=\!\beta)$\, and 
\,$r \e R^s_d[f]\}$, 
where the tags of these children are: $t(q_1(\ref{TFor},r))$ = 
$((\beta,H(\psi\!:=\!\beta),f,r),\min,w_1(\ref{TFor},r))$. 
So \,$w_0 = \max\{w_1(\ref{TFor},r) : \beta r \nte H(\psi\!:=\!\beta)$\, and 
\,$r \e R^s_d[f]\}$. 

Let the set of children of $p_1(\ref{TFor},r_0)$ be 
\,$\{p_2(\ref{Tset},r_0)\}$ $\cup$ 
$\{p_2(\ref{TDftd},s) : s \e \Foe(\psi,f,r_0)\}$, 
where the tags of these children are: 
\,$t(p_2(\ref{Tset},r_0)) = ((\psi,H$+$\psi r_0,A(r_0)),\min,v_2(\ref{Tset},r_0))$; and  
\,$t(p_2(\ref{TDftd},s)) = ((\psi,H,f,r_0,s),\max,v_2(\ref{TDftd},s))$. 
So \,$+1 = v_1(\ref{TFor},r_0)$ = $\min[\{v_2(\ref{Tset},r_0)\}$ $\cup$ 
$\{v_2(\ref{TDftd},s) : s \e \Foe(\psi,f,r_0)\}]$. 
Hence \,$v_2(\ref{Tset},r_0) = +1$; and for each $s$ in $\Foe(\psi,f,r_0)$, 
\,$v_2(\ref{TDftd},s) = +1$. 

Let the set of children of $q_1(\ref{TFor},r_0)$ be 
\,$\{q_2(\ref{Tset},r_0)\}$ $\cup$ 
$\{q_2(\ref{TDftd},s) : s \e \Foe(\beta,f,r_0)\}$, 
where the tags of these children are: 
\,$t(q_2(\ref{Tset},r_0)) = ((\beta,H(\psi\!:=\!\beta)$+$\beta r_0,A(r_0)),\min,w_2(\ref{Tset},r_0))$; and 
\,$t(q_2(\ref{TDftd},s)) = ((\beta,H(\psi\!:=\!\beta),f,r_0,s),\max,w_2(\ref{TDftd},s))$. \nl
So \,$w_1(\ref{TFor},r_0)$ = $\min[\{w_2(\ref{Tset},r_0)\}$ $\cup$ 
$\{w_2(\ref{TDftd},s) : s \e \Foe(\beta,f,r_0)\}]$. 

Since \,$|T[\psi,H$+$\psi r_0,A(r_0)]| \leq k$\, and 
\,$T(\psi,H$+$\psi r_0,A(r_0)) = v_2(\ref{Tset},r_0) = +1$, by $Y(k)$ we have 
\,$T(\beta,H(\psi\!:=\!\beta)$+$\beta r_0,A(r_0)) = w_2(\ref{Tset},r_0) = +1$. 
Hence \nl (*) \,$w_1(\ref{TFor},r_0)$ = 
$\min\{w_2(\ref{TDftd},s) : s \e \Foe(\beta,f,r_0)\}$. 

For each $s$ in $\Foe(\psi,f,r_0)$ let the set of children of 
$p_2(\ref{TDftd},s)$ be \nl
$\{p_3(\ref{Tset},t) : \psi t \nte H$ and $t \e R^s_d[f;s]\}$ $\cup$ 
$\{p_3(\ref{Tminus},s) : \psi's \nte H\}$, 
where the tags of these children are: 
\,$t(p_3(\ref{Tset},t)) = ((\psi,H$+$\psi t,A(t)), \min, v_3(\ref{Tset},t))$; 
\ and \ 
\,$t(p_3(\ref{Tminus},s)) = (-(\psi',H$+$\psi's,A(s)), -, v_3(\ref{Tminus},s))$. \nl
So \,$+1 = v_2(\ref{TDftd},s) = 
\max[\{v_3(\ref{Tset},t) : \psi t \nte H$ and $t \e R^s_d[f;s]\}$ $\cup$ 
$\{v_3(\ref{Tminus},s) : \psi's \nte H\}]$. 

For each $s$ in $\Foe(\beta,f,r_0)$ let the set of children of 
$q_2(\ref{TDftd},s)$ be $\{q_3(\ref{Tset},t) : 
\beta t \nte H(\psi\!:=\!\beta)$ and $t \e R^s_d[f;s]\}$ $\cup$ 
$\{q_3(\ref{Tminus},s) : \beta' s \nte H(\psi\!:=\!\beta)\}$, 
where the tags of these children are: 
\,$t(q_3(\ref{Tset},t)) = ((\beta,H(\psi\!:=\!\beta)$+$\beta t,A(t)), \min, w_3(\ref{Tset},t))$; \ and \ 
\,$t(q_3(\ref{Tminus},s)) = 
(-(\beta',H(\psi\!:=\!\beta)$+$\beta' s,A(s)), -, w_3(\ref{Tminus},s))$. \nl
So \,$w_2(\ref{TDftd},s) = \max[\{w_3(\ref{Tset},t) : 
\beta t \nte H(\psi\!:=\!\beta)$ and $t \e R^s_d[f;s]\}$ $\cup$ 
$\{w_3(\ref{Tminus},s) : \beta' s \nte H(\psi\!:=\!\beta)\}]$. 

Take any $s$ in $\Foe(\beta,f,r_0)$. 
We shall show that \,$w_2(\ref{TDftd},s) = +1$. 
Observe that \,$\Foe(\psi,f,r_0) = \Foe(\beta,f,r_0)$. 

Suppose there exists $t_0$ such that \,$\psi t_0 \nte H$\, and 
\,$t_0 \e R^s_d[f;s]$\, and 
\,$+1 = v_3(\ref{Tset},t_0) = T(\psi,H$+$\psi t_0,A(t_0))$. 
Then $p_3(\ref{Tset},t_0)$ exists. 
Since \,$\psi t_0 \e H$\, iff \,$\beta t_0 \e H(\psi\!:=\!\beta)$, 
we have \,$\beta t_0 \nte H(\psi\!:=\!\beta)$\, 
and so $q_3(\ref{Tset},t_0)$ exists. 
Since \,$|T[\psi,H$+$\psi t_0,A(t_0)]| \leq k$, then by $Y(k)$ we have 
\,$T(\beta,H(\psi\!:=\!\beta)$+$\beta t_0,A(t_0)) = w_3(\ref{Tset},t_0) = +1$. 
Hence \,$w_2(\ref{TDftd},s) = +1$. 

So suppose that such a $t_0$ does not exist. 
Then \,$v_3(\ref{Tminus},s) = +1$\, and so $p_3(\ref{Tminus},s)$ exists 
and \,$\psi's \nte H$. 
Since \,$\psi' s \e H$\, iff \,$\beta' s \e H(\psi\!:=\!\beta)$, 
we have \,$\beta' s \nte H(\psi\!:=\!\beta)$, 
and so $q_3(\ref{Tminus},s)$ exists. 
Let the child of $p_3(\ref{Tminus},s)$ be $p_4(\ref{Tset},s)$ where 
\,$t(p_4(\ref{Tset},s)) = 
((\psi',H$+$\psi's,A(s)), \min, v_4(\ref{Tset},s))$. 
Then \,$+1 = v_3(\ref{Tminus},s) = -v_4(\ref{Tset},s)$. 
So \,$v_4(\ref{Tset},s) = -1$\, and hence \,$T(\psi',H$+$\psi's,A(s)) = -1$. 
Since \,$|T[\psi',H$+$\psi' s,A(s)]| \leq k$, then by $Z(k)$, 
\,$T(\beta',H(\psi\!:=\!\beta)$+$\beta' s,A(s)) = -1$. 

Let the child of $q_3(\ref{Tminus},s)$ be $q_4(\ref{Tset},s)$ where 
\,$t(q_4(\ref{Tset},s)) = 
((\beta',H(\psi\!:=\!\beta)$+$\beta' s,A(s)), \min, w_4(\ref{Tset},s))$. 
Then \,$w_3(\ref{Tminus},s)$ = $-w_4(\ref{Tset},s)$ 
= $-T(\beta',H(\psi\!:=\!\beta)$+$\beta' s,A(s)) = +1$. 
Therefore \,$w_2(\ref{TDftd},s) = +1$. 

Thus for all $s$ in $\Foe(\beta,f,r_0)$, \,$w_2(\ref{TDftd},s) = +1$. 
So by (*), \,$w_1(\ref{TFor},r_0) = +1$\, and hence \,$w_0 = +1$, as required. 

Since the subject of $p_0$ is $(\psi,H,x)$, $p_0$ does not satisfy 
T\ref{TFor}, or T\ref{TDftd}, or T\ref{Tminus}. 

Thus $Y(k\!+\!1)$ is proved. 

\smallskip

To prove $Z(k\!+\!1)$ we suppose the antecedent of $Z(k\!+\!1)$ holds and 
\,$|T[\psi',H,x]| = k\!+\!1$. 
We must show \,$T(\beta',H(\psi\!:=\!\beta),x) = -1$. 
Let $m_0$ be the root of $T[\psi',H,x]$ 
and $n_0$ be the root of $T[\beta',H(\psi\!:=\!\beta),x]$. 
Then $m_0$ has a child and \,$\pv(m_0) = T(\psi',H,x) = -1$. 

If $m_0$ satisfies T\ref{Tset} then let $x$ be $F$. 
We see that $n_0$ also satisfies T\ref{Tset}. 
So \,$t(m_0) = ((\psi',H,F), \min, -1)$\, and 
\,$t(n_0) = ((\beta',H(\psi\!:=\!\beta),F), \min, \pv(n_0))$, where 
\,$\pv(n_0) = T(\beta',H(\psi\!:=\!\beta),F) \in \{+1,-1\}$. 
Let $\{m_{\!f} : f \e F\}$ be the set of children of $m_0$ 
and $\{n_{\!f} : f \e F\}$ be the set of children of $n_0$. 
Let $f$ be any formula in $F$. 
Then the subject of $m_{\!f}$ is $(\psi',H,f)$, 
and the subject of $n_{\!f}$ is $(\beta',H(\psi\!:=\!\beta),f)$. 
Also \,$|T[\psi',H,f]| \leq k$. 
There exists \,$f_0 \e F$\, such at \,$\pv(m_{\!f_0}) = -1$\, 
because $m_0$ is a min node with proof value $-1$. 
So by $Z(k)$, \,$\pv(n_{\!f_0}) = T(\beta',H(\psi\!:=\!\beta),f_0) = -1$. 
But $n_0$ is a min node, so \,$\pv(n_0) = -1$, as required. 

Since \,$\pv(m_0) = T(\psi',H,x) = -1$, $m_0$ does not satisfy T\ref{Tfact}. 

If $m_0$ satisfies T\ref{Tfml} then let $x$ be $f$. 
We see that $n_0$ also satisfies T\ref{Tfml}. 
So \,$t(m_0) = ((\psi',H,f), \max, -1)$\, and 
\,$t(n_0) = ((\beta',H(\psi\!:=\!\beta),f), \max, \pv(n_0))$, where 
\,$\pv(n_0) = T(\beta',H(\psi\!:=\!\beta),f) \in \{+1, -1\}$. 

We shall adopt the following naming conventions. 
Each non-root node of $T[\psi',H,f]$ is denoted by $m_l(\#,y)$ where $l$ is the level of the node, $\#$ is the number in $[\ref{Tset}..\ref{Tminus}]$ such that the node satisfies T\#, and $y$ is a rule, or a formula, or a set, which distinguishes siblings. 
For non-root nodes in $T[\beta',H(\psi\!:=\!\beta),f]$ 
we shall use $n_l(\#,y)$. 

Let the set of children of $m_0$ be 
\,$\{m_1(\ref{TFor},r) : \psi' r \nte H$\, and \,$r \e R^s_d[f]\}$, 
where the tags of these children are: 
\,$t(m_1(\ref{TFor},r))$ = $((\psi',H,f,r),\min,\pv(m_1(\ref{TFor},r)))$. 
Recall that $m_0$ has at least one child. 
So \,$-1 = \max\{\pv(m_1(\ref{TFor},r)) : 
\psi' r \nte H$\, and \,$r \e R^s_d[f]\}$. 
Hence if \,$\psi' r \nte H$\, and \,$r \e R^s_d[f]$\, 
then \,$\pv(m_1(\ref{TFor},r)) = -1$. 
Therefore \,$t(m_1(\ref{TFor},r))$ = $((\psi',H,f,r),\min,-1)$. 

Let the set of children of $n_0$ be 
\,$\{n_1(\ref{TFor},r) : \beta' r \nte H(\psi\!:=\!\beta)$\, and 
\,$r \e R^s_d[f]\}$, 
where the tags of these children are: $t(n_1(\ref{TFor},r))$ = 
$((\beta',H(\psi\!:=\!\beta),f,r),\min,\pv(n_1(\ref{TFor},r)))$. 
So \,$\pv(n_0) = \max\{\pv(n_1(\ref{TFor},r)) : 
\beta' r \nte H(\psi\!:=\!\beta)$\, and \,$r \e R^s_d[f]\}$. 
If $n_0$ does not have a child then \,$\pv(n_0) = \max\{\} = -1$, as required. 
So suppose that $n_0$ has at least one child. 

If \,$\psi' r \nte H$\, and \,$r \e R^s_d[f]$\, 
then let the set of children of $m_1(\ref{TFor},r)$ be 
\,$\{m_2(\ref{Tset},r)\}$ $\cup$ $\{m_2(\ref{TDftd},s) : s \e R[\neg f;r]\}$, 
where the tags of these children are: \,$t(m_2(\ref{Tset},r)) 
= ((\psi',H$+$\psi'r,A(r)),\min,\pv(m_2(\ref{Tset},r)))$; and  
\,$t(m_2(\ref{TDftd},s)) = ((\psi',H,f,r,s),\max,\pv(m_2(\ref{TDftd},s)))$. 
So \,$-1 = \pv(m_1(\ref{TFor},r))$ = $\min[\{\pv(m_2(\ref{Tset},r))\}$ $\cup$ 
$\{\pv(m_2(\ref{TDftd},s)) : s \e R[\neg f;r]\}]$. 
Therefore either \,$\pv(m_2(\ref{Tset},r)) = -1$\, or there exists 
$s_0$ in $R[\neg f;r]$ such that \,$\pv(m_2(\ref{TDftd},s_0)) = -1$. 

If \,$\beta' r \nte H(\psi\!:=\!\beta)$\, and \,$r \e R^s_d[f]$\, 
then let the set of children of $n_1(\ref{TFor},r)$ be \nl
\,$\{n_2(\ref{Tset},r)\}$ $\cup$ 
$\{n_2(\ref{TDftd},s) : s \e \Foe(\beta',f,r)\}$, 
where the tags of these children are: \,$t(n_2(\ref{Tset},r)) 
= ((\beta',H(\psi\!:=\!\beta)$+$\beta' r,A(r)),\min,\pv(n_2(\ref{Tset},r)))$; 
and \,$t(n_2(\ref{TDftd},s)) 
= ((\beta',H(\psi\!:=\!\beta),f,r,s),\max,\pv(n_2(\ref{TDftd},s)))$. \nl
So \,$\pv(n_1(\ref{TFor},r))$ = $\min[\{\pv(n_2(\ref{Tset},r))\}$ $\cup$ 
$\{\pv(n_2(\ref{TDftd},s)) : s \e \Foe(\beta',f,r)\}]$. 

We show that for each $r$ in $R^s_d[f]$ such that 
\,$\beta' r \nte H(\psi\!:=\!\beta)$, \,$\pv(n_1(\ref{TFor},r)) = -1$. 

We have \,$|T[\psi',H$+$\psi' r,A(r)]| \leq k$. 
If \,$-1 = \pv(m_2(\ref{Tset},r)) = T(\psi',H$+$\psi'r,A(r))$\, 
then by $Z(k)$, \,$\pv(n_2(\ref{Tset},r)) 
= T(\beta',H(\psi\!:=\!\beta)$+$\beta' r,A(r)) = -1$. 
Hence \,$\pv(n_1(\ref{TFor},r)) = -1$. 

So suppose there exists $s_0$ in $R[\neg f;r]$ such that 
\,$T(\psi',H,f,r,s_0) = \pv(m_2(\ref{TDftd},s_0)) = -1$. 
Then \,$s_0 \e \Foe(\beta',f,r)$. 
We shall show that \,$T(\beta',H(\psi\!:=\!\beta),f,r,s_0) = \pv(n_2(\ref{TDftd},s_0)) = -1$, 
and hence that \,$\pv(n_1(\ref{TFor},r)) = -1$. 

Let the set of children of $m_2(\ref{TDftd},s_0)$ be \nl
$\{m_3(\ref{Tset},t) : \psi' t \nte H$ and $t \e R^s_d[f;s_0]\}$ $\cup$ 
$\{m_3(\ref{Tminus},s_0) : \psi s_0 \nte H\}$, 
where the tags of these children are: 
\,$t(m_3(\ref{Tset},t)) = 
((\psi',H$+$\psi't,A(t)),\min,\pv(m_3(\ref{Tset},t)))$, 
\ and \ \,$t(m_3(\ref{Tminus},s_0)) = 
(-(\psi,H$+$\psi s_0,A(s_0)), -, \pv(m_3(\ref{Tminus},s_0)))$. \nl
So \,$-1 = \pv(m_2(\ref{TDftd},s_0)) = 
\max[\{\pv(m_3(\ref{Tset},t)) : \psi' t \nte H$ and $t \e R^s_d[f;s_0]\}$ 
$\cup$ $\{\pv(m_3(\ref{Tminus},s_0)) : \psi s_0 \nte H\}]$. 
Hence for all $t$ in $R^s_d[f;s_0]$ such that \,$\psi' t \nte H$\, 
we have \,$-1 = \pv(m_3(\ref{Tset},t)) = T(\psi',H$+$\psi't,A(t))$. 
Also if \,$\psi s_0 \nte H$\, then 
\,$-1 = \pv(m_3(\ref{Tminus},s_0)) = T(-(\psi,H$+$\psi s_0,A(s_0)))$. 

Let the set of children of $n_2(\ref{TDftd},s_0)$ be \nl
$\{n_3(\ref{Tset},t) : \beta' t \nte H(\psi\!:=\!\beta)$ and 
$t \e R^s_d[f;s_0]\}$ $\cup$ 
$\{n_3(\ref{Tminus},s_0) : \beta s_0 \nte H(\psi\!:=\!\beta)\}$, 
where the tags of these children are: 
\,$t(n_3(\ref{Tset},t)) = 
((\beta',H(\psi\!:=\!\beta)$+$\beta' t,A(t)),\min,\pv(n_3(\ref{Tset},t)))$, 
\ and \ \,$t(n_3(\ref{Tminus},s_0)) = 
(-(\beta,H(\psi\!:=\!\beta)$+$\beta s_0,A(s_0)),-,\pv(n_3(\ref{Tminus},s_0)))$. \nl
So \,$\pv(n_2(\ref{TDftd},s_0)) = \max[\{\pv(n_3(\ref{Tset},t)) : 
\beta' t \nte H(\psi\!:=\!\beta)$ and $t \e R^s_d[f;s_0]\}$ $\cup$ 
$\{\pv(n_3(\ref{Tminus},s_0)) : \beta s_0 \nte H(\psi\!:=\!\beta)\}]$. 
If $n_2(\ref{TDftd},s_0)$ has no children then 
\,$\pv(n_2(\ref{TDftd},s_0)) = \max\{\} = -1$, as required. 
So suppose that $n_2(\ref{TDftd},s_0)$ has at least one child. 

Case 1: $n_3(\ref{Tset},t)$ is a child of $n_2(\ref{TDftd},s_0)$. \nl
Then \,$\beta' t \nte H(\psi\!:=\!\beta)$\, and \,$t \e R^s_d[f;s_0]$. 
Hence \,$\psi' t \nte H$. 
So from above, \,$-1 = \pv(m_3(\ref{Tset},t)) = T(\psi',H$+$\psi't,A(t))$. 
Also \,$|T[\psi',H$+$\psi't,A(t)]| \leq k$. 
So by $Z(k)$, \,$-1 = T(\beta',H(\psi\!:=\!\beta)$+$\beta' t,A(t)) 
= \pv(t(n_3(\ref{Tset},t)))$. 

Case 2: $n_3(\ref{Tminus},s_0)$ is a child of $n_2(\ref{TDftd},s_0)$. \nl
Then \,$\beta s_0 \nte H(\psi\!:=\!\beta)$. 
Hence \,$\psi s_0 \nte H$. 
So from above, 
\,$-1 = \pv(m_3(\ref{Tminus},s_0)) = T(-(\psi,H$+$\psi s_0,A(s_0)))$. 
Therefore \,$T(\psi,H$+$\psi s_0,A(s_0)) = +1$. 
Also \,$|T[\psi,H$+$\psi s_0,A(s_0)]| \leq k$. 
So by $Y(k)$, \,$T(\beta,H(\psi\!:=\!\beta)$+$\beta s_0,A(s_0)) = +1$. 
But \,$\pv(n_3(\ref{Tminus},s_0)) 
= -T(\beta,H(\psi\!:=\!\beta)$+$\beta s_0,A(s_0)) = - +1 = -1$. 

These two cases show that \,$\pv(n_2(\ref{TDftd},s_0)) = -1$, as required. 
Hence \,$\pv(n_1(\ref{TFor},r)) = -1$. 
Therefore \,$\pv(n_0) = -1$. 
Thus $Z(k\!+\!1)$ is proved. 

\smallskip

Therefore $Y(k)$ and $Z(k)$ are proved by induction, and so the lemma is proved.

\noindent \eop{EndProofLem\ref{Lem:Th(psi) subset Th(beta) subset Th(psi')}} 
\end{Lem} 
\begin{Defn} \label{Defn:(gamma:lambda)}  
Suppose \,$\{\alpha, \gamma, \lambda\} \!\subseteq\! \Alg$, 
$H$ is a $\alpha$-history, 
and $T$ is an evaluation tree of some plausible theory. 
If \,$\alpha \nte \{\gamma,\gamma',\lambda,\lambda'\}$\, then define 
\,$\alpha(\gamma\!:\!\lambda) = \alpha$; else define 
\,$\gamma(\gamma\!:\!\lambda) = \lambda$, 
\,$\gamma'(\gamma\!:\!\lambda) = \lambda'$, 
\,$\lambda(\gamma\!:\!\lambda) = \gamma$, and 
\,$\lambda'(\gamma\!:\!\lambda) = \gamma'$. 

If \,$H = (\alpha_1r_1, ..., \alpha_nr_n)$\, then define 
\,$H(\gamma\!:\!\lambda) = 
(\alpha_1(\gamma\!:\!\lambda)r_1, ..., \alpha_n(\gamma\!:\!\lambda)r_n)$. 

Define $T(\gamma\!:\!\lambda)$ to be the tree formed from $T$ by 
only changing the subject of each node as follows. 
For each node $p$ of $T$ replace $\alg(p)$ by $\alg(p)(\gamma\!:\!\lambda)$, 
and replace $\Hist(p)$ by $\Hist(p)(\gamma\!:\!\lambda)$. 
\end{Defn} 
\begin{Lem} 
\label{Lem:If > is empty then psi' = pi'.}
\raggedright \parindent = 1.5em

Suppose \,$\calP = (R,>)$\, is a plausible theory such that $>$ is empty. 
If $T$ is an evaluation tree of $\calP$ 
then $T(\pi\!:\!\psi)$ is an evaluation tree of $\calP$. \nl
Hence \ $\calP(\psi) = \calP(\pi)$ \ and \ $\calP(\pi') = \calP(\psi')$. 

\noindent \pf{Proof}

Let \,$\calP = (R,>)$\, be a plausible theory such that $>$ is empty. 
Then \,$\Foe(\psi',f,r) = \{\} = \Foe(\pi',f,r)$\, 
and \,$\Foe(\pi,f,r) = R[\neg f] = \Foe(\psi,f,r)$. 
Let $Y(n)$ denote the following conditional statement. \nl 
``If $T$ is an evaluation tree of $\calP$ and \,$|T| \leq n$\, 
then $T(\pi\!:\!\psi)$ is an evaluation tree of $\calP$." \nl
By Definition \ref{Defn:PlausibleTheory, PlausibleLogic}, 
it suffices to prove $Y(n)$ by induction on $n$. 

Suppose \,$n=1$\, and the antecedent of $Y(1)$ holds. 
Let the root of $T$ be $p_0$ and \,$\alg(p_0) = \alpha$. 
If \,$\alpha \nte \{\pi,\psi,\psi',\pi'\}$\, then 
\,$T(\pi\!:\!\psi) = T$\, and so $Y(1)$ holds. 
So suppose \,$\alpha \e \{\pi,\psi,\psi',\pi'\}$. 
Let the root of $T(\pi\!:\!\psi)$ be $q_0$, and let \,$\alg(q_0) = \lambda$. 
So \,$\alpha(\pi\!:\!\psi) = \lambda$. 
Since $p_0$ has no children, $q_0$ has no children. \nl
If $p_0$ satisfies T\ref{Tset} then let \,$\Subj(p_0) = (\alpha,H,\{\})$. 
Hence \,$\Subj(q_0) = (\lambda,H(\pi\!:\!\psi),\{\})$\, 
and so $q_0$ satisfies T\ref{Tset}. 
Thus $T(\pi\!:\!\psi)$ is an evaluation tree of $\calP$. \nl
If $p_0$ satisfies T\ref{Tfact} then let \,$\Subj(p_0) = (\alpha,H,f)$. 
Hence \,$\Subj(q_0) = (\lambda,H(\pi\!:\!\psi),f)$\, 
and so $q_0$ satisfies T\ref{Tfact}. 
Thus $T(\pi\!:\!\psi)$ is an evaluation tree of $\calP$. \nl
If $p_0$ satisfies T\ref{Tfml} then let \,$\Subj(p_0) = (\alpha,H,f)$. 
Hence \,$\Subj(q_0) = (\lambda,H(\pi\!:\!\psi),f)$. 
Since $p_0$ has no children, \,$S(p_0) = \{\}$. 
So if \,$r \e R^s_d[f]$\, then \,$\alpha r \e H$. 
Now \,$\alpha r \e H$\, iff \,$\lambda r \e H(\pi\!:\!\psi)$. 
Hence \,$S(q_0) = \{\}$\, and so $q_0$ satisfies T\ref{Tfml}. 
Thus $T(\pi\!:\!\psi)$ is an evaluation tree of $\calP$. \nl
Since $p_0$ and $q_0$ have no children, 
$p_0$ and $q_0$ satisfy neither T\ref{TFor} nor T\ref{Tminus}. \nl 
If $p_0$ satisfies T\ref{TDftd} then let \,$\Subj(p_0) = (\alpha,H,f,r,s)$. 
Hence \,$\Subj(q_0) = (\lambda,H(\pi\!:\!\psi),f,r,s)$. 
Since $p_0$ has no children, \,$S(p_0) = \{\}$. 
Hence if \,$t \e R^s_d[f;s]$\, then \,$\alpha t \e H$. 
Now \,$\alpha t \e H$\, iff \,$\lambda t \e H(\pi\!:\!\psi)$. 
Also \,$\alpha's \e H$. 
Hence \,$\lambda' s \e H(\pi\!:\!\psi)$. 
So \,$S(q_0) = \{\}$. 
Thus $q_0$ satisfies T\ref{TDftd} and so 
$T(\pi\!:\!\psi)$ is an evaluation tree of $\calP$. 

All cases have been considered and so $Y(1)$ holds. 

If $T$ is any tree and $p$ is any node of $T$ for which $\Subj(p)$ is defined, 
then define the set $S(p,T)$ of subjects of the children of $p$ in $T$ by 
\,$S(p,T) = \{\Subj(c) : c$ is a child of $p$ in $T\}$. 

Take any integer $n$ such that \,$n \geq 1$. 
Suppose that $Y(n)$ is true. 
We shall prove $Y(n\!+\!1)$. 
Suppose the antecedent of $Y(n\!+\!1)$ holds and that \,$|T| = n\!+\!1$. 
Let the root of $T$ be $p_0$ and \,$\alg(p_0) = \alpha$. 
If \,$\alpha \nte \{\pi,\psi,\psi',\pi'\}$\, then 
\,$T(\pi\!:\!\psi) = T$\, and so $Y(n\!+\!1)$ holds. 
So suppose \,$\alpha \e \{\pi,\psi,\psi',\pi'\}$. 
Let the root of $T(\pi\!:\!\psi)$ be $q_0$, and let \,$\alg(q_0) = \lambda$. 
So \,$\alpha(\pi\!:\!\psi) = \lambda$. 

If $p_0$ satisfies T\ref{Tset} then let \,$\Subj(p_0) = (\alpha,H,F)$. 
Hence \,$\Subj(q_0) = (\lambda,H(\pi\!:\!\psi),F)$. 
So $q_0$ satisfies T\ref{Tset}. 
Recall that \,$S(p_0,T[\alpha,H,F])$ = $\{(\alpha,H,f) : f \e F\}$. 
So \,$S(q_0,T[\alpha,H,F](\pi\!:\!\psi))$ 
= $\{(\lambda,H(\pi\!:\!\psi),f) : f \e F\}$ 
= $S(q_0,T[\lambda,H(\pi\!:\!\psi),F])$, by T\ref{Tset}. 
But for each $(\alpha,H,f)$ in $S(p_0,T[\alpha,H,F])$, $T[\alpha,H,f]$ is an 
evaluation tree of $\calP$ and \,$|T[\alpha,H,f]| \leq n$. 
So by $Y(n)$, $T[\alpha,H,f](\pi\!:\!\psi)$ 
is an evaluation tree of $\calP$. 
Hence \,$T[\alpha,H,f](\pi\!:\!\psi)$ 
= $T[\lambda,H(\pi\!:\!\psi),f]$. 
Thus \,$T(\pi\!:\!\psi)$ = $T[\alpha,H,F](\pi\!:\!\psi)$ 
= $T[\lambda,H(\pi\!:\!\psi),F]$\, 
which is an evaluation tree of $\calP$. 

Since $p_0$ has a child, $p_0$ does not satisfy T\ref{Tfact}. 

If $p_0$ satisfies T\ref{Tfml} then let \,$\Subj(p_0) = (\alpha,H,f)$. 
Hence \,$\Subj(q_0) = (\lambda,H(\pi\!:\!\psi),f)$. 
So $q_0$ satisfies T\ref{Tfml}. 
Recall that \,$S(p_0,T[\alpha,H,f])$ = 
$\{(\alpha,H,f,r) : \alpha r \nte H$\, and \,$r \e R^s_d[f]\}$. 
Since \,$\alpha r \e H$\, iff \,$\lambda r \e H(\pi\!:\!\psi)$\, we have 
\,$\alpha r \nte H$\, iff \,$\lambda r \nte H(\pi\!:\!\psi)$. 
So \,$S(q_0,T[\alpha,H,f](\pi\!:\!\psi))$ 
= $\{(\lambda,H(\pi\!:\!\psi),f,r) : 
	\alpha r \nte H$\, and \,$r \e R^s_d[f]\}$ 
= $\{(\lambda,H(\pi\!:\!\psi),f,r) : 
	\lambda r \nte H(\pi\!:\!\psi)$\, and \,$r \e R^s_d[f]\}$ 
= $S(q_0,T[\lambda,H(\pi\!:\!\psi),f])$, by T\ref{Tfml}. 
But for each $(\alpha,H,f,r)$ in $S(p_0,T[\alpha,H,f])$, $T[\alpha,H,f,r]$ is an 
evaluation tree of $\calP$ and \,$|T[\alpha,H,f,r]| \leq n$. 
So by $Y(n)$, $T[\alpha,H,f,r](\pi\!:\!\psi)$ 
is an evaluation tree of $\calP$. 
Hence \,$T[\alpha,H,f,r](\pi\!:\!\psi)$ 
= $T[\lambda,H(\pi\!:\!\psi),f,r]$. 
Thus \,$T(\pi\!:\!\psi)$ = $T[\alpha,H,f](\pi\!:\!\psi)$ 
= $T[\lambda,H(\pi\!:\!\psi),f]$\, 
which is an evaluation tree of $\calP$. 

If $p_0$ satisfies T\ref{TFor} then let \,$\Subj(p_0) = (\alpha,H,f,r)$. 
Hence \,$\Subj(q_0) = (\lambda,H(\pi\!:\!\psi),f,r)$. 
Since \,$\alpha r \e H$\, iff \,$\lambda r \e H(\pi\!:\!\psi)$\, we have 
\,$\alpha r \nte H$\, iff \,$\lambda r \nte H(\pi\!:\!\psi)$. 
So $q_0$ satisfies T\ref{TFor}. 
Recall that \,$S(p_0,T[\alpha,H,f,r])$ = $\{(\alpha,H$+$\alpha r,A(r))\}$ 
$\!\cup\!$ $\{(\alpha,H,f,r,s) : s \e \Foe(\alpha,f,r)\}$. 
Since \,$\Foe(\alpha,f,r) = \Foe(\lambda,f,r)$, 
\,$S(q_0,T[\alpha,H,f,r](\pi\!:\!\psi))$ 
= $\{(\lambda,H(\pi\!:\!\psi)$+$\lambda r,A(r))\}$ $\!\cup\!$ 
	$\{(\lambda,H(\pi\!:\!\psi),f,r,s) : s \e \Foe(\alpha,f,r)\}$ 
= $S(q_0,T[\lambda,H(\pi\!:\!\psi),f,r])$, by T\ref{TFor}. 
But $T[\alpha,H$+$\alpha r,A(r)]$ is an evaluation tree of $\calP$ and 
\,$|T[\alpha,H$+$\alpha r,A(r)]| \leq n$. 
So by $Y(n)$, $T[\alpha,H$+$\alpha r,A(r)](\pi\!:\!\psi)$ 
is an evaluation tree of $\calP$. 
Hence \,$T[\alpha,H$+$\alpha r,A(r)](\pi\!:\!\psi)$ 
= $T[\lambda,H(\pi\!:\!\psi)$+$\lambda r,A(r)]$. 
Also for each $(\alpha,H,f,r,s)$ in $S(p_0,T[\alpha,H,f,r])$, 
$T[\alpha,H,f,r,s]$ is an evaluation tree of $\calP$ and 
\,$|T[\alpha,H,f,r,s]| \leq n$. 
So by $Y(n)$, $T[\alpha,H,f,r,s](\pi\!:\!\psi)$ 
is an evaluation tree of $\calP$. 
Hence \,$T[\alpha,H,f,r,s](\pi\!:\!\psi)$ 
= $T[\lambda,H(\pi\!:\!\psi),f,r,s]$. 
Thus \,$T(\pi\!:\!\psi)$ = $T[\alpha,H,f,r](\pi\!:\!\psi)$ 
= $T[\lambda,H(\pi\!:\!\psi),f,r]$\, 
which is an evaluation tree of $\calP$. 

If $p_0$ satisfies T\ref{TDftd} then let \,$\Subj(p_0) = (\alpha,H,f,r,s)$. 
Then \,$\alpha \neq \pi'$. 
Since \,$s \e \Foe(\alpha,f,r)$, \,$\Foe(\alpha,f,r) \neq \{\}$\, 
and so \,$\alpha \neq \psi'$. 
Therefore \,$\alpha \e \{\pi,\psi\}$\, and so \,$\lambda \e \{\pi,\psi\}$. 
Now \,$\Subj(q_0) = (\lambda,H(\pi\!:\!\psi),f,r,s)$. 
Since \,$\alpha r \e H$\, iff \,$\lambda r \e H(\pi\!:\!\psi)$\, we have 
\,$\alpha r \nte H$\, iff \,$\lambda r \nte H(\pi\!:\!\psi)$. 
So $q_0$ satisfies T\ref{TDftd}. 
Recall that \,$S(p_0,T[\alpha,H,f,r,s])$ = $\{(\alpha,H$+$\alpha t,A(t)) : 
\alpha t \nte H$\, and \,$t \e R^s_d[f;s]\}$ $\!\cup\!$ 
$\{-(\alpha',H$+$\alpha' s,A(s)) : \alpha' s \nte H\}$. 
Also since \,$\alpha' s \e H$\, iff \,$\lambda' s \e H(\pi\!:\!\psi)$\, 
we have \,$\alpha' s \nte H$\, iff \,$\lambda' s \nte H(\pi\!:\!\psi)$. 
So \,$S(q_0,T[\alpha,H,f,r,s](\pi\!:\!\psi))$ \nl
= $\{(\lambda,H(\pi\!:\!\psi)$+$\lambda t,A(t)) : 
\alpha t \nte H$\, and \,$t \e R^s_d[f;s]\}$ $\!\cup\!$ 
$\{-(\lambda',H(\pi\!:\!\psi)$+$\lambda' s,A(s)) : \alpha' s \nte H\}$ \nl 
= $\{(\lambda,H(\pi\!:\!\psi)$+$\lambda t,A(t)) : 
\lambda t \nte H(\pi\!:\!\psi)$\, and \,$t \e R^s_d[f;s]\}$ $\!\cup\!$ 
$\{-(\lambda',H(\pi\!:\!\psi)$+$\lambda' s,A(s)) : 
\lambda' s \nte H(\pi\!:\!\psi)\}$ \nl 
= $S(q_0,T[\lambda,H(\pi\!:\!\psi),f,r,s])$, by T\ref{TDftd}. 

But for each $(\alpha,H$+$\alpha t,A(t))$ in $S(p_0,T[\alpha,H,f,r,s])$, 
$T[\alpha,H$+$\alpha t,A(t)]$ is an evaluation tree of $\calP$ and 
\,$|T[\alpha,H$+$\alpha t,A(t)]| \leq n$. 
So by $Y(n)$, $T[\alpha,H$+$\alpha t,A(t)](\pi\!:\!\psi)$ 
is an evaluation tree of $\calP$. 
Hence \,$T[\alpha,H$+$\alpha t,A(t)](\pi\!:\!\psi)$ 
= $T[\lambda,H(\pi\!:\!\psi)$+$\lambda t,A(t)]$. 
Also if \,$-(\alpha',H$+$\alpha' s,A(s)) \in S(p_0,T[\alpha,H,f,r,s])$, 
then $T[-(\alpha',H$+$\alpha' s,A(s))]$ is an evaluation tree of $\calP$ and 
\,$|T[-(\alpha',H$+$\alpha' s,A(s))]| \leq n$. 
So by $Y(n)$, $T[-(\alpha',H$+$\alpha' s,A(s))](\pi\!:\!\psi)$ 
is an evaluation tree of $\calP$. 
Hence \,$T[-(\alpha',H$+$\alpha' s,A(s))](\pi\!:\!\psi)$ 
= $T[-(\lambda',H(\pi\!:\!\psi)$+$\lambda' s,A(s))]$. 

Thus \,$T(\pi\!:\!\psi)$ = $T[\alpha,H,f,r,s](\pi\!:\!\psi)$ 
= $T[\lambda,H(\pi\!:\!\psi),f,r,s]$\, 
which is an evaluation tree of $\calP$. 

If $p_0$ satisfies T\ref{Tminus} then let \,$\Subj(p_0) = -(\alpha',H,F)$. 
Then \,$\alpha \e \{\pi,\psi\}$\, and so \,$\lambda \e \{\psi,\pi\}$. 
Hence \,$\Subj(q_0) = -(\lambda',H(\pi\!:\!\psi),F)$. 
So $q_0$ satisfies T\ref{Tminus}. 
Recall that \,$S(p_0,T[-(\alpha',H,F)])$ = $\{(\alpha',H,F)\}$. 
So \,$S(q_0,T[-(\alpha',H,F)](\pi\!:\!\psi))$ 
= $\{(\lambda',H(\pi\!:\!\psi),F)\}$ 
= $S(q_0,T[-(\lambda',H(\pi\!:\!\psi),F)])$, by T\ref{Tminus}. 
But $T[\alpha',H,F]$ is an evaluation tree of $\calP$ and 
\,$|T[\alpha',H,F]| \leq n$. 
So by $Y(n)$, $T[\alpha',H,F](\pi\!:\!\psi)$ 
is an evaluation tree of $\calP$. 
Hence \,$T[\alpha',H,F](\pi\!:\!\psi)$ 
= $T[\lambda',H(\pi\!:\!\psi),F]$. 
Thus \,$T(\pi\!:\!\psi)$ = $T[-(\alpha',H,F)](\pi\!:\!\psi)$ 
= $T[-(\lambda',H(\pi\!:\!\psi),F)]$\, 
which is an evaluation tree of $\calP$. 

Therefore $Y(n\!+\!1)$, and hence the lemma, is proved by induction. 

\noindent \eop{EndProofLem\ref{Lem:If > is empty then psi' = pi'.}} 
\end{Lem}  
\begin{Thm} [Theorem \ref{Thm:Hierarchy} The proof algorithm hierarchy]
\label{Thm:The proof algorithm hierarchy} 
\raggedright \parindent = 1.5em

Suppose \,$\calP = (R,>)$\, is a plausible theory. \nl
1) \,$\calP(\varphi) \subseteq \calP(\pi) \subseteq \calP(\psi) 
\subseteq \calP(\beta) = \calP(\beta') \subseteq \calP(\psi') \subseteq \calP(\pi')$. \nl
2) If $>$ is empty then \,$\calP(\varphi) \subseteq \calP(\pi) = \calP(\psi) 
\subseteq \calP(\beta) = \calP(\beta') \subseteq \calP(\psi') = \calP(\pi')$. 

\noindent \pf{Proof} 

Suppose \,$\calP = (R,>)$\, is a plausible theory. 
By Lemma \ref{Lem:If (phi,I)|- x then (alpha,H)|- x.}, 
\,$\calP(\varphi) \subseteq \calP(\pi)$. 
By Lemma \ref{Lem:If (pi,H) |- x then (psi,H(pi:=psi)) |- x.}, 
\,$\calP(\pi) \subseteq \calP(\psi)$. 
By Lemma \ref{Lem:Th(psi) subset Th(beta) subset Th(psi')}(\ref{Th(psi) subset Th(beta)}), 
\,$\calP(\psi) \subseteq \calP(\beta)$. 
By Lemma \ref{Lem:beta is isomorphic to beta'}, 
\,$\calP(\beta) = \calP(\beta')$. 
By Lemma \ref{Lem:Th(psi) subset Th(beta) subset Th(psi')}(\ref{Th(beta) subset Th(psi')}), 
\,$\calP(\beta') \subseteq \calP(\psi')$. 
By Lemma \ref{Lem:If alpha|-x then pi'|-x.}, 
\,$\calP(\psi') \subseteq \calP(\pi')$. 
So part (1) holds. 

Part (2) holds by part (1) and Lemma \ref{Lem:If > is empty then psi' = pi'.}. 

\noindent \eop{EndProofThm\ref{Thm:The proof algorithm hierarchy}} 
\end{Thm} 

\bibliographystyle{theapa}
\bibliography{Bib170401}

\end{document}